\documentclass{article}
\usepackage{iclr2027_conference,times}

\usepackage{amsmath,amsfonts,bm}

\def\eqref#1{equation~\ref{#1}}
\def\1{\bm{1}}

\DeclareMathAlphabet{\mathsfit}{\encodingdefault}{\sfdefault}{m}{sl}
\SetMathAlphabet{\mathsfit}{bold}{\encodingdefault}{\sfdefault}{bx}{n}

\usepackage{hyperref}
\usepackage{url}
\usepackage{xspace}
\usepackage{algorithm}
\usepackage{algorithmic}
\usepackage{graphicx}
\usepackage{fontawesome5}

\newcommand{\hflogo}{\raisebox{-0.25em}{\includegraphics[height=1.15em]{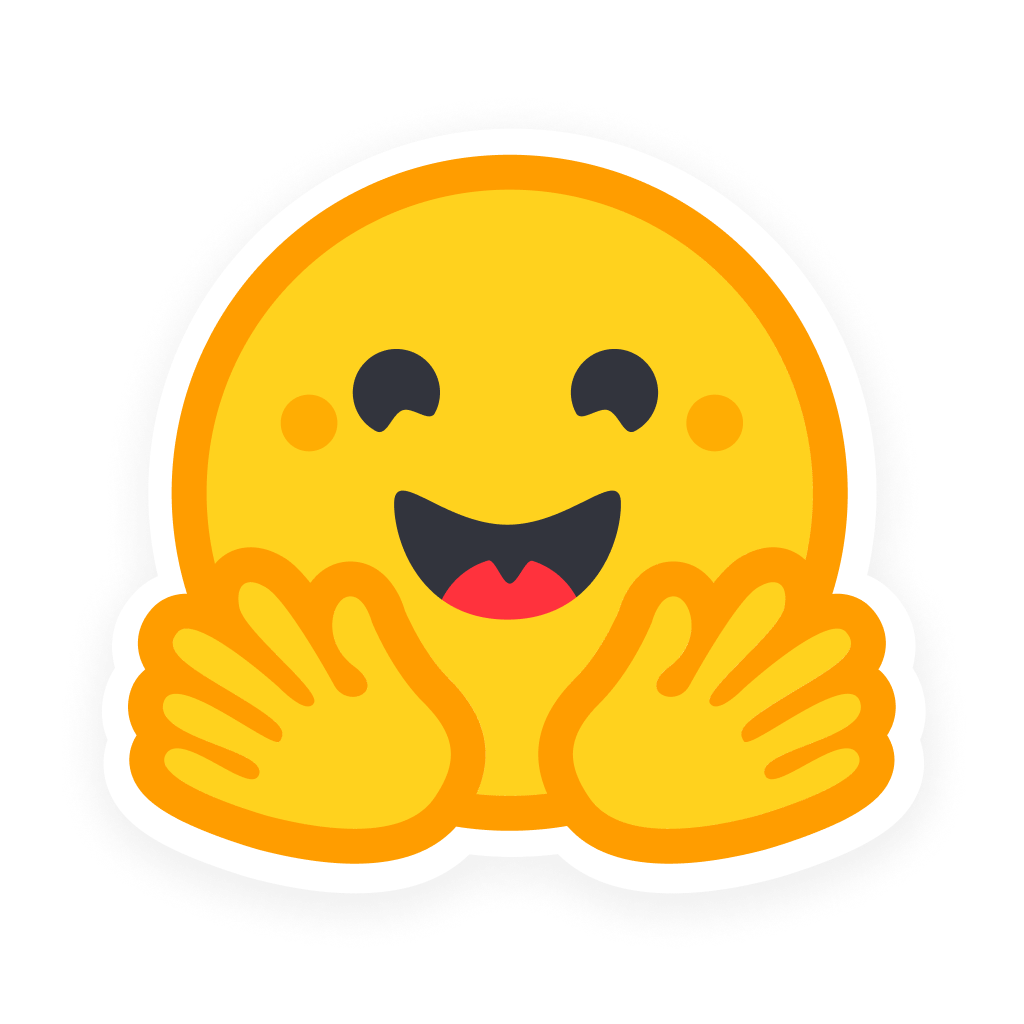}}}

\definecolor{cornellred}{rgb}{0.7, 0.11, 0.11}
\definecolor{cadmiumgreen}{rgb}{0.0, 0.42, 0.24}
\definecolor{aliceblue}{rgb}{0.91, 0.94, 0.97}
\definecolor{darkblue}{rgb}{0.83, 0.89, 0.97}
\definecolor{Red7}{rgb}{0.941, 0.243, 0.243}
\definecolor{Green7}{RGB}{55, 178, 77}
\definecolor{Blue9}{rgb}{0.098,0.3,0.9}
\definecolor{tocnavy}{RGB}{27,58,107}

\hypersetup{
  linkcolor = cornellred,
  citecolor  = cadmiumgreen,
  colorlinks = true,
  urlcolor = Blue9
}

\title{Learning the 4D World \\through Point Trajectory Forecasting}

\author{
  \hspace*{-1mm}Byungwoo Jeon$^{1}$\thanks{Equal contribution.} \quad
  Sunghwan Hong$^{2}$\footnotemark[1] \quad
  Dohyeon Kim$^{3}$ \quad
  Jonghoon Lee$^{1}$ \\
  \textbf{Jungwoo Park}$^{1}$ \quad
  \textbf{Inhee Lee}$^{3}$ \quad
  \textbf{Doosan Baek}$^{6}$ \quad
  \textbf{Jinil Kim}$^{6}$ \quad
  \textbf{Taehyun Lim}$^{6}$ \\
  \textbf{Hanbyul Joo}$^{3,5}$ \quad
  \textbf{Marc Pollefeys}$^{2}$ \quad
  \textbf{Saining Xie}$^{4}$\thanks{Equal advising.} \quad
  \textbf{Jinwoo Shin}$^{1,5}$\footnotemark[2] \\
  $^{1}$KAIST \quad $^{2}$ETH Zürich \quad $^{3}$Seoul National University \quad $^{4}$New York University \\ $^{5}$RLWRLD \quad $^{6}$ Physics Sim Lab
}

\newcommand{\llname}{Video-Point Model (VPM)}

\newcommand{\sname}{VPM\xspace}

\newcommand{\bname}{VP-WorldBench\xspace}

\usepackage{tikz}
\usepackage{tikzducks}
\usepackage{xcolor}
\usepackage[table]{xcolor}
\usepackage{enumitem}
\usepackage{booktabs}
\usepackage{multirow}
\usepackage{hyperref}
\usepackage{cleveref}
\usepackage{subcaption}
\usepackage{xcolor}
\usepackage{pifont}
\usepackage{titletoc}
\usepackage{array}

\newcolumntype{L}[1]{>{\raggedright\arraybackslash}p{#1}}

\titlecontents{lsection}[1.8em]
  {\addvspace{3.5pt}\bfseries}
  {\contentslabel{1.8em}}
  {\hspace*{-1.8em}}
  {\hfill\contentspage}
\titlecontents{lsubsection}[4.3em]
  {}
  {\contentslabel{2.5em}}
  {\hspace*{-2.5em}}
  {\titlerule*[0.6pc]{.}\contentspage}

\newcommand{\eg}{\textit{e.g.}}
\newcommand{\ie}{\textit{i.e.}}
\newcommand{\cmark}{\textcolor{green!60!black}{\ding{51}}}
\newcommand{\xmark}{\textcolor{red!80!black}{\ding{55}}}

\iclrfinalcopy
\begin{document}

\maketitle
\maketitle
\vspace{-8mm}
{\centering
  \href{https://rootyjeon.github.io/video-point-model/}{\textcolor{tocnavy}{\faGlobe~\textbf{Project Page}}}
  \hspace{2em}
  \href{https://huggingface.co/datasets/rooty2020/VPM-Data}{\hflogo~\textcolor{tocnavy}{\textbf{VPM-Data}}}
  \hspace{2em}
  \href{https://huggingface.co/datasets/rooty2020/VP-WorldBench}{\hflogo~\textcolor{tocnavy}{\textbf{VP-WorldBench}}}
\par}
\vspace{-1mm}
\begin{figure}[h]
    \centering\centerline{\includegraphics[width=\textwidth]{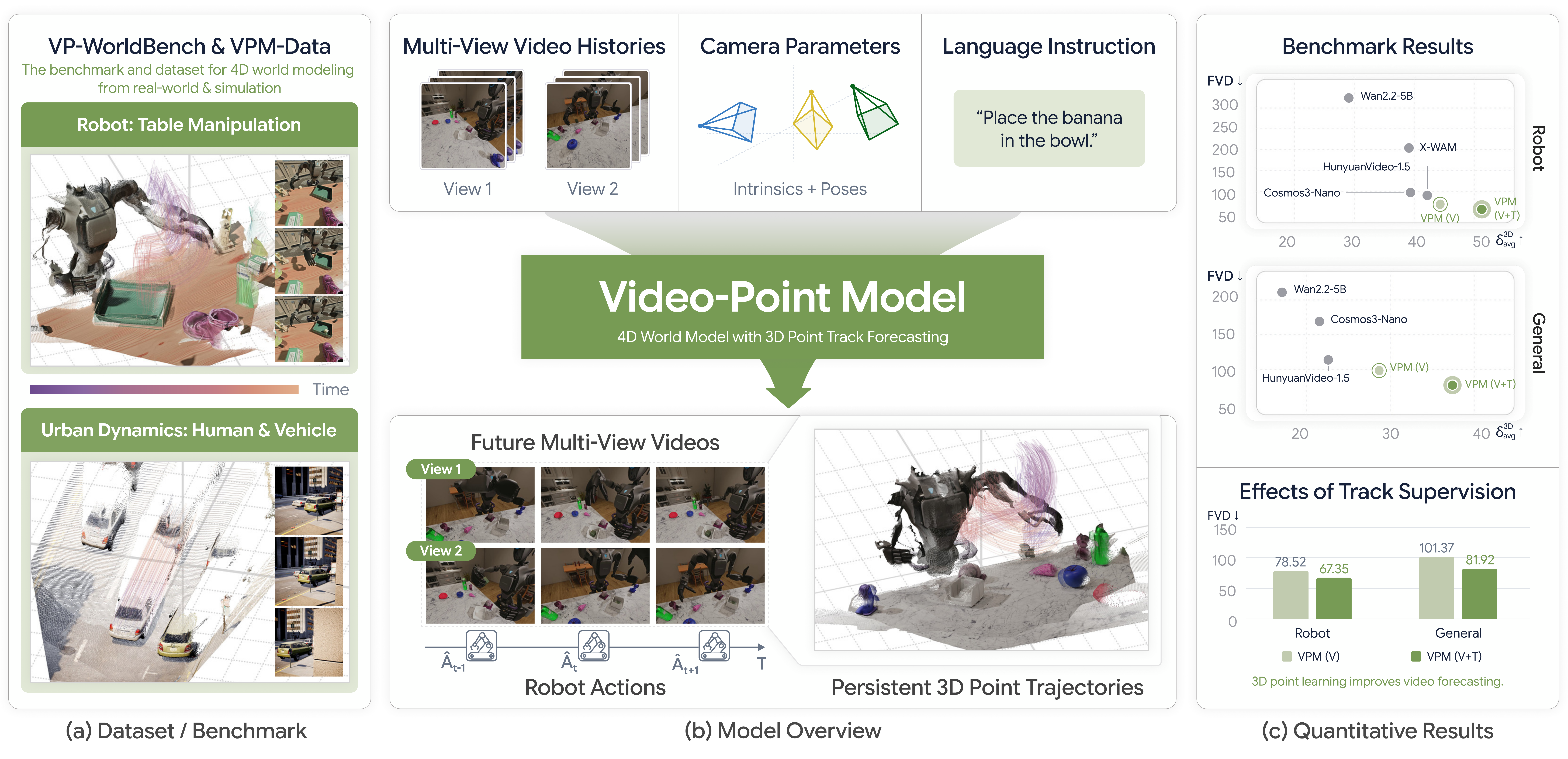}}
    \vspace{-0.05in}
    \caption{\textbf{Teaser.}
We introduce simulated and real-world multi-view videos paired with 3D point tracks (left), jointly forecast future videos and point trajectories (center), and show that point supervision improves
visual forecasting (right).}
\end{figure}

\begin{abstract}
Reliable multi-view forecasting requires videos from different cameras to describe the same evolving 3D scene. 
To this end, we introduce \llname{}, a world model that conditions on multi-view video histories and a language instruction to jointly forecast future videos from arbitrary viewpoints and the 3D trajectories of persistent scene points.
We extend a pretrained world model with camera-conditioned multi-view processing that integrates complementary observations across cameras, and add a trajectory head that predicts how physical points move through time. 
Each trajectory tracks a single point across all views, and supervising these trajectories alongside video generation therefore anchors appearance changes in every camera to a shared description of 3D motion.
To support training and evaluation, we build VPM-Data, a simulation corpus pairing synchronized multi-view videos with persistent point trajectories, and VP-WorldBench, a forecasting benchmark spanning simulated and real environments. 
We further extend \sname{} to robot control by adding robot actions as a token modality in the generator. 
Action tokens share the point-supervised backbone and are denoised jointly with future videos, thereby grounding the policy in the same 3D-aware representation.
On VP-WorldBench and existing 4D generation benchmarks, point supervision improves both video
fidelity and 3D motion accuracy over a video-only counterpart, and in simulated and real-robot experiments, it also raises the manipulation success of the action-enabled model.
\end{abstract}
\section{Introduction}
\label{sec:intro}

World models aim to anticipate how scenes evolve from past
observations. Recent models pursue this goal by predicting
future videos and, for robot interactions, the associated
actions~\citep{nvidia2026cosmos3}.
However, visual plausibility alone does not establish whether
these forecasts describe coherent 3D motion.
Multi-view histories provide richer evidence for predicting
this motion by revealing scene structure and ongoing dynamics
that may be obscured in individual cameras~\citep{rajivc2025mvtracker}.
To exploit this evidence, a model must relate observations
across cameras to a common prediction of scene evolution.
Its forecasts from arbitrary viewpoints should therefore
depict the same future event, despite differences in
perspective and visibility.

To connect these predictions, we build on geometric
supervision~\citep{lee2026mvtrack4gen} for video modeling, including pointmap prediction
and trajectory-derived representation
alignment~\citep{liu2026geometry,yang20264d}.
Specifically, we investigate jointly forecasting future images
and persistent 3D point trajectories. Because each trajectory
follows the same scene point over time, it provides a common
description of motion across cameras. Learning this motion
alongside image generation encourages the model to relate
appearance changes to the movement of identifiable points.
Unlike conventional tracking, both predictions extend beyond
the observed history, requiring the model to anticipate
correspondences without access to future frames.
Our hypothesis is that learning where scene points will move
helps the visual predictor produce more faithful forecasts.

Accordingly, we introduce \textbf{\sname{}},
a multi-view video--point world model built on
Cosmos~3~\citep{nvidia2026cosmos3}.
Given posed video histories and a language instruction,
\sname{} predicts future videos from arbitrary viewpoints
together with the trajectories of points selected from
observed frames. To integrate information across cameras,
we process their video tokens in a shared sequence, using
ray-based embeddings to encode camera geometry and a view
axis in positional encoding to distinguish camera streams.
A trajectory forecasting head is optimized jointly with
video generation, allowing explicit point-motion supervision
to shape the representation used for visual prediction.

To support this learning objective, we construct
\textbf{\sname{}-Data}, a simulation corpus covering tabletop
manipulation and urban dynamics. Each sequence pairs synchronized
multi-view videos with persistent point trajectories, camera
geometry, visibility annotations, and language descriptions,
providing aligned supervision for future appearance and motion.
We complement this training corpus with \textbf{\bname},
an evaluation suite spanning simulated and real environments.
Its simulation component assesses visual and geometric forecasts
on the same held-out events, allowing image quality and motion
accuracy to be examined together.

We evaluate \sname{} on our benchmark and existing benchmarks,
assessing future-video quality and 3D trajectory accuracy.
Controlled ablations demonstrate that point supervision improves
visual forecasting. We further investigate whether this benefit
extends to robot control through an action-enabled variant
that jointly learns video prediction, point trajectories,
and robot commands. Across simulated benchmarks and real-robot
experiments, point supervision consistently improves task success
rates of the action-enabled model. Our contributions are threefold:
\begin{itemize}
    \item We introduce \sname{}, a camera-conditioned model
    that jointly forecasts multi-view videos and persistent
    3D point trajectories, using explicit point-motion supervision
    to improve visual prediction from arbitrary viewpoints.

    \item We develop \sname{}-Data and \bname,
    providing aligned visual and geometric training data
    and a forecasting evaluation suite spanning simulated
    manipulation, urban dynamics, and real-world recordings.

    \item Through extensive experiments and controlled ablations,
    we demonstrate the benefits of point supervision for
    multi-view forecasting and assess its contribution to
    robot action learning in an action-enabled extension.
\end{itemize}
\section{Related Work}
\label{sec:related_work}

\noindent{\bf Geometry- and Motion-Aware Video Prediction.}
Auxiliary objectives improve video generation by supervising internal
features with 2D point tracking~\citep{jeong2025track4gen,lee20263d} or
optical flow~\citep{chefer2025videojam}, or by aligning them with
video and 3D foundation models~\citep{zhang2505videorepa, wu2026geometry,han2026geometric}. For multi-view prediction~\citep{balice2026no},
SV4D~\citep{xie2025sv4d} synthesizes novel views of observed videos,
Geometry-aware 4D Video Generation~\citep{liu2026geometry} forecasts RGB and
pointmaps, and X-WAM~\citep{guo2026unified} adds actions to multi-view RGB-D
forecasting, while GEM-4D~\citep{zhou2026gem},
MECo-WAM~\citep{zhang2026learning}, and 4D-WAM~\citep{yang20264d} distill
geometry or trajectory fields into video and video--action representations.
Unlike these single-view or per-frame targets, our persistent 3D trajectories
retain point identity across views and time. Closest to ours,
MVTrack4Gen~\citep{lee2026mvtrack4gen} supervises novel-view generation with
multi-view point tracking; we differ from this in that we \textit{forecast} beyond observed frames, lift supervision from 2D to metric 3D, and extend it to action learning.

% move from 2D to metric 3D and to action learning.

\noindent\textbf{Video World Models and Action Learning.}
Video foundation models such as
HunyuanVideo-1.5~\citep{wu2025hunyuanvideo} and
Wan2.2-TI2V-5B~\citep{team2025wan} provide pretrained
representations for visual prediction.
VideoVLA~\citep{shen2026videovla} and
Cosmos~3~\citep{nvidia2026cosmos3} jointly predict future
observations and actions, while SpatialVAM~\citep{li2026spatialvam}
pairs multi-view RGB prediction with spatial action heatmaps.
We instead supervise multi-view forecasting with future 3D point
motion and examine whether it also benefits action learning.

\noindent\textbf{3D Motion Forecasting and Benchmarks.}
Point trackers estimate correspondences within observed
videos~\citep{karaev2025cotracker3}, and
TAPIP3D~\citep{zhang2026tapip3d} and
MVTracker~\citep{rajivc2025mvtracker} extend tracking to 3D from
monocular and multi-view input.
MotionForesight~\citep{bharadhwaj2026motionforesight} and
MolmoMotion~\citep{zhang2026molmomotion} instead forecast motion
beyond observed frames, the latter also supporting video generation
and robot learning. 
Among benchmarks, TAPVid-MV~\citep{koppula2026tapvidmv} provides
multi-view tracking annotations and
PointMotionBench~\citep{zhang2026molmomotion} targets
language-conditioned motion prediction; ours adds paired visual
and geometric supervision for jointly evaluating future appearance
and point motion.
\section{Method}
\label{sec:method}

\subsection{Problem formulation}
\begin{figure}[t]
    \centering
    \centerline{\includegraphics[width=\textwidth]{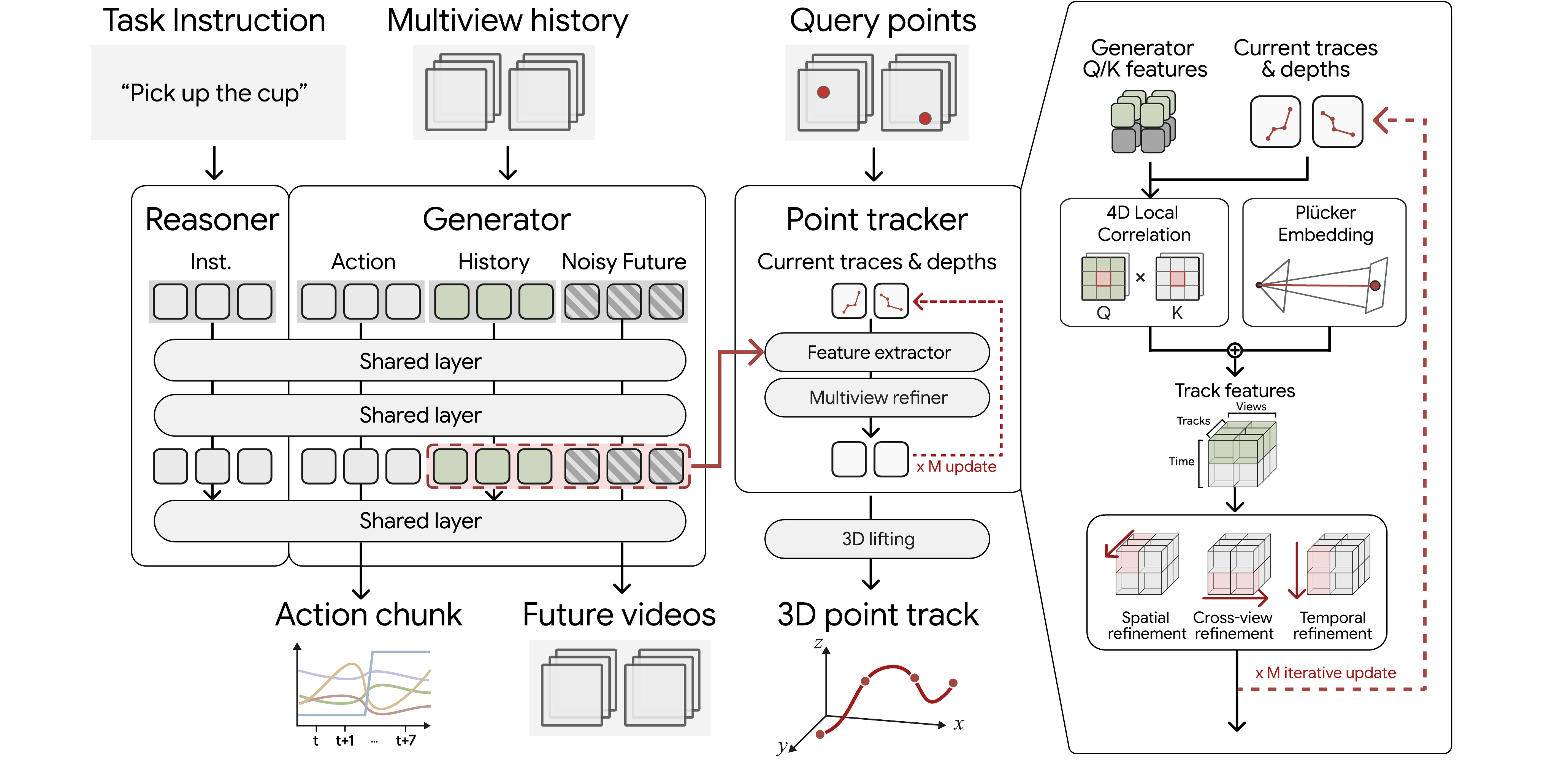}}
       \caption{\textbf{Overview of \sname{}.} From posed multi-view histories and a task instruction, \sname{} forecasts future videos and the trajectories of points queried from observed frames. The 3D point tracker iteratively refines point positions and depths before lifting them to 3D. Joint point supervision shapes the generator's representation, and the action-enabled extension also outputs robot action chunks.}\vspace{-10pt}
    \label{fig:overview}
\end{figure}

Let $\mathcal{V}_{\mathrm{in}}$ and
$\mathcal{V}_{\mathrm{out}}$ denote the input and output camera
sets. For camera $v$ at time $\tau$, let
$\mathbf{O}^v_\tau \in \mathbb{R}^{H \times W \times 3}$,
$\mathbf{K}^v$, and $\mathbf{T}^v_\tau \in \mathrm{SE}(3)$
denote its RGB frame, intrinsics, and camera-to-world pose.
Given the observed histories
$\{\mathbf{O}^v_{t-h'+1:t}\}_{v\in\mathcal{V}_{\mathrm{in}}}$,
the camera parameters of the observed and requested future views,
and a language instruction $\ell$, the model predicts the next
$h$ frames at arbitrary output viewpoints and the future
trajectories of $N$ points selected from the observed frames.
Each point $n$ is specified by a query
$(t_n,v_n,\mathbf{u}_n)$ within the history, and
$\mathbf{P}_{\tau,n}\in\mathbb{R}^3$ denotes its persistent
3D position. Trajectories are expressed in a fixed reference
frame defined by camera $v_{\mathrm{ref}}$ at the first observed
timestep $t_0=t-h'+1$. Our architecture overview is shown in \Cref{fig:overview}.

\subsection{Preliminaries: Mixture-of-Transformers for World Modeling}
\label{sec:prelim}

We build on Cosmos~3~\citep{nvidia2026cosmos3}, whose
Mixture-of-Transformers backbone comprises a reasoner and a
generator. The two towers are initialized from a pretrained
vision-language model and maintain separate layer parameters,
interacting through a joint attention operation.
In the forecasting configuration, the reasoner processes instruction
tokens with causal attention restricted to its own sequence,
while generator tokens attend to both sequences.

The generator is trained with rectified flow
matching~\citep{liu2022flow,lipman2022flow}.
Given clean future-video tokens $\mathbf{x}_0$ and Gaussian
noise $\boldsymbol{\epsilon}$, we construct
$\mathbf{x}_\sigma=(1-\sigma)\mathbf{x}_0+
\sigma\boldsymbol{\epsilon}$ at noise level
$\sigma\in[0,1]$. The denoiser $v_\theta$ predicts
$\boldsymbol{\epsilon}-\mathbf{x}_0$, conditioned on
the noise level and context $c$:
\begin{equation}
\mathcal{L}_{\mathrm{FM}}
=
\mathbb{E}_{\mathbf{x}_0,c,\boldsymbol{\epsilon},\sigma}
\left[
\left\|
v_\theta(\mathbf{x}_\sigma,\sigma,c)
-
(\boldsymbol{\epsilon}-\mathbf{x}_0)
\right\|_2^2
\right].
\label{eq:fm}
\end{equation}
Here, $c$ contains the observed history and conditioning
information. Clean conditioning tokens are excluded from
the generation loss.

\subsection{Camera-Conditioned Multi-View Prediction}
\label{sec:camera}

We pack latent video tokens across cameras into a single
sequence on a shared temporal grid. Observed history tokens
provide context, while noisy future tokens are associated
with the requested output cameras.
To distinguish camera geometry from view identity, we introduce
per-token camera embeddings and an additional view axis in
rotary positional encoding.

\noindent\textbf{Camera embeddings.}
For each video token $i$, we compute the
Pl\"ucker ray~\citep{sitzmann2021light,he2024cameractrl}
of its image region using the corresponding intrinsics and
pose. Expressed in the fixed reference frame, the ray is
$\mathbf{r}_i=(\mathbf{d}_i,\mathbf{o}_i\times\mathbf{d}_i)$,
where $\mathbf{o}_i$ and $\mathbf{d}_i$ are its origin and
unit direction. A lightweight encoder produces
$\mathbf{e}_i=\phi(\mathbf{r}_i)$, which conditions the
backbone through additive features and camera-dependent
rotations of attention queries and keys:
\begin{equation}
\begin{aligned}
\mathbf{h}_i
&\leftarrow \mathbf{h}_i+\mathbf{W}\mathbf{e}_i,\\
\mathbf{q}_i
&\leftarrow
\mathbf{R}\big(\psi(\mathbf{e}_i)\big)\mathbf{q}_i,
\qquad
\mathbf{k}_i
\leftarrow
\mathbf{R}\big(\psi(\mathbf{e}_i)\big)\mathbf{k}_i.
\end{aligned}
\label{eq:camera}
\end{equation}
Here, $\mathbf{W}$ is a learned projection and
$\mathbf{R}(\cdot)$ applies per-head rotary phases predicted
by $\psi$. These query and key rotations are applied in each
transformer layer~\citep{park2026redirector}, allowing attention
to depend on the camera geometry associated with its tokens.

\noindent\textbf{4D RoPE.}
Tokens from different cameras can share the same temporal
and spatial coordinates. Standard 3D rotary positional encoding
therefore does not distinguish their view identities.
We add a fourth axis, assigning $p_{\mathrm{view},i}=v$
to tokens from camera $v$. The rotary phase becomes
\begin{equation}
\theta_{i,k}
=
p_{t,i}\omega^t_k
+
p_{h,i}\omega^h_k
+
p_{w,i}\omega^w_k
+
p_{\mathrm{view},i}\omega^{\mathrm{view}}_k,
\label{eq:rope4d}
\end{equation}
where $\omega^t$, $\omega^h$, and $\omega^w$ are the
pretrained frequencies, and $\omega^{\mathrm{view}}$ is
learnable. Initializing the view frequencies to zero recovers
the original 3D positional encoding at initialization.
This formulation preserves the temporal and spatial indexing
within each view while accommodating different numbers of
camera streams. 

\subsection{Joint Video and Trajectory Learning}
\label{sec:track}

We connect point-motion supervision to video prediction in
two ways. A trajectory head forecasts point locations from
intermediate generator features, while a correspondence
objective directly supervises the attention features from
which those trajectories are inferred.
Both objectives provide geometric learning signals to the
shared video representation.

\noindent\textbf{Trajectory forecasting head.}
The head operates on query and key features from an intermediate
self-attention layer, reshaped into per-view feature grids
$\mathbf{F}^v_q$ and $\mathbf{F}^v_k$ over the latent frames.
For each query $(t_n,v_n,\mathbf{u}_n)$, it samples a source
feature and computes correlations with key features around
the current point estimate in each view and timestep.
A transformer with attention factorized over space, time,
and views~\citep{koo2026mv} refines
the estimates over $M$ iterations.
At each latent timestep, the head predicts a 2D position
$\widehat{\mathbf{u}}^v_{\tau,n}$, depth
$\widehat{d}^v_{\tau,n}$, and confidence
$\widehat{c}^v_{\tau,n}$ for every view.

We unproject each estimate into the fixed reference frame.
Let
$\mathbf{G}^v_\tau=
(\mathbf{T}^{v_{\mathrm{ref}}}_{t_0})^{-1}\mathbf{T}^v_\tau$
be the camera-to-reference transform, with rotation
$\mathbf{R}^v_\tau$ and translation $\mathbf{b}^v_\tau$.
The corresponding 3D estimate is
\begin{equation}
\widehat{\mathbf{P}}^{\,v}_{\tau,n}
=
\mathbf{R}^v_\tau
\left(
\widehat{d}^v_{\tau,n}
(\mathbf{K}^v)^{-1}
\begin{bmatrix}
\widehat{\mathbf{u}}^v_{\tau,n}\\
1
\end{bmatrix}
\right)
+
\mathbf{b}^v_\tau.
\label{eq:unproject}
\end{equation}

During training, future-frame features are obtained from
partially denoised tokens rather than clean future observations.
The head therefore learns to infer motion from the model's
predictive features. 

\noindent\textbf{Trajectory supervision.}
Extending the multi-view tracking
objective~\citep{lee2026mvtrack4gen}, we supervise the
iterative predictions with
\begin{equation}
\mathcal{L}_{\mathrm{track}}
=
\lambda_{\mathrm{seq}}\mathcal{L}_{\mathrm{seq}}
+
\lambda_{\mathrm{depth}}\mathcal{L}_{\mathrm{depth}}
+
\lambda_{\mathrm{conf}}\mathcal{L}_{\mathrm{conf}}.
\label{eq:track}
\end{equation}
$\mathcal{L}_{\mathrm{seq}}$ and $\mathcal{L}_{\mathrm{depth}}$
apply a truncated Huber loss to 2D positions with their
frame-to-frame displacements and to log-depth, respectively;
truncation limits the influence of large errors from unstable
refinements on heavily noised features.
$\mathcal{L}_{\mathrm{conf}}$ is a binary cross-entropy whose
target indicates whether the predicted position lies within a
threshold of the ground truth. All terms are summed over the $M$
refinement iterations and averaged over visible points. The 2D
and depth targets are projected from the same ground-truth 3D
trajectories using known camera parameters, keeping supervision
geometrically consistent across views.

\noindent\textbf{Cross-view correspondence supervision.}
To supervise the features used for trajectory refinement
directly, we treat correspondence as classification over
the spatial cells of a target frame.
Let $\mathbf{C}^{v_1,v_2}_{i,j}$ denote the attention map
between latent frame $i$ in view $v_1$ and frame $j$ in
view $v_2$, obtained by normalizing query--key similarities
over target cells and averaging across attention heads.
We minimize
\begin{equation}
\mathcal{L}_{\mathrm{corr}}
=
-\frac{1}{|\Omega|}
\sum_{(n,v_2,j)\in\Omega}
\log
\mathbf{C}^{v_n,v_2}_{t_n,j}
\left[
\mathbf{u}_n,\mathbf{u}^{v_2}_{j,n}
\right],
\label{eq:corr}
\end{equation}
where $\Omega$ contains query--target pairs for which the
ground-truth correspondence is visible, and the bracket
selects the attention weight at that correspondence.
This objective encourages the generator's attention features
to associate the same physical point across views and time,
complementing supervision of the head's predicted positions.

\subsection{Training and Inference}
\label{sec:pipeline}

\noindent\textbf{Training.}
We first adapt the generation pathway, camera embeddings, and
view frequencies using $\mathcal{L}_{\mathrm{FM}}$ on all
multi-view sources, including those without trajectory annotations.
Next, we train only the trajectory head and view frequencies
with $\mathcal{L}_{\mathrm{track}}$, freezing the remaining
parameters. Finally, we jointly optimize the video backbone
and trajectory head with
\begin{equation}
\mathcal{L}_{\mathrm{forecast}}
=
\mathcal{L}_{\mathrm{FM}}
+
\lambda_{\mathrm{track}}\mathcal{L}_{\mathrm{track}}
+
\lambda_{\mathrm{corr}}\mathcal{L}_{\mathrm{corr}}.
\label{eq:forecast}
\end{equation}

\noindent\textbf{Inference.}
Conditioned on the observed histories, camera parameters, and
instruction, we denoise future-video tokens from Gaussian noise
at $\sigma=1$ to $0$, keeping history tokens clean.
Trajectories are decoded from the predicted clip's backbone
features and transformed into the reference frame.

\section{Dataset and Benchmark}
\label{sec:data}

We construct \textbf{\sname{}-Data}, 10K training clips pairing synchronized
multi-view videos with persistent 3D trajectories for tabletop manipulation
and urban dynamics, and \textbf{\bname{}}, an evaluation benchmark spanning
tabletop, indoor, and urban scenes in simulation and the real world
(\Cref{tab:data_bench}).

\subsection{Training Dataset}
\label{sec:training_data}

\noindent\textbf{Generation and capture.}
Tabletop clips replay physics-simulated pick-and-place episodes
of a Fourier GR-1~T2 robot in Unreal Engine~5, with randomized
clutter, surface appearance, scene placement, and camera poses.
Distractor interactions are validated and recorded before rendering.
Urban clips capture pedestrians and vehicles from a running
traffic and crowd simulation, varying their appearance,
camera intrinsics, and camera paths. In both domains, images
and geometric annotations are exported from the same simulation
timestep to maintain alignment across views.

\noindent\textbf{Annotations and validation.}
We annotate 256 surface points per object and 512 additional
points on the robot's arms and hands. Rigid transforms or
triangle identities and barycentric coordinates propagate
these points through motion and occlusion.
Each clip also provides metric depth, camera parameters,
point-validity and per-view visibility masks, object visibility
states, and a vision-language-model-generated caption.
Clips undergo checks for geometric alignment, annotation
completeness, depth validity, camera motion, target visibility,
and caption consistency, followed by a final human audit. Generation and validation details
are provided in Appendix~\ref{app:data}.

\subsection{\bname}
\label{sec:bench}

\begin{table}[t]
\centering
\setlength{\tabcolsep}{3pt}
\caption{\textbf{Overview of \sname{}-Data and \bname.}
}\vspace{-10pt}
\label{tab:data_bench}
\label{tab:data}
\label{tab:bench_sim}
\begin{minipage}[c]{0.49\linewidth}
\centering
\resizebox{\linewidth}{!}{%
\begin{tabular}{lrrrrrr}
\toprule
Domain & Samples & Views & fps & Frames & 3D tracks/clip & Hard \\
\midrule
\multicolumn{7}{l}{\textbf{\sname{}-Data}} \\
Tabletop manipulation & 8,000 & 3 & 20 & 81--181 & 768--2,048 & - \\
Urban dynamics & 2,000 & 4 & 15 & 81--181 & 256 & - \\
\midrule
\multicolumn{7}{l}{\textbf{\bname (Sim)}} \\
Tabletop manipulation & 500 & 3 & 20 & 81--680 & 768--1,536 & 27 \\
Urban dynamics & 500 & 4 & 15 & 81--181 & 256 & 95 \\
\midrule
\multicolumn{7}{l}{\textbf{\bname (Real)}} \\
Tabletop manipulation & 100 & 2 & 15 & 139--325 & 768 & - \\
Indoor scenes & 100 & 48 & 15 & 150 & 512 & - \\
\bottomrule
\end{tabular}
}
\end{minipage}
\hfill
\begin{minipage}[c]{0.50\linewidth}
\centering
\includegraphics[width=\linewidth]{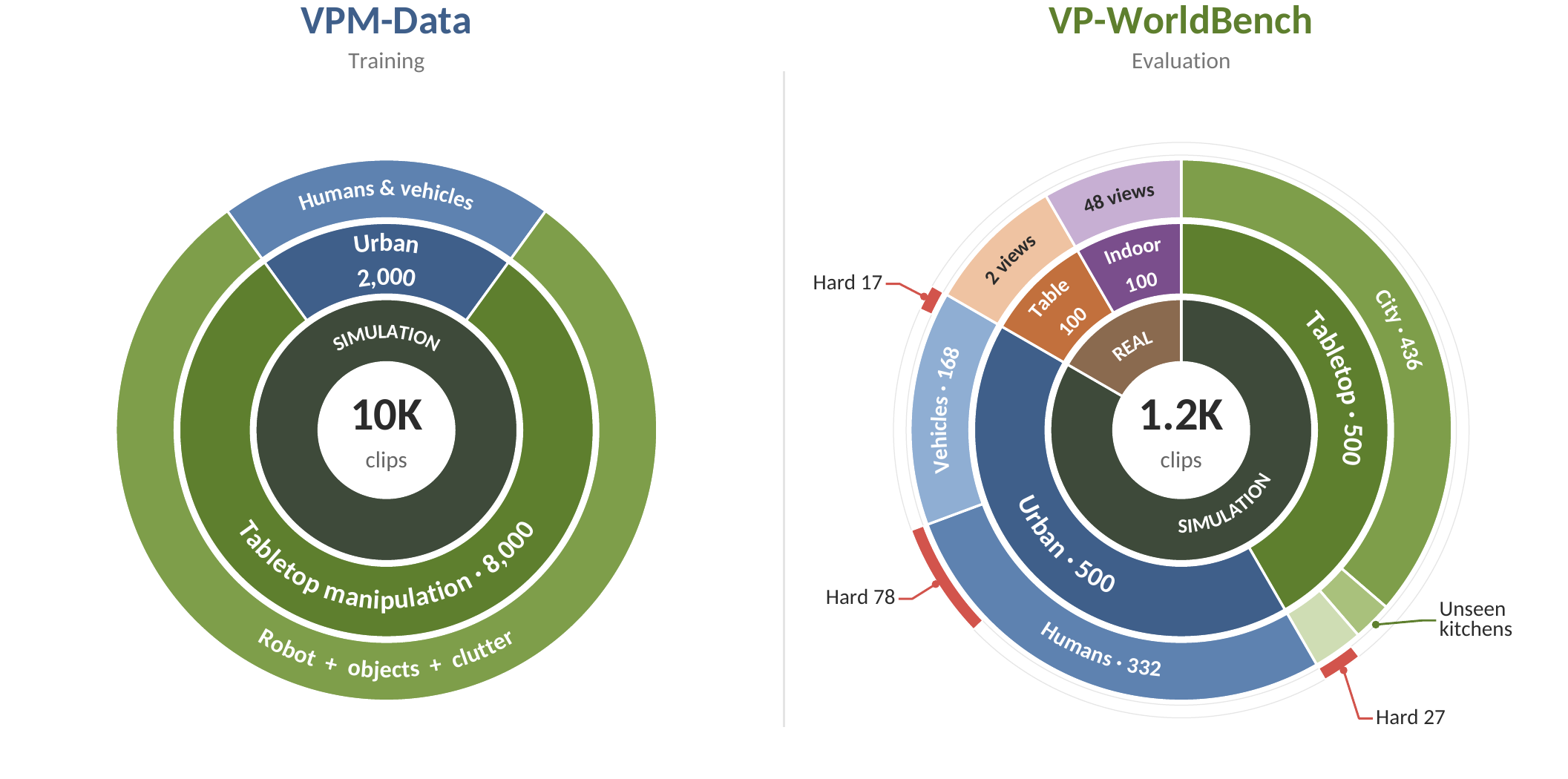}
\end{minipage}
\vspace{-15pt}
\end{table}
\begin{figure}[t]
    \centering
    \label{qual:bench}\centerline{\includegraphics[width=\textwidth]{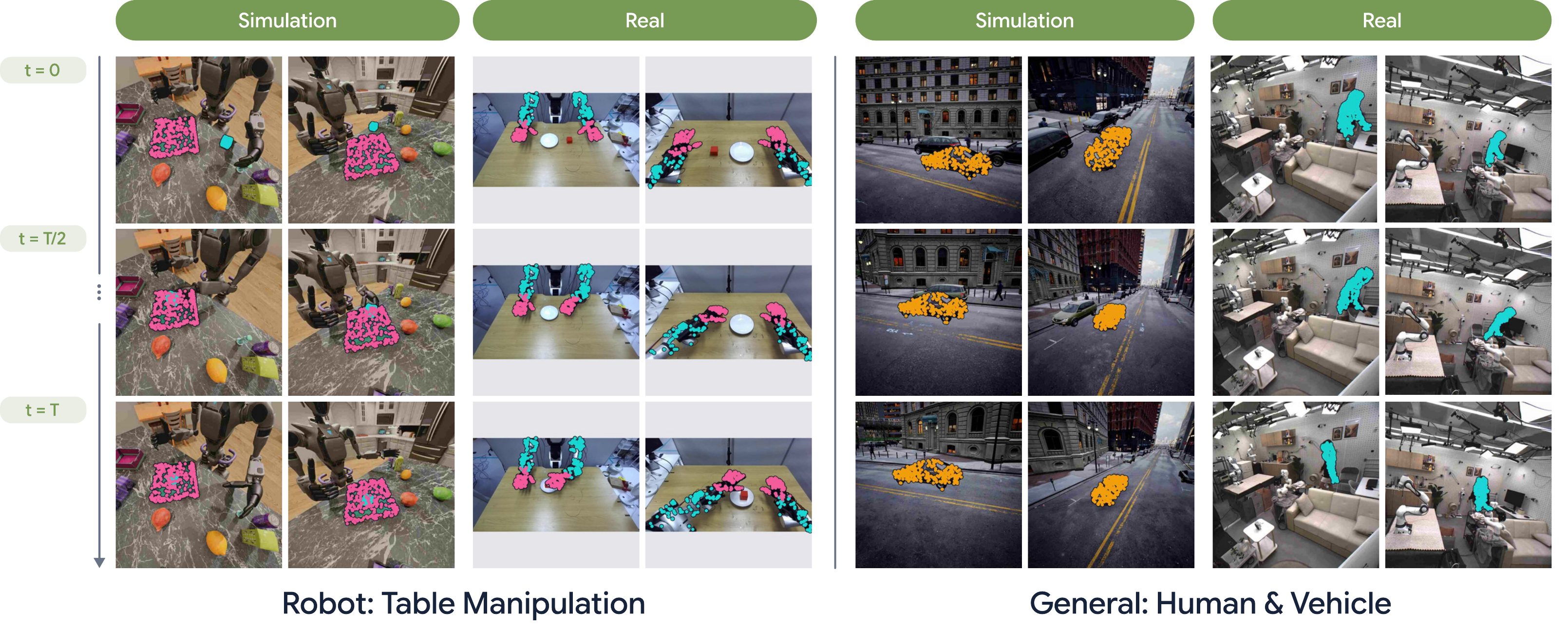}}
    \vspace{-0.05in}
   \caption{\textbf{Qualitative examples of \sname{}-Data and \bname{}.}}\vspace{-10pt}
\end{figure}

\noindent\textbf{Evaluation splits.}
The simulation benchmark covers three generalization regimes: 436 new
tabletop episodes at varied locations in city $C_1$, whose scenery appears
in \sname{}-Data only as static background; 64 tabletop episodes in two
unseen kitchens; and urban clips of moving agents in $C_1$, whose traffic and
crowd dynamics are unseen, as urban training uses a separate city $C_2$.
Each domain includes a hard subset in which individual cameras lose sight of
the target for at least 30 consecutive frames while at least one camera sees
it at every frame. In tabletop hard clips, all three cameras meet this
threshold and the target moves while hidden; in urban hard clips, all four
cameras lose and re-acquire the target, with two or three meeting the
threshold. Details are in Appendix~\ref{app:bench}.

\noindent\textbf{Real-world benchmarks.}
For robot manipulation, we collect 100 teleoperated episodes and sample 3D
point tracks on the robot surface from recorded robot states
(Appendix~\ref{app:robotpipeline}). For general scenes, we record 100
ten-second sequences of people moving, handling objects, and interacting with
furniture in a furnished home-like space with 48 synchronized RGB cameras.
Each sequence pairs multi-view videos with a 4D reconstruction for evaluating
visual forecasting and geometric accuracy, and passes the same quality checks
as the training data. Ground-truth 3D tracks are obtained by tracking points
in 2D across all views, triangulating them per frame, and keeping only tracks
that are depth-consistent across views.
Details are in
Appendix~\ref{app:real_bench}.

\section{Extension: Robot Control and Joint Action Learning}
\label{sec:robot}

To test whether point supervision also benefits control, we extend
\sname{} with robot action generation. The action-enabled model keeps
the multi-view video and trajectory objectives and additionally learns
robot commands from action-annotated interactions. Following Cosmos~3, we represent a future
action sequence $\mathbf{A}_t\in\mathbb{R}^{h_A\times d_A}$, with
horizon $h_A$ and dimension $d_A$, as an additional token modality
that shares the predictive backbone with video tokens. Action learning
can thus draw on representations trained with point-motion supervision.

Clean action tokens $\mathbf{a}_0$ are corrupted as
$\mathbf{a}_{\sigma_a}=(1-\sigma_a)\mathbf{a}_0+\sigma_a\boldsymbol{\epsilon}_a$,
where $\boldsymbol{\epsilon}_a$ is Gaussian noise and $\sigma_a$ is
sampled independently of the video noise level. The shared denoiser's
action output $v^a_\theta$ is trained jointly with the forecasting
objective in the final training stage:
\begin{equation}
\mathcal{L}_{\mathrm{robot}}
=\mathcal{L}_{\mathrm{forecast}}
+\lambda_{\mathrm{act}}
\underbrace{\mathbb{E}\left[\left\|
v^a_\theta(\mathbf{a}_{\sigma_a},\sigma_a,\mathbf{x}_\sigma,c)
-(\boldsymbol{\epsilon}_a-\mathbf{a}_0)
\right\|_2^2\right]}_{\mathcal{L}_{\mathrm{act}}}.
\label{eq:robot}
\end{equation}
At inference, video and action tokens are initialized from noise and
denoised jointly, conditioned only on the observations and instruction.
\section{Experiments}
We validate the effectiveness of \sname{} and its components by investigating the following questions:

\begin{itemize}[leftmargin=*,itemsep=0mm]
    \item Can \sname{} improve the generation quality of multi-view videos? (Table~\ref{tab:vp-worldbench},~\ref{tab:4d-recamandcmu},~\ref{tab:4d-robocasakitchen}) 
    \item Does point supervision benefit robot control? (Table~\ref{tab:robot_manipulation})
    \item Can explicitly learning future point motion improve world models? (Figure~\ref{fig:gen_qual}, Table~\ref{tab:analysis_modality})
\end{itemize}

\subsection{Experimental Setup}
\begin{table*}[t]
\centering
\caption{\textbf{Results on \bname{}.} Numbers in parentheses denote clips
per split. V and T denote video and point-trajectory supervision. }
\label{tab:vp-worldbench}
\vspace{-0.1in}
\setlength{\tabcolsep}{3pt}
\resizebox{\linewidth}{!}{%
\begin{tabular}{lccccccccccccccccc}
\toprule
& & \multicolumn{8}{c}{Simulation} & \multicolumn{8}{c}{Real-world} \\
\cmidrule(lr){3-10} \cmidrule(lr){11-18}
& & \multicolumn{4}{c}{Robot (500)} & \multicolumn{4}{c}{General (500)}
  & \multicolumn{4}{c}{Robot (100)} & \multicolumn{4}{c}{General (100)} \\
\cmidrule(lr){3-6} \cmidrule(lr){7-10} \cmidrule(lr){11-14} \cmidrule(lr){15-18}
Method & MV
& FVD $\downarrow$ & PSNR $\uparrow$ & MEt3R $\downarrow$ & $\delta^{\text{3D}}_{\text{avg}}$ $\uparrow$
& FVD $\downarrow$ & PSNR $\uparrow$ & MEt3R $\downarrow$ & $\delta^{\text{3D}}_{\text{avg}}$ $\uparrow$
& FVD $\downarrow$ & PSNR $\uparrow$ & MEt3R $\downarrow$ & $\delta^{\text{3D}}_{\text{avg}}$ $\uparrow$
& FVD $\downarrow$ & PSNR $\uparrow$ & MEt3R $\downarrow$ & $\delta^{\text{3D}}_{\text{avg}}$ $\uparrow$ \\
\midrule
HV-1.5 & \xmark
& 98.34 & 20.21 & 0.023 & 41.62 & 115.63 & 11.45 & 0.061 & 23.10
& 426.76 & 16.04 & 0.102 & 36.17 & 105.83 & 20.00 & 0.014 & 21.35 \\
Wan2.2-5B & \xmark
& 315.13 & 15.84 & 0.108 & 29.51 & 205.73 & 10.43 & 0.076 & 18.00
& 550.55 & 14.58 & 0.159 & 29.65 & 115.52 & 16.97 & 0.030 & 16.82 \\
X-WAM & \cmark
& 203.83 & 19.97 & 0.038 & 38.80 & - & - & - & -
& 492.31 & 17.63 & 0.091 & \underline{50.86} & - & - & - & - \\
Cosmos3-N & \xmark
& 104.74 & 20.05 & 0.040 & 39.03 & 167.02 & 10.99 & 0.062 & 22.12
& 326.11 & 16.37 & 0.099 & 41.50 & 66.24 & 20.89 & 0.015 & 20.47 \\
\midrule
\textbf{\sname{} (V)} & \cmark
& \underline{78.52} & \underline{21.41} & \underline{0.022} & \underline{43.62} & \underline{101.37} & \underline{16.95} & \underline{0.057} & \underline{28.71}
& \underline{296.83} & \underline{18.42} & \underline{0.074} & 50.83 & \underline{62.85} & \underline{20.96} & \underline{0.010} & \underline{26.54} \\
\textbf{\sname{} (V+T)} & \cmark
& \textbf{67.35} & \textbf{22.08} & \textbf{0.016} & \textbf{50.05} & \textbf{81.92} & \textbf{19.16} & \textbf{0.050} & \textbf{36.80}
& \textbf{255.19} & \textbf{20.57} & \textbf{0.064} & \textbf{64.75} & \textbf{56.48} & \textbf{22.10} & \textbf{0.007} & \textbf{33.92} \\
\bottomrule
\end{tabular}
}
\vspace{-10pt}
\end{table*}
\begin{table*}[t]
\centering
\begin{minipage}[t]{0.53\linewidth}
\centering
\caption{\textbf{Results of 4D generation on ReCamMaster and CMU Panoptic.}}
\label{tab:4d-recamandcmu}
\vspace{-0.07in}
\label{tab:recam_panoptic}
\setlength{\tabcolsep}{3pt}
\resizebox{\linewidth}{!}{%
\begin{tabular}{lccccccc}
\toprule
& & \multicolumn{2}{c}{ReCamMaster (100)} & \multicolumn{4}{c}{CMU Panoptic (100)} \\
\cmidrule(lr){3-4} \cmidrule(lr){5-8}
Method & MV
& FVD $\downarrow$ & PSNR $\uparrow$
& FVD $\downarrow$ & PSNR $\uparrow$ & MEt3R $\downarrow$ & $\delta^{\text{3D}}_{\text{avg}}$ $\uparrow$ \\
\midrule
HV-1.5 & \xmark
& 183.48 & 14.03 & 352.15 & \underline{21.60} & 0.041 & 31.67 \\
Wan2.2-5B & \xmark
& 457.16 & 12.10
& 258.06 & 18.65 & 0.061 & 27.88 \\
Cosmos3-N & \xmark
& 125.70 & 13.99
& 162.36 & 20.72 & 0.060 & 30.12 \\
\midrule
\textbf{\sname (V)} & \cmark
& \underline{103.12} & \underline{15.09}
& \underline{150.63} & 20.85 & \underline{0.037} & \underline{34.72} \\
\textbf{\sname (V+T)} & \cmark
& \textbf{79.63} & \textbf{17.06}
& \textbf{128.64} & \textbf{23.24} & \textbf{0.029} & \textbf{42.03} \\
\bottomrule
\end{tabular}
}
\end{minipage}
\hfill
\begin{minipage}[t]{0.45\linewidth}
\centering
\caption{\textbf{Results of 4D generation on RoboCasa Kitchen.}}
\label{tab:4d-robocasakitchen}
\vspace{-0.07in}
\label{tab:4d_recon}
\setlength{\tabcolsep}{3pt}
\resizebox{\linewidth}{!}{%
\begin{tabular}{lcccccc}
\toprule
 && \multicolumn{3}{c}{RGB} & \multicolumn{2}{c}{Geometry} \\
\cmidrule(lr){3-5} \cmidrule(lr){6-7}
Method & MV & PSNR$\uparrow$ & SSIM$\uparrow$ & LPIPS$\downarrow$ & MEt3R$\downarrow$ & $\delta^{\text{3D}}_{\text{avg}}\uparrow$ \\
\midrule
DreamZero & \xmark & 20.47 & 0.7612 & 0.1651 & 0.061 & 30.83 \\
Robot4DGen & \cmark & 22.05 & 0.8035 & 0.1078 & \underline{0.025} & \underline{46.96} \\
X-WAM & \cmark & 22.41 & 0.8729 & 0.0571 & 0.042 & 42.10 \\
\midrule
\textbf{\sname (V)} & \cmark & \underline{22.61} & \underline{0.8752} & \underline{0.0558} & 0.033 & 46.94 \\
\textbf{\sname (V+T)} & \cmark & \textbf{22.83} & \textbf{0.8801} & \textbf{0.0537} & \textbf{0.020} & \textbf{53.70} \\
\bottomrule
\end{tabular}
}
\end{minipage}\vspace{-10pt}
\end{table*}

\noindent{\bf Evaluation setup.} We evaluate \sname{} on three aspects: the
fidelity of generated videos, their geometric consistency, and the
performance of robot policies built on the action-enabled model. For a fair
comparison, no method, including \sname{}, receives camera parameters for
future frames;  thus, all methods must infer camera motion from the observed
history. Additional results with camera conditioning can be found in the Appendix. For video fidelity, we report FVD and PSNR, which measure
distribution-level and pixel-level fidelity, respectively. For geometric
consistency, we use cross-view MEt3R~\citep{asim2025met3r} and  extract 3D point tracks from the generated videos with
MVTracker~\citep{rajivc2025mvtracker} and compare them with ground-truth
trajectories, reporting the average fraction of
points within depth-adaptive thresholds ($\delta^{\text{3D}}_{\text{avg}}$).
For robot manipulation, we report task success
rates (\%). Results for the trajectory head's own forecasts are provided in
Appendix~\ref{app:own_head}.

\begin{table}[t]
\centering
\caption{\textbf{Robot manipulation results.}
 A denotes action. $\triangle$ denotes methods that observe multiple cameras without camera
conditioning, either as separate input images or tiled into a single frame. \cmark~denotes camera-conditioned
multi-view generation.
}\vspace{-5pt}
\label{tab:robot_manipulation}
\setlength{\tabcolsep}{6pt}
\resizebox{0.85\linewidth}{!}{%
\begin{tabular}{lclcccccc}
\toprule
& & & \multicolumn{3}{c}{Simulation} & \multicolumn{3}{c}{Real-world} \\
\cmidrule(lr){4-6} \cmidrule(lr){7-9}
Method & MV & Backbone
& LIBERO & LIBERO-Plus & RoboCasa
& \textit{PnP Banana} & \textit{PnP Cup} & \textit{PnP Cube} \\
\midrule
\multicolumn{9}{c}{\textit{VLA-based Methods}}\\
$\pi_{0.5}$      & $\triangle$ & - & 96.8 & 68.0 & 62.1 & 41.7 & 47.9 & 12.5 \\
GR00T N1.5       & $\triangle$ & - & 86.6 & 66.4 & 65.6 & - & - & - \\
GR00T N1.7       & $\triangle$ & - & - & - & 70.0 & 31.3 & 27.1 & 4.2 \\
\midrule
\multicolumn{9}{c}{\textit{WAM-based Methods}}\\
Cosmos3-Nano     & $\triangle$ & -            & 96.8 & 81.6 & 72.0 & 50.0 & 54.2 & 41.7 \\
DreamZero        & $\triangle$ & Wan2.1-14B   & - & - & - & 54.2 & 68.8 & 45.8 \\
X-WAM            & \cmark & Wan2.2-5B    & - & - & 73.8 & 14.6 & 27.1 & 4.2 \\
\midrule
\textbf{\sname (V+A)} & \cmark & Cosmos3-Nano & 96.8 & 82.0 & 73.4 & 58.3 & 64.6 & 43.8 \\
\textbf{\sname (V+T+A)} & \cmark & Cosmos3-Nano & \textbf{97.2} & \textbf{84.6} & \textbf{76.2} & \textbf{79.2} & \textbf{70.8} & \textbf{54.2} \\
\bottomrule
\end{tabular}
}
\vspace{-10pt}
\end{table}
% \vspace{0.05in}
\noindent{\bf Baselines.}
We compare \sname{} with recent video world models, which are fine-tuned on the same data as ours for fair comparison, and world-action models:
HunyuanVideo-1.5~\citep{wu2025hunyuanvideo}, Wan2.2-5B~\citep{team2025wan},
Cosmos3-Nano~\citep{nvidia2026cosmos3},  DreamZero~\citep{ye2026dreamzero},
Robot4DGen~\citep{liu2026geometry}, and X-WAM~\citep{guo2026unified}. Single-view models generate each requested view independently.  We additionally include VLA models for real-world robot manipulation experiments:  $\pi_{0.5}$~\citep{pi05} and GR00T N-series~\citep{gr00tn1_2025}.

\subsection{\bname{} Results}
\noindent{\bf Setup.} We first evaluate \sname{} on \bname{}, which covers
the same domains as VPM-Data, robot manipulation and general scene
dynamics, using held-out episodes and environments along with real-world
recordings. Video fidelity metrics are computed for each generated view and
averaged across views. We compare two variants: \sname{} (V), trained only
on multi-view videos, and \sname{} (V+T), our default model, trained on both
multi-view videos and 3D point trajectories. Both variants use the same
training budget, so the comparison isolates the effect of point supervision.

\noindent{\bf Results.}
As shown in \Cref{tab:vp-worldbench}, \sname{} (V+T) achieves the best
performance on every split and metric, generating videos with higher
fidelity and stronger geometric consistency in both simulated and
real-world scenes. Even without point supervision, \sname{} (V) improves
over its Cosmos3-Nano backbone, showing the benefit of multi-view
processing. Adding point supervision under the same training budget further
improves both fidelity and 3D motion accuracy on all splits, supporting our
hypothesis that learning where scene points will move yields more faithful
multi-view forecasts.

\begin{table*}[t]
\centering
\caption{\textbf{Modality-wise ablation on \bname{}.}}\vspace{-5pt}
\label{tab:analysis_modality}
\setlength{\tabcolsep}{3pt}
\resizebox{\linewidth}{!}{%
\begin{tabular}{lcccccccccccccccc}
\toprule
& \multicolumn{8}{c}{Simulation} & \multicolumn{8}{c}{Real-world} \\
\cmidrule(lr){2-9} \cmidrule(lr){10-17}
& \multicolumn{4}{c}{Robot (500)} & \multicolumn{4}{c}{General (500)}
  & \multicolumn{4}{c}{Robot (100)} & \multicolumn{4}{c}{General (100)} \\
\cmidrule(lr){2-5} \cmidrule(lr){6-9} \cmidrule(lr){10-13} \cmidrule(lr){14-17}
Method
& FVD $\downarrow$ & PSNR $\uparrow$ & MEt3R $\downarrow$ & $\delta^{\text{3D}}_{\text{avg}}$ $\uparrow$
& FVD $\downarrow$ & PSNR $\uparrow$ & MEt3R $\downarrow$ & $\delta^{\text{3D}}_{\text{avg}}$ $\uparrow$
& FVD $\downarrow$ & PSNR $\uparrow$ & MEt3R $\downarrow$ & $\delta^{\text{3D}}_{\text{avg}}$ $\uparrow$
& FVD $\downarrow$ & PSNR $\uparrow$ & MEt3R $\downarrow$ & $\delta^{\text{3D}}_{\text{avg}}$ $\uparrow$ \\
\midrule
Cosmos3-Nano
& 104.74 & 20.05 & 0.040 & 39.03 & 167.02 & 10.99 & 0.062 & 22.12
& 326.11 & 16.37 & 0.099 & 41.50 & 66.24 & 20.89 & 0.015 & 20.47 \\
\midrule
\sname{} (V)
& 78.52 & 21.41 & 0.022 & 43.62 & 101.37 & 16.95 & 0.057 & 28.71
& 296.83 & 18.42 & 0.074 & 50.83 & 62.85 & 20.96 & 0.010 & 26.54 \\
\sname{} (V+A)
& 74.29 & 21.59 & 0.021 & 44.27 & 98.08 & 17.42 & 0.054 & 30.20
& 291.05 & 18.87 & 0.069 & 53.21 & 61.31 & 21.24 & 0.009 & 28.06 \\
\sname{} (V+T)
& 67.35 & 22.08 & 0.016 & 50.05 & 81.92 & 19.16 & 0.050 & 36.80
& 255.19 & 20.57 & 0.064 & 64.75 & 56.48 & 22.10 & 0.007 & 33.92 \\
\sname{} (V+T+A)
& \textbf{58.97} & \textbf{22.43} & \textbf{0.015} & \textbf{53.14} & \textbf{71.23} & \textbf{20.02} & \textbf{0.048} & \textbf{38.57}
& \textbf{234.59} & \textbf{21.10} & \textbf{0.058} & \textbf{66.94} & \textbf{55.48} & \textbf{22.85} & \textbf{0.006} & \textbf{37.03} \\
\bottomrule
\end{tabular}
}
\vspace{-10pt}
\end{table*}

\subsection{Zero-Shot 4D Generation Results}
\label{exp:zero_shot_4d_gen}
\noindent{\bf Setup.}
To assess out-of-domain generalization, we further evaluate  on three datasets: ReCamMaster~\citep{bai2025recammaster}, CMU
Panoptic~\citep{joo2015panoptic}, and RoboCasa
Kitchen~\citep{nasiriany2026robocasa365}. 

\noindent{\bf Results.}
As shown in \Cref{tab:4d-recamandcmu}, \sname{} (V+T) achieves the best
results on both ReCamMaster and CMU Panoptic. Compared with \sname{} (V),
point supervision improves both video fidelity and 3D motion accuracy,
indicating that the generated videos depict more coherent 3D motion. The same
trend holds on RoboCasa Kitchen (\Cref{tab:4d-robocasakitchen}), where
\sname{} (V+T) outperforms all baselines. Notably, \sname{} (V) trails the geometry-aware Robot4DGen on multi-view consistency, whereas \sname{} (V+T) surpasses it, suggesting
that point supervision is key to the 3D consistency of generated videos.

\subsection{Robot Manipulation Results}
\noindent{\bf Setup.}
We test whether point supervision also benefits robot control
using the action-enabled extension of \sname{} (\Cref{sec:robot}).
Simulation experiments follow each benchmark's standard protocol
with three synchronized views: one wrist camera and one or two
third-person cameras on either side of the robot.
In the real world, we use the DROID platform (a Franka Research 3 arm
with a Robotiq 2F-85 gripper) and train only on an 89K-trajectory subset of DROID dataset.
We evaluate three pick-and-place tasks of varying difficulty,
with 24 trials each. We provide more details in Appendix~\ref{app:robot_task}.

\noindent{\bf Results.}
As shown in \Cref{tab:robot_manipulation}, \sname{} (V+T+A) achieves the best
success rates on all simulation benchmarks, outperforming both VLA policies
and world-action models. This advantage
carries over to the real world (\Cref{tab:robot_manipulation}), where \sname{} (V+T+A)
achieves the highest success rate on every task. In both simulated and real-world settings, \sname{}
(V+T+A) consistently improves over \sname{} (V+A), which is trained without
point supervision. These results indicate that forecasting persistent 3D
point trajectories provides a useful learning signal for robot actions.

\begin{figure}[t]
    \centering
    \centerline{\includegraphics[width=\textwidth]{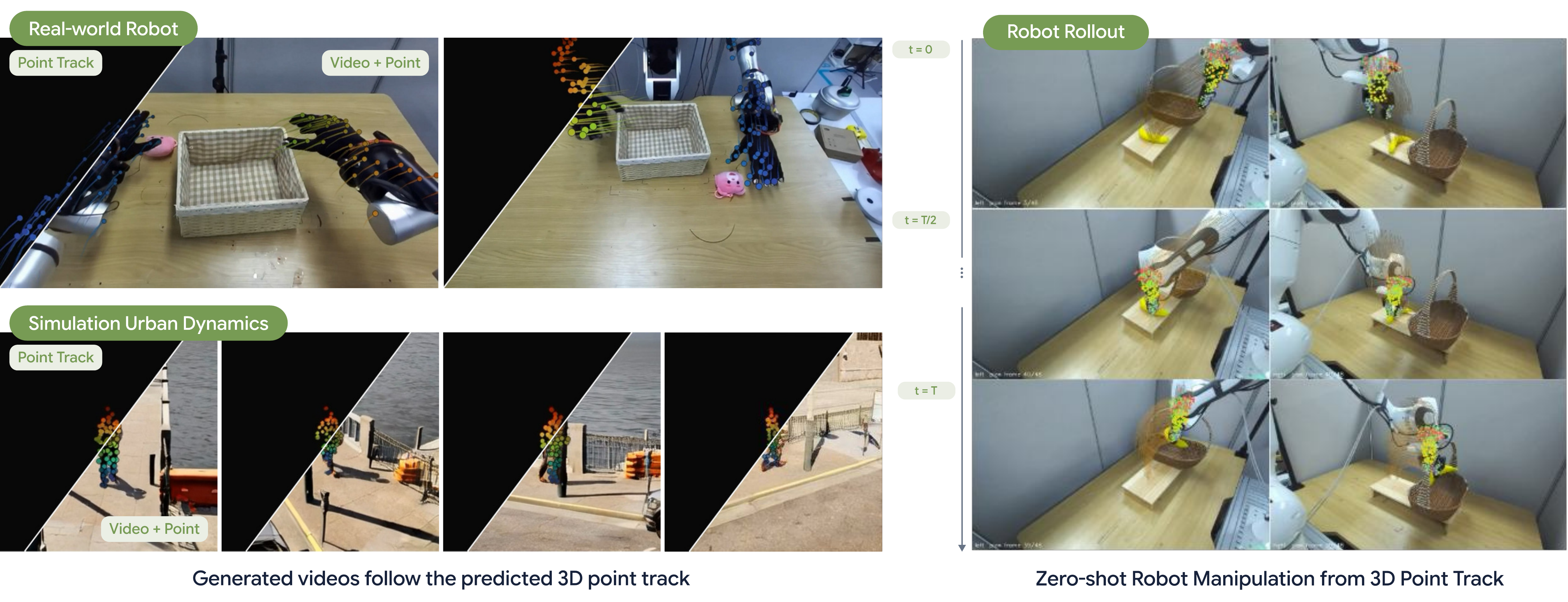}}
    \vspace{-0.05in}
  \caption{\textbf{Predicted 3D point trajectories guide generation and
action.} \textbf{Left:} Predicted trajectories alone and overlaid on
generated videos.
\textbf{Right:} Zero-shot real-robot rollout obtained by converting predicted
trajectories into commands with inverse kinematics.}
   \label{fig:gen_qual}
   \vspace{-15pt}
\end{figure}
\subsection{Analysis and Ablation Studies}

\noindent{\bf Modality-wise ablation.}
We train \sname{} with combinations of video (V), point trajectories (T), and
actions (A). In \Cref{tab:analysis_modality}, adding trajectories yields the
largest gains in video fidelity and 3D motion accuracy on every split, while
adding actions yields modest improvements. Training with all three performs
best on all splits and metrics, indicating that action supervision
complements point supervision. Together with \Cref{tab:robot_manipulation},
where point supervision raises success rates on every benchmark, these
results show point trajectories benefit both generation and action. We next
examine why.

\noindent{\bf Why 3D point trajectory forecasting helps multi-view video generation.}
The trajectory head predicts point motion from the generator's intermediate
 features, which the remaining blocks process into future
frames, so point supervision directly shapes the representation from which
videos are generated. \Cref{fig:gen_qual} illustrates this coupling: the
robot hands and the pedestrian
closely follow the predicted trajectories overlaid on the generated frames.
Since each trajectory follows a single physical point, generation that
follows them remains consistent across views.

\noindent{\bf Why 3D point trajectory forecasting helps action prediction.}
3D point forecasting supervision consistently improves the action-enabled model
(\Cref{tab:robot_manipulation}), which we attribute to three properties.
First, on robot data, trajectories are computed from joint states by forward
kinematics, describing the same motion as actions but grounded in 3D and in
every view, which teaches the backbone to connect visual features with where
the robot will move. Second, they are dense, constraining the robot's full
future configuration with hundreds of points per timestep rather than one
low-dimensional action. Third, object trajectories, where available, specify
where manipulated objects should go, encoding the task's intended outcome.
Action tokens, sharing this backbone, thus draw on a representation that
already encodes the robot's future motion. Supporting this, converting
predicted trajectories into commands via inverse kinematics, without action
tokens, completes unseen real-robot tasks zero-shot (\Cref{fig:gen_qual}).
Though not yet consistent enough for quantitative evaluation, these rollouts
show that forecast trajectories alone carry actionable information
(Appendix~\ref{app:robot_manip}).
\section{Conclusion and Limitations}
We presented \sname{}, a multi-view world model that jointly forecasts future
videos and persistent 3D point trajectories from posed video histories and
language instructions. With \sname{}-Data and \bname{}, our experiments show
that supervising where scene points will move improves visual forecasting on
simulated and real benchmarks and carries over to robot action learning.
Limitations remain. Our point supervision covers only the
robot and annotated objects rather than the full scene; skeletal annotations
also omit independent cloth and hair motion. \sname{} assumes known camera
intrinsics and poses at inference, and its cost grows with the number of
views, as all camera streams share one token sequence. Future work includes
deriving trajectory supervision from real multi-view videos at scale,
extending it to denser, non-rigid content, and relaxing the need for
calibrated cameras.
\section*{Acknowledgments}
This work was partly supported by Institute for Information \& Communications Technology Planning \& Evaluation (IITP) grant funded by the Korea government (MSIT) (RS-2019-II190075, Artificial Intelligence Graduate School Program (KAIST); RS-2024-00509279, Global AI Frontier Lab), RLWRLD Inc., and Physics Sim Lab, Inc.; the Swiss AI Initiative through a grant from the Swiss National Supercomputing Centre (CSCS) under project ID ab036 on the Alps infrastructure; the Swiss National Science Foundation (Advanced Grant 216260: ``Beyond Frozen Worlds: Capturing Functional 3D Digital Twins from the Real World''); the European Union’s Horizon Europe research and innovation programme under grant agreement No. 101214398 (ELLIOT); the Horizon Europe programme under grant agreement No. 101298421 (GRAIL); and the ETH AI Center through an ETH AI Center postdoc fellowship awarded to Sunghwan Hong.

The authors thank Seonghu Jeon and Seoyeon Byun for their proofreading and assistance with improving the visualizations in the manuscript; Junbeom Kim, Minha Lee, Seong Hyeon Park, and Minsu Cho for their insightful discussions; and Beomjun Kim, Seunghoon Sim, Jaewon Youm, and Heeseung Kwon for providing feedback on model implementation and real-robot experiments.

\bibliography{iclr2027_conference}

@article{nvidia2026cosmos3,
  title={Cosmos 3: Omnimodal world models for physical ai},
  author={Agarwal, Niket and Ali, Arslan and Allen, Jon and Antolini, Martin and Aubame, Adeline and Azzolini, Alisson and Bai, Junjie and Bala, Maciej and Balaji, Yogesh and Bapst, Josh and others},
  journal={arXiv preprint arXiv:2606.02800},
  year={2026}
}

@article{han2026geometric,
  title={Geometric Action Model for Robot Policy Learning},
  author={Han, Jisang and Jeon, Seonghu and Jung, Jaewoo and Zurbr{\"u}gg, Ren{\'e} and An, Honggyu and Portela, Tifanny and Hutter, Marco and Pollefeys, Marc and Kim, Seungryong and Hong, Sunghwan},
  journal={arXiv preprint arXiv:2606.17046},
  year={2026}
}

@article{balice2026no,
  title={No Pose, No Problem in 4D: Feed-Forward Dynamic Gaussians from Unposed Multi-View Videos},
  author={Balice, Matteo and Kunzi, Yanik and Zhang, Chenyangguang and Matteucci, Matteo and Pollefeys, Marc and Hong, Sunghwan},
  journal={arXiv preprint arXiv:2605.22190},
  year={2026}
}

@inproceedings{lee20263d,
  title={3d scene prompting for scene-consistent camera-controllable video generation},
  author={Lee, JoungBin and Jung, Jaewoo and Han, Jisang and Narihira, Takuya and Fukuda, Kazumi and Seo, Junyoung and Hong, Sunghwan and Mitsufuji, Yuki and Kim, Seungryong},
  booktitle={International Conference on Learning Representations},
  volume={2026},
  pages={73753--73777},
  year={2026}
}

@article{li2026spatialvam,
  title={Spatial-Aware Multi-View Video Diffusion as a Data-Efficient Robot Policy},
  author={Li, Peiyan and Chen, Yixiang and Xu, Yuan and Yang, Jiabing and Wu, Xiangnan and Guo, Jun and Sun, Nan and Qian, Long and Li, Xinghang and Xiao, Xin and others},
  journal={arXiv preprint arXiv:2604.03181},
  year={2026}
}

@inproceedings{xie2025sv4d,
  title={Sv4d: Dynamic 3d content generation with multi-frame and multi-view consistency},
  author={Xie, Yiming and Yao, Chun-Han and Voleti, Vikram and Jiang, Huaizu and Jampani, Varun},
  booktitle={International Conference on Learning Representations},
  volume={2025},
  pages={33421--33441},
  year={2025}
}

@inproceedings{ravi2025sam,
  title={Sam 2: Segment anything in images and videos},
  author={Ravi, Nikhila and Gabeur, Valentin and Hu, Yuan-Ting and Hu, Ronghang and Ryali, Chaitanya and Ma, Tengyu and Khedr, Haitham and R{\"a}dle, Roman and Rolland, Chloe and Gustafson, Laura and others},
  booktitle={International Conference on Learning Representations},
  volume={2025},
  pages={28085--28128},
  year={2025}
}

@inproceedings{liu2026geometry,
  title={Geometry-aware 4d video generation for robot manipulation},
  author={Liu, Zeyi and Li, Shuang and Cousineau, Eric and Feng, Siyuan and Burchfiel, Benjamin and Song, Shuran},
  booktitle={International Conference on Learning Representations},
  volume={2026},
  pages={140751--140773},
  year={2026}
}

@article{zhang2026tapip3d,
  title={Tapip3d: Tracking any point in persistent 3d geometry},
  author={Zhang, Bowei and Ke, Lei and Harley, Adam and Fragkiadaki, Katerina},
  journal={Advances in Neural Information Processing Systems},
  volume={38},
  pages={135284--135303},
  year={2026}
}

@article{wu2025hunyuanvideo,
  title={Hunyuanvideo 1.5 technical report},
  author={Wu, Bing and Zou, Chang and Li, Changlin and Huang, Duojun and Yang, Fang and Tan, Hao and Peng, Jack and Wu, Jianbing and Xiong, Jiangfeng and Jiang, Jie and others},
  journal={arXiv preprint arXiv:2511.18870},
  year={2025}
}

@article{guo2026unified,
  title={Unified 4d world action modeling from video priors with asynchronous denoising},
  author={Guo, Jun and Li, Qiwei and Li, Peiyan and Chen, Zilong and Sun, Nan and Su, Yifei and Wang, Heyun and Zhang, Yuan and Li, Xinghang and Liu, Huaping},
  journal={arXiv preprint arXiv:2604.26694},
  year={2026}
}

@article{bharadhwaj2026motionforesight,
  title={MotionForesight: Re-purposing Video Models for Future 3D Scene-Flow Prediction},
  author={Bharadhwaj, Homanga and Jangir, Yash},
  journal={arXiv preprint arXiv:2607.16192},
  year={2026}
}

@article{zhang2026learning,
  title={Learning 4D Geometric Priors for Inference-Efficient World Action Models},
  author={Zhang, Jianjun and Zhu, Jian and Su, Taiyi and Ma, Chong and Huang, Zitai and Xu, Yi and Wang, Hanli},
  journal={arXiv preprint arXiv:2607.05468},
  year={2026}
}

@article{yang20264d,
  title={4D-WAM: Infusing Spatiotemporal Awareness into World Action Models through Trajectory Fields},
  author={Yang, Lishan and Song, Wenxuan and Wang, Xi and Sheng, Pingyue and Fang, Zheng and Zhou, Ziyang and He, Junjie and Yan, Haodong and Chen, Jiayi and Sun, Nan and others},
  journal={arXiv preprint arXiv:2608.08023},
  year={2026}
}

@article{zhou2026gem,
  title={Gem-4d: Geometry-enhanced video world models for robot manipulation},
  author={Zhou, Kaichen and Chen, Yuzhen and Zhan, Fangneng and Hua, Hang and Chen, Grace and Chang, Xinhai and Qu, Ao and Du, Yilun and Liu, Zhuang and Liang, Paul Pu and others},
  journal={arXiv preprint arXiv:2605.22882},
  year={2026}
}

@article{shen2026videovla,
  title={Videovla: Video generators can be generalizable robot manipulators},
  author={Shen, Yichao and Wei, Fangyun and Du, Zhiying and Liang, Yaobo and Lu, Yan and Yang, Jiaolong and Zheng, Nanning and Guo, Baining},
  journal={Advances in neural information processing systems},
  volume={38},
  pages={95597--95621},
  year={2026}
}

@article{team2025wan,
  title={Wan: Open and advanced large-scale video generative models},
  author={Team, Wan and Wang, Ang and Ai, Baole and Wen, Bin and Mao, Chaojie and others},
  journal={arXiv preprint arXiv:2503.20314},
  volume={2},
  number={3},
  pages={4},
  year={2025}
}

@inproceedings{karaev2025cotracker3,
  title={CoTracker3: Simpler and Better Point Tracking by Pseudo-Labeling Real Videos},
  author={Karaev, Nikita and Makarov, Yuri and Wang, Jianyuan and Neverova, Natalia and Vedaldi, Andrea and Rupprecht, Christian},
  booktitle={2025 IEEE/CVF International Conference on Computer Vision (ICCV)},
  pages={1--10},
  year={2025},
  organization={IEEE}
}

@article{koppula2026tapvidmv,
  title={TAPVid-MV: A Benchmark for Tracking Any Point in 3D Across Multiple Views},
  author={Koppula, Skanda and Rajic, Frano and Rahman, Abdullah Faiz Ur and Yang, Yi and Rocco, Ignacio and Thakwani, Jeet and Kabra, Rishabh and Zisserman, Andrew and Carreira, Joao and Tang, Siyu and others},
  journal={arXiv preprint arXiv:2609.01899},
  year={2026}
}

@article{tesch2026bedlam2,
  title={BEDLAM2. 0: Synthetic humans and cameras in motion},
  author={Tesch, Joachim and Becherini, Giorgio and Achar, Prerana and Yiannakidis, Anastasios and Kocabas, Muhammed and Patel, Priyanka and Black, Michael},
  journal={Advances in Neural Information Processing Systems},
  volume={38},
  year={2026}
}

@article{pi05,
  title={$\pi_{0.5}$: a Vision-Language-Action Model with Open-World Generalization},
  author={Intelligence, Physical and Black, Kevin and Brown, Noah and Darpinian, James and Dhabalia, Karan and Driess, Danny and Esmail, Adnan and Equi, Michael and Finn, Chelsea and Fusai, Niccolo and others},
  journal={arXiv preprint arXiv:2504.16054},
  year={2025}
}

@inproceedings{gr00tn1_2025,
  archivePrefix = {arxiv},
  eprint     = {2503.14734},
  title      = {{GR00T} {N1}: An Open Foundation Model for Generalist Humanoid Robots},
  author     = {NVIDIA and Johan Bjorck and Fernando Castañeda and Nikita Cherniadev and Xingye Da and Runyu Ding and Linxi "Jim" Fan and Yu Fang and Dieter Fox and Fengyuan Hu and Spencer Huang and Joel Jang and Zhenyu Jiang and Jan Kautz and Kaushil Kundalia and Lawrence Lao and Zhiqi Li and Zongyu Lin and Kevin Lin and Guilin Liu and Edith Llontop and Loic Magne and Ajay Mandlekar and Avnish Narayan and Soroush Nasiriany and Scott Reed and You Liang Tan and Guanzhi Wang and Zu Wang and Jing Wang and Qi Wang and Jiannan Xiang and Yuqi Xie and Yinzhen Xu and Zhenjia Xu and Seonghyeon Ye and Zhiding Yu and Ao Zhang and Hao Zhang and Yizhou Zhao and Ruijie Zheng and Yuke Zhu},
  month      = {March},
  year       = {2025},
  booktitle  = {ArXiv Preprint},
}

@inproceedings{wu2026geometry,
  title={Geometry forcing: Marrying video diffusion and 3d representation for consistent world modeling},
  author={Wu, Haoyu and Wu, Diankun and He, Tianyu and Guo, Junliang and Ye, Yang and Duan, Yueqi and Bian, Jiang},
  booktitle={International Conference on Learning Representations},
  volume={2026},
  pages={119186--119209},
  year={2026}
}

@article{zhang2505videorepa,
  title={Videorepa: Learning physics for video generation through relational alignment with foundation models (2025)},
  author={Zhang, X and Liao, J and Zhang, S and Meng, F and Wan, X and Yan, J and Cheng, Y},
  journal={arXiv preprint arXiv:2505.23656},
  year={2025}
}

@article{chefer2025videojam,
  title={Videojam: Joint appearance-motion representations for enhanced motion generation in video models},
  author={Chefer, Hila and Singer, Uriel and Zohar, Amit and Kirstain, Yuval and Polyak, Adam and Taigman, Yaniv and Wolf, Lior and Sheynin, Shelly},
  journal={arXiv preprint arXiv:2502.02492},
  year={2025}
}

@inproceedings{jeong2025track4gen,
  title={Track4gen: Teaching video diffusion models to track points improves video generation},
  author={Jeong, Hyeonho and Huang, Chun-Hao P and Ye, Jong Chul and Mitra, Niloy J and Ceylan, Duygu},
  booktitle={2025 IEEE/CVF Conference on Computer Vision and Pattern Recognition (CVPR)},
  pages={7276--7287},
  year={2025},
  organization={IEEE}
}

@inproceedings{joo2015panoptic,
  title={Panoptic studio: A massively multiview system for social motion capture},
  author={Joo, Hanbyul and Liu, Hao and Tan, Lei and Gui, Lin and Nabbe, Bart and Matthews, Iain and Kanade, Takeo and Nobuhara, Shohei and Sheikh, Yaser},
  booktitle={Proceedings of the IEEE international conference on computer vision},
  pages={3334--3342},
  year={2015}
}

@article{liu2022flow,
  title={Flow straight and fast: Learning to generate and transfer data with rectified flow},
  author={Liu, Xingchao and Gong, Chengyue and Liu, Qiang},
  journal={arXiv preprint arXiv:2209.03003},
  year={2022}
}

@inproceedings{bai2025recammaster,
  title={Recammaster: Camera-controlled generative rendering from a single video},
  author={Bai, Jianhong and Xia, Menghan and Fu, Xiao and Wang, Xintao and Mu, Lianrui and Cao, Jinwen and Liu, Zuozhu and Hu, Haoji and Bai, Xiang and Wan, Pengfei and others},
  booktitle={2025 IEEE/CVF International Conference on Computer Vision (ICCV)},
  pages={14834--14844},
  year={2025},
  organization={IEEE}
}

@article{lipman2022flow,
  title={Flow matching for generative modeling},
  author={Lipman, Yaron and Chen, Ricky TQ and Ben-Hamu, Heli and Nickel, Maximilian and Le, Matt},
  journal={arXiv preprint arXiv:2210.02747},
  year={2022}
}

@article{sitzmann2021light,
  title={Light field networks: Neural scene representations with single-evaluation rendering},
  author={Sitzmann, Vincent and Rezchikov, Semon and Freeman, Bill and Tenenbaum, Josh and Durand, Fredo},
  journal={Advances in neural information processing systems},
  volume={34},
  pages={19313--19325},
  year={2021}
}

@article{he2024cameractrl,
  title={Cameractrl: Enabling camera control for text-to-video generation},
  author={He, Hao and Xu, Yinghao and Guo, Yuwei and Wetzstein, Gordon and Dai, Bo and Li, Hongsheng and Yang, Ceyuan},
  journal={arXiv preprint arXiv:2404.02101},
  year={2024}
}

@inproceedings{park2026redirector,
  title={Redirector: Creating any-length video retakes with rotary camera encoding},
  author={Park, Byeongjun and Kim, Byung-Hoon and Chung, Hyungjin and Ye, Jong Chul},
  booktitle={Proceedings of the IEEE/CVF Conference on Computer Vision and Pattern Recognition},
  pages={11163--11173},
  year={2026}
}

@article{lee2026mvtrack4gen,
  title={MVTrack4Gen: Multi-View Point Tracking as Geometric Supervision for 4D Video Generation},
  author={Lee, JoungBin and Jung, Jaewoo and Lee, Jongmin and Kim, Tongmin and Kim, Hyunsung and Narihira, Takuya and Fukuda, Kazumi and Koo, Jahyeok and Han, Jisang and Mitsufuji, Yuki and others},
  journal={arXiv preprint arXiv:2606.26087},
  year={2026}
}

@inproceedings{koo2026mv,
  title={Mv-tap: Tracking any point in multi-view videos},
  author={Koo, Jahyeok and Kim, In{\`e}s Hyeonsu and Kim, Mungyeom and Park, Junghyun and Park, Seohyeon and Kim, Jaeyeong and Yi, Jung and Cho, Seokju and Kim, Seungryong},
  booktitle={Proceedings of the IEEE/CVF Conference on Computer Vision and Pattern Recognition},
  pages={20932--20941},
  year={2026}
}

@article{ye2026dreamzero,
  title={World action models are zero-shot policies},
  author={Ye, Seonghyeon and Ge, Yunhao and Zheng, Kaiyuan and Gao, Shenyuan and Yu, Sihyun and Kurian, George and Indupuru, Suneel and Tan, You Liang and Zhu, Chuning and Xiang, Jiannan and others},
  journal={arXiv preprint arXiv:2602.15922},
  year={2026}
}

@article{lin2025da3,
  title={Depth anything 3: Recovering the visual space from any views},
  author={Lin, Haotong and Chen, Sili and Liew, Junhao and Chen, Donny Y and Li, Zhenyu and Shi, Guang and Feng, Jiashi and Kang, Bingyi},
  journal={arXiv preprint arXiv:2511.10647},
  year={2025}
}

@inproceedings{grauman2024egoexo4d,
  title={Ego-exo4d: Understanding skilled human activity from first-and third-person perspectives},
  author={Grauman, Kristen and Westbury, Andrew and Torresani, Lorenzo and Kitani, Kris and Malik, Jitendra and Afouras, Triantafyllos and Ashutosh, Kumar and Baiyya, Vijay and Bansal, Siddhant and Boote, Bikram and others},
  booktitle={Proceedings of the IEEE/CVF conference on computer vision and pattern recognition},
  pages={19383--19400},
  year={2024}
}

@article{khazatsky2024droid,
  title={Droid: A large-scale in-the-wild robot manipulation dataset},
  author={Khazatsky, Alexander and Pertsch, Karl and Nair, Suraj and Balakrishna, Ashwin and Dasari, Sudeep and Karamcheti, Siddharth and Nasiriany, Soroush and Srirama, Mohan Kumar and Chen, Lawrence Yunliang and Ellis, Kirsty and others},
  journal={arXiv preprint arXiv:2403.12945},
  year={2024}
}

@inproceedings{nasiriany2026robocasa365,
  title={Robocasa365: A large-scale simulation framework for training and benchmarking generalist robots},
  author={Nasiriany, Soroush and Nasiriany, Sep and Maddukuri, Abhiram and Zhu, Yuke},
  booktitle={International Conference on Learning Representations},
  volume={2026},
  pages={98643--98667},
  year={2026}
}

@article{zhang2026molmomotion,
  title={MolmoMotion: Forecasting Point Trajectories in 3D with Language Instruction},
  author={Zhang, Jianing and Zheng, Chenhao and Yang, Yajun and Argus, Max and Soraki, Rustin and Han, Winson and Anderson, Taira and Li, Chun-Liang and Liu, Shuo and Duan, Jiafei and others},
  journal={arXiv preprint arXiv:2606.18558},
  year={2026}
}

@article{liang2024mot,
  title={Mixture-of-transformers: A sparse and scalable architecture for multi-modal foundation models},
  author={Liang, Weixin and Yu, Lili and Luo, Liang and Iyer, Srinivasan and Dong, Ning and Zhou, Chunting and Ghosh, Gargi and Lewis, Mike and Yih, Wen-tau and Zettlemoyer, Luke and others},
  journal={arXiv preprint arXiv:2411.04996},
  year={2024}
}

@inproceedings{rajivc2025mvtracker,
  title={Multi-view 3d point tracking},
  author={Raji{\v{c}}, Frano and Xu, Haofei and Mihajlovic, Marko and Li, Siyuan and Demir, Irem and G{\"u}ndo{\u{g}}du, Emircan and Ke, Lei and Prokudin, Sergey and Pollefeys, Marc and Tang, Siyu},
  booktitle={Proceedings of the IEEE/CVF International Conference on Computer Vision},
  pages={59--68},
  year={2025}
}

@inproceedings{zhou2019continuity,
  title={On the continuity of rotation representations in neural networks},
  author={Zhou, Yi and Barnes, Connelly and Lu, Jingwan and Yang, Jimei and Li, Hao},
  booktitle={2019 IEEE/CVF Conference on Computer Vision and Pattern Recognition (CVPR)},
  pages={5738--5746},
  year={2019},
  organization={IEEE}
}

@article{zhang2000a,
author = {Zhang, Zhengyou},
title = {A Flexible New Technique for Camera Calibration},
year = {2000},
month = {December},
pages = {1330-1334},
journal = {IEEE Transactions on Pattern Analysis and Machine Intelligence},
volume = {22},
note = {MSR-TR-98-71, Updated March 25, 1999},
}

@inproceedings{schoenberger2016sfm,
    author={Sch\"{o}nberger, Johannes Lutz and Frahm, Jan-Michael},
    title={Structure-from-Motion Revisited},
    booktitle={Conference on Computer Vision and Pattern Recognition (CVPR)},
    year={2016},
}

@inproceedings{li2026ambisur,
  title={Revisiting Photometric Ambiguity for Accurate Gaussian-Splatting Surface Reconstruction},
  author={Li, Jiahe and Zhang, Jiawei and Bai, Xiao and Zheng, Jin and Yu, Xiaohan and Gu, Lin and Lee, Gim Hee},
  booktitle={International Conference on Machine Learning},
  year={2026},
  organization={PMLR}
}

@inproceedings{carreira2017i3d,
  title={Quo vadis, action recognition? a new model and the kinetics dataset},
  author={Carreira, Joao and Zisserman, Andrew},
  booktitle={2017 IEEE conference on computer vision and pattern recognition (CVPR)},
  pages={4724--4733},
  year={2017},
  organization={IEEE}
}

@article{unterthiner2018towards,
  title={Towards accurate generative models of video: A new metric \& challenges},
  author={Unterthiner, Thomas and Van Steenkiste, Sjoerd and Kurach, Karol and Marinier, Raphael and Michalski, Marcin and Gelly, Sylvain},
  journal={arXiv preprint arXiv:1812.01717},
  year={2018}
}

@article{wang2026moge,
  title={Moge-2: Accurate monocular geometry with metric scale and sharp details},
  author={Wang, Ruicheng and Xu, Sicheng and Dong, Yue and Deng, Yu and Xiang, Jianfeng and Lv, Zelong and Sun, Guangzhong and Tong, Xin and Yang, Jiaolong},
  journal={Advances in Neural Information Processing Systems},
  volume={38},
  pages={35928--35959},
  year={2026}
}

@article{liu2023libero,
  title={Libero: Benchmarking knowledge transfer for lifelong robot learning},
  author={Liu, Bo and Zhu, Yifeng and Gao, Chongkai and Feng, Yihao and Liu, Qiang and Zhu, Yuke and Stone, Peter},
  journal={Advances in Neural Information Processing Systems},
  volume={36},
  pages={44776--44791},
  year={2023}
}

@article{han2025d,
  title={D\^{} 2USt3R: Enhancing 3D Reconstruction with 4D Pointmaps for Dynamic Scenes},
  author={Han, Jisang and An, Honggyu and Jung, Jaewoo and Narihira, Takuya and Seo, Junyoung and Fukuda, Kazumi and Kim, Chaehyun and Hong, Sunghwan and Mitsufuji, Yuki and Kim, Seungryong},
  journal={arXiv e-prints},
  pages={arXiv--2504},
  year={2025}
}

@article{cho2021cats,
  title={Cats: Cost aggregation transformers for visual correspondence},
  author={Cho, Seokju and Hong, Sunghwan and Jeon, Sangryul and Lee, Yunsung and Sohn, Kwanghoon and Kim, Seungryong},
  journal={Advances in Neural Information Processing Systems},
  volume={34},
  pages={9011--9023},
  year={2021}
}

@inproceedings{hong2021deep,
  title={Deep matching prior: Test-time optimization for dense correspondence},
  author={Hong, Sunghwan and Kim, Seungryong},
  booktitle={Proceedings of the IEEE/CVF international conference on computer vision},
  pages={9907--9917},
  year={2021}
}

@article{hong2024unifying2,
  title={Unifying feature and cost aggregation with transformers for semantic and visual correspondence},
  author={Hong, Sunghwan and Cho, Seokju and Kim, Seungryong and Lin, Stephen},
  journal={arXiv preprint arXiv:2403.11120},
  year={2024}
}

@article{cho2022cats++,
  title={Cats++: Boosting cost aggregation with convolutions and transformers},
  author={Cho, Seokju and Hong, Sunghwan and Kim, Seungryong},
  journal={IEEE Transactions on Pattern Analysis and Machine Intelligence},
  volume={45},
  number={6},
  pages={7174--7194},
  year={2022},
  publisher={IEEE}
}

@inproceedings{an2025cross,
  title={Cross-view completion models are zero-shot correspondence estimators},
  author={An, Honggyu and Kim, Jin Hyeon and Park, Seonghoon and Jung, Jaewoo and Han, Jisang and Hong, Sunghwan and Kim, Seungryong},
  booktitle={Proceedings of the Computer Vision and Pattern Recognition Conference},
  pages={1103--1115},
  year={2025}
}

@inproceedings{hong2024unifying,
  title={Unifying correspondence pose and nerf for generalized pose-free novel view synthesis},
  author={Hong, Sunghwan and Jung, Jaewoo and Shin, Heeseong and Yang, Jiaolong and Kim, Seungryong and Luo, Chong},
  booktitle={Proceedings of the IEEE/CVF Conference on Computer Vision and Pattern Recognition},
  pages={20196--20206},
  year={2024}
}

@article{hong2024pf3plat,
  title={Pf3plat: Pose-free feed-forward 3d gaussian splatting},
  author={Hong, Sunghwan and Jung, Jaewoo and Shin, Heeseong and Han, Jisang and Yang, Jiaolong and Luo, Chong and Kim, Seungryong},
  journal={arXiv preprint arXiv:2410.22128},
  year={2024}
}

@article{lee2026tora,
  title={TORA: Topological Representation Alignment for 3D Shape Assembly},
  author={Lee, Nahyuk and Chen, Zhiang and Pollefeys, Marc and Hong, Sunghwan},
  journal={arXiv preprint arXiv:2604.04050},
  year={2026}
}

@article{yue2025litept,
  title={Litept: Lighter yet stronger point transformer},
  author={Yue, Yuanwen and Robert, Damien and Wang, Jianyuan and Hong, Sunghwan and Wegner, Jan Dirk and Rupprecht, Christian and Schindler, Konrad},
  journal={arXiv preprint arXiv:2512.13689},
  year={2025}
}

@article{chen2026prose,
  title={PROSE: Training-Free Egocentric Scene Registration with Vision-Language Models},
  author={Chen, Zhiang and Lee, Nahyuk and Sun, Boyang and Kwon, Taein and Pollefeys, Marc and Bauer, Zuria and Hong, Sunghwan},
  journal={arXiv preprint arXiv:2606.16569},
  year={2026}
}

@article{fei2025liberoplus,
  title={Libero-plus: In-depth robustness analysis of vision-language-action models},
  author={Fei, Senyu and Wang, Siyin and Shi, Junhao and Dai, Zihao and Cai, Jikun and Qian, Pengfang and Ji, Li and He, Xinzhe and Zhang, Shiduo and Fei, Zhaoye and others},
  journal={arXiv preprint arXiv:2510.13626},
  year={2025}
}

@article{lim2026hrdexdb,
  title={HRDexDB: A Paired Human-Robot Dataset for Cross-Embodiment Dexterous Grasping},
  author={Lim, Jongbin and Ha, Taeyun and Choi, Mingi and Kim, Jisoo and Kim, Byungjun and Jeon, Subin and Joo, Hanbyul},
  journal={arXiv preprint arXiv:2604.14944},
  year={2026}
}

@article{banerjee2024hot3d,
  title={Introducing hot3d: An egocentric dataset for 3d hand and object tracking},
  author={Banerjee, Prithviraj and Shkodrani, Sindi and Moulon, Pierre and Hampali, Shreyas and Zhang, Fan and Fountain, Jade and Miller, Edward and Basol, Selen and Newcombe, Richard and Wang, Robert and others},
  journal={arXiv preprint arXiv:2406.09598},
  year={2024}
}

@article{pont2017davis,
  title={The 2017 davis challenge on video object segmentation},
  author={Pont-Tuset, Jordi and Perazzi, Federico and Caelles, Sergi and Arbel{\'a}ez, Pablo and Sorkine-Hornung, Alex and Van Gool, Luc},
  journal={arXiv preprint arXiv:1704.00675},
  year={2017}
}

@inproceedings{feng2025st4rtrack,
  title={St4rtrack: Simultaneous 4d reconstruction and tracking in the world},
  author={Feng, Haiwen and Zhang, Junyi and Wang, Qianqian and Ye, Yufei and Yu, Pengcheng and Black, Michael J and Darrell, Trevor and Kanazawa, Angjoo},
  booktitle={2025 IEEE/CVF International Conference on Computer Vision (ICCV)},
  pages={8503--8513},
  year={2025},
  organization={IEEE}
}

@inproceedings{zhou2026pai,
  title={Pai-bench: A comprehensive benchmark for physical ai},
  author={Zhou, Fengzhe and Huang, Jiannan and Li, Jialuo and Ramanan, Deva and Shi, Humphrey},
  booktitle={Proceedings of the IEEE/CVF Conference on Computer Vision and Pattern Recognition},
  pages={21522--21536},
  year={2026}
}

@inproceedings{deng2026rbench,
  title={Rethinking video generation model for the embodied world},
  author={Deng, Yufan and Pan, Zilin and Zhang, Hongyu and Li, Xiaojie and Hu, Ruoqing and Ding, Yufei and Zou, Yiming and Zeng, Yan and Zhou, Daquan},
  booktitle={Forty-third International Conference on Machine Learning},
  year={2026}
}

@article{yan2021videogpt,
  title={Videogpt: Video generation using vq-vae and transformers},
  author={Yan, Wilson and Zhang, Yunzhi and Abbeel, Pieter and Srinivas, Aravind},
  journal={arXiv preprint arXiv:2104.10157},
  year={2021}
}

@article{kondratyuk2023videopoet,
  title={Videopoet: A large language model for zero-shot video generation},
  author={Kondratyuk, Dan and Yu, Lijun and Gu, Xiuye and Lezama, Jos{\'e} and Huang, Jonathan and Schindler, Grant and Hornung, Rachel and Birodkar, Vighnesh and Yan, Jimmy and Chiu, Ming-Chang and others},
  journal={arXiv preprint arXiv:2312.14125},
  year={2023}
}

@article{teng2025magi,
  title={Magi-1: Autoregressive video generation at scale},
  author={Teng, Hansi and Jia, Hongyu and Sun, Lei and Li, Lingzhi and Li, Maolin and Tang, Mingqiu and Han, Shuai and Zhang, Tianning and Zhang, WQ and Luo, Weifeng and others},
  journal={arXiv preprint arXiv:2505.13211},
  year={2025}
}

@inproceedings{wang2024dust3r,
  title={Dust3r: Geometric 3d vision made easy},
  author={Wang, Shuzhe and Leroy, Vincent and Cabon, Yohann and Chidlovskii, Boris and Revaud, Jerome},
  booktitle={2024 IEEE/CVF Conference on Computer Vision and Pattern Recognition (CVPR)},
  pages={20697--20709},
  year={2024},
  organization={IEEE}
}

@inproceedings{asim2025met3r,
  title={Met3r: Measuring multi-view consistency in generated images},
  author={Asim, Mohammad and Wewer, Christopher and Wimmer, Thomas and Schiele, Bernt and Lenssen, Jan Eric},
  booktitle={2025 IEEE/CVF Conference on Computer Vision and Pattern Recognition (CVPR)},
  pages={6034--6044},
  year={2025},
  organization={IEEE}
}

@inproceedings{peebles2023scalable,
  title={Scalable diffusion models with transformers},
  author={Peebles, William and Xie, Saining},
  booktitle={2023 IEEE/CVF International Conference on Computer Vision (ICCV)},
  pages={4172--4182},
  year={2023},
  organization={IEEE}
}

@article{oquab2023dinov2,
  title={Dinov2: Learning robust visual features without supervision},
  author={Oquab, Maxime and Darcet, Timoth{\'e}e and Moutakanni, Th{\'e}o and Vo, Huy and Szafraniec, Marc and Khalidov, Vasil and Fernandez, Pierre and Haziza, Daniel and Massa, Francisco and El-Nouby, Alaaeldin and others},
  journal={arXiv preprint arXiv:2304.07193},
  year={2023}
}

@inproceedings{leroy2024grounding,
  title={Grounding image matching in 3d with mast3r},
  author={Leroy, Vincent and Cabon, Yohann and Revaud, J{\'e}r{\^o}me},
  booktitle={European conference on computer vision},
  pages={71--91},
  year={2024},
  organization={Springer}
}

@inproceedings{fu2024featup,
  title={Featup: A model-agnostic framework for features at any resolution},
  author={Fu, Stephanie and Hamilton, Mark and Brandt, Laura E and Feldmann, Axel and Zhang, Zhoutong and Freeman, William},
  booktitle={International Conference on Learning Representations},
  volume={2024},
  pages={45324--45350},
  year={2024}
}

@inproceedings{jeon2026spatialboost,
  title={Spatialboost: Enhancing visual representation through language-guided reasoning},
  author={Jeon, Byungwoo and Kim, Dongyoung and Jang, Huiwon and Kim, Insoo and Shin, Jinwoo},
  booktitle={European Conference on Computer Vision},
  pages={390--408},
  year={2026},
  organization={Springer}
}
\bibliographystyle{iclr2027_conference}

\clearpage
\appendix

\begin{center}
  {\LARGE\bfseries Appendix Contents}
\end{center}

{\small
\setlength{\parskip}{1.2pt}%
\hypersetup{linkcolor=cadmiumgreen!75!black}%
\startcontents[appendix]
\printcontents[appendix]{l}{1}{\setcounter{tocdepth}{2}}
}

\clearpage

\section{Implementation Details}
\label{app:impl}

\subsection{Training Details}
\label{app:training}

We adopt Cosmos3-Nano~\citep{nvidia2026cosmos3} as the backbone and follow its Mixture-of-Transformers architecture~\citep{liang2024mot}. We train \sname{} with the three-stage pipeline described in \Cref{sec:pipeline}.

In Stage 1, we train the generation pathway, camera embeddings, and view frequencies with $\mathcal{L}_{\mathrm{FM}}$ on all
multi-view sources, including those without trajectory annotations.
In Stage 2, we freeze the remaining parameters and train only the trajectory head and view frequencies with $\mathcal{L}_{\mathrm{track}}$. In Stage 3, we jointly optimize the video backbone and the trajectory head with $\mathcal{L}_{\mathrm{forecast}}$. The reasoner is frozen throughout all stages. For video supervision, we treat the camera pose of each view as an optional condition. For instance, views from \sname{}-Data, where camera control is important, are conditioned on their future camera poses, whereas wrist-camera views from the robot datasets, whose future poses depend on the actions being predicted, are not conditioned on any pose. This also allows \sname{} to forecast without future camera parameters, as required by our main evaluation protocol.
Detailed hyperparameters are provided in \Cref{tab:hyperparams}.

The trajectory head operates on query and key features from the 18th self-attention layer of the generator and performs $M=4$ refinement iterations. During training, we sample $N=512$ query points per clip
from the observed history. 
For the action-enabled variants, we apply $\mathcal{L}_{\mathrm{act}}$ in Stage 3 on action-annotated interactions. 
Following Cosmos~3~\citep{nvidia2026cosmos3}, unified actions are represented as 9-dimensional vectors~\citep{zhou2019continuity}.
\sname{} (V) and \sname{} (V+A) are trained without $\mathcal{L}_{\mathrm{track}}$ and $\mathcal{L}_{\mathrm{corr}}$.
All training is conducted on 32 NVIDIA
GH200 Grace Hopper GPUs.

\begin{table}[h]
\centering
\caption{\textbf{Hyperparameters for each training stage.}}
\label{tab:hyperparams}
\small
\resizebox{0.8\linewidth}{!}{%
\begin{tabular}{lccc}
\toprule
Hyperparameter & Stage 1 & Stage 2 & Stage 3 \\
\midrule
Trainable modules
& Generator, Camera emb.
& Traj. head
& Generator, Traj. head \\
Objective
& $\mathcal{L}_{\mathrm{FM}}$
& $\mathcal{L}_{\mathrm{track}}$
& $\mathcal{L}_{\mathrm{forecast}}$ ($+\lambda_{\mathrm{act}}\mathcal{L}_{\mathrm{act}}$) \\
Optimizer & \multicolumn{3}{c}{AdamW} \\
Learning rate & 1e-4 & 1e-4 & 2e-5 \\
Tokens & 60,000 & 60,000 & 24,000 \\
Training iterations & 5,000 & 2,000 & 5,000 \\
Warm-up steps & \multicolumn{3}{c}{100} \\
Video resolution & \multicolumn{3}{c}{256--480px} \\
Maximum frames & \multicolumn{3}{c}{181} \\
History frames ($h'$) & \multicolumn{3}{c}{1-16} \\
Query points ($N$) & - & 512 & 512 \\
Refinement iterations ($M$) & - & 4 & 4 \\
$\lambda_{\mathrm{seq}}$ / $\lambda_{\mathrm{depth}}$ / $\lambda_{\mathrm{conf}}$
& - & 1 / 1 / 1 & 1 / 1 / 1 \\
$\lambda_{\mathrm{track}}$ & - & - & 0.01 \\
$\lambda_{\mathrm{corr}}$ & - & - & 0.01 \\
$\lambda_{\mathrm{act}}$ & - & - & 0.1 \\
\bottomrule
\end{tabular}
}
\vspace{-0.15in}
\end{table}

\subsection{Evaluation Details}
\label{app:eval}

\noindent\textbf{Evaluation protocol.}
For Image-to-Video (I2V) evaluation, each model receives a single frame. For Video-to-Video (V2V) evaluation, each model receives $h'=16$ frames of multi-view history. The model then predicts the next $h=64$ frames and the output videos are resized to 480p. We use each model’s default number of denoising steps. To evaluate single-view baselines, \ie, HunyuanVideo-1.5, Wan2.2-5B, and Cosmos3-Nano, we generate videos from each view independently.

\noindent\textbf{Video fidelity.}
We compute FVD~\citep{unterthiner2018towards} using I3D~\citep{carreira2017i3d} features and PSNR between generated and ground-truth frames. Both metrics are computed independently for each output view and averaged across views.

\begin{figure}[t]
\centering
\includegraphics[width=0.60\linewidth]{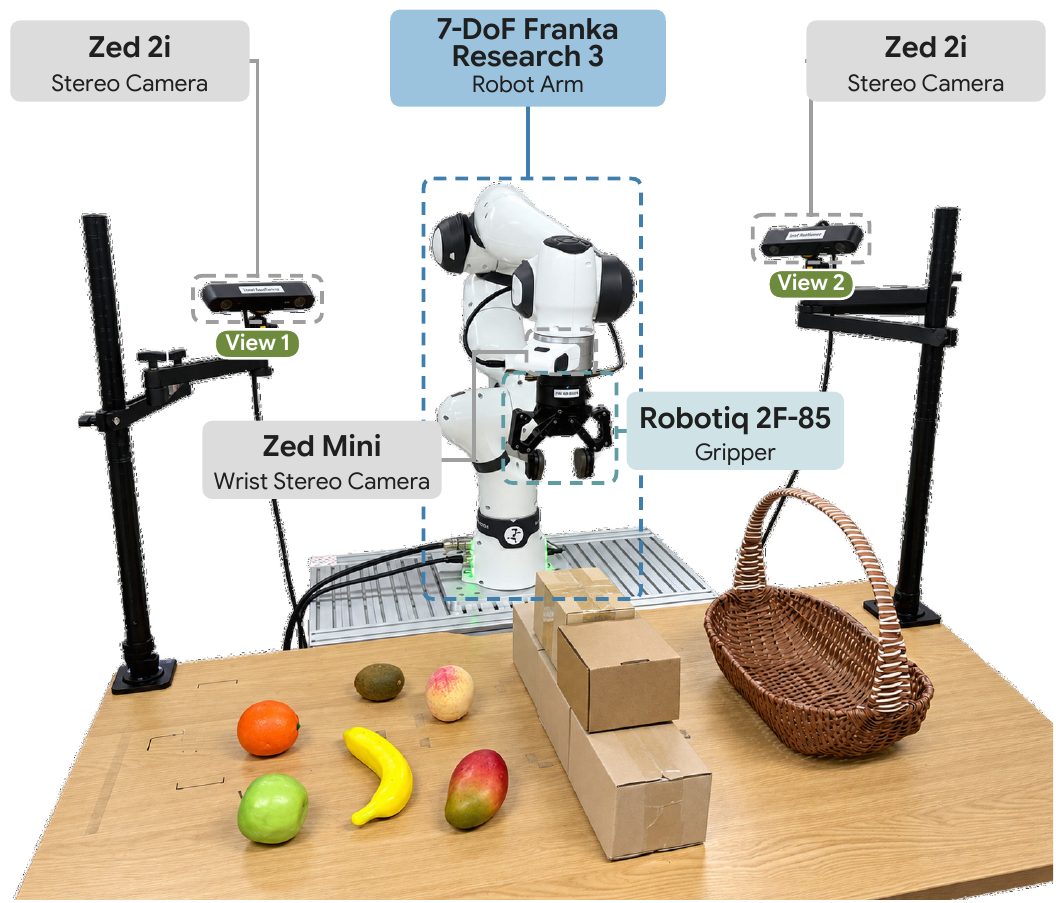}
\caption{\textbf{Real-world hardware platform.}
A 7-DoF Franka Research 3 robot arm equipped with a Robotiq 2F-85 gripper and a wrist-mounted Zed Mini stereo camera.
Two external Zed 2i stereo cameras (view 1 and view 2) observe the workspace from different viewpoints, providing synchronized multi-view observations.}
\label{fig:hardware}
\vspace{-0.15in}
\end{figure}

\noindent\textbf{Geometric consistency of multi-view videos.}
We measure geometric consistency with cross-view MEt3R and $\delta^{\text{3D}}_{\text{avg}}$.
MEt3R~\citep{asim2025met3r} scores the 3D consistency of two images $I_1$ and $I_2$ without ground truth, comparing their DINO features~\citep{oquab2023dinov2} in a common view.
Chosen over different geometric foundation models~\citep{wang2024dust3r,han2025d}, MASt3R~\citep{leroy2024grounding} regresses point maps of both images in the camera frame of $I_1$, and upsampled features through FeatUp~\citep{fu2024featup} attached to these points are rendered into the view of $I_1$, yielding $\hat{F}_1$ and $\hat{F}_2$.
The score averages their cosine dissimilarity over the co-visible pixels $\Omega_{12}$, symmetrized over the two orderings:
\begin{equation}
S(I_1, I_2) = \frac{1}{|\Omega_{12}|} \sum_{p \in \Omega_{12}} \big[ 1 - \cos( \hat{F}_1(p), \hat{F}_2(p) ) \big], \quad
\text{MEt3R}(I_1, I_2) = \tfrac{1}{2} \big[ S(I_1, I_2) + S(I_2, I_1) \big].
\end{equation}
However, MEt3R is typically applied to consecutive frames of a monocular video, assuming large overlap between the two images.
On multi-view videos, raw MEt3R fails to rank videos by quality: real view pairs with little overlap already score high, while generated views drifting apart go unpenalized, since a pair without overlap scores 0.
We thus introduce cross-view MEt3R, measuring the deviation of a generated video's consistency from that of the ground truth.
For each clip $s$ in a split $\mathcal{S}$, we average MEt3R over all pairs of output views at
timesteps $\mathcal{T} = \{t + 4k\}_{k=1}^{h/4}$, obtaining $m(s)$, and report
\begin{equation}
  \text{Cross-view MEt3R} = \frac{1}{|\mathcal{S}|} \sum_{s \in \mathcal{S}}
  \big| m^{\mathrm{gen}}(s) - m^{\mathrm{gt}}(s) \big| .
\end{equation}
Both videos are evaluated with the official implementation on $256 \times 256$ center crops.

For $\delta^{\text{3D}}_{\text{avg}}$, we track query points sampled from the observed history with MVTracker~\citep{rajivc2025mvtracker}, using the generated videos and depth maps from MoGe-2~\citep{wang2026moge}, expressing the trajectories in the fixed reference frame.
With $e_{\tau,n}=\|\widehat{\mathbf{P}}_{\tau,n}-\mathbf{P}_{\tau,n}\|_2$ denoting the error of point $n$ at future timestep $\tau$,
\begin{equation}
\delta^{\text{3D}}_{\text{avg}}
=\frac{1}{|\mathcal{D}|}\sum_{\delta\in\mathcal{D}}
\frac{1}{Nh}\sum_{n=1}^{N}\sum_{\tau=t+1}^{t+h}
\mathbb{1}\left[e_{\tau,n}<\delta\,\frac{d_{\tau,n}}{f}\right],
\end{equation}
where the pixel thresholds $\mathcal{D}=\{1,2,4,8,16\}$ are converted into 3D distances at the ground-truth depth $d_{\tau,n}$ with focal length $f$, following TAPVid-MV~\citep{koppula2026tapvidmv}.

Neither metric reaches its ideal value even for a perfect generator.
For $\delta^{\text{3D}}_{\text{avg}}$, we report an upper bound, evaluating the ground-truth videos with the same pipeline.
For cross-view MEt3R, a generator as consistent as the ground truth still scores above 0 when forecasting a different future, as MEt3R fluctuates over time with the content.
We estimate this floor from the ground-truth videos as
\begin{equation}
  \text{floor} = \frac{1}{|\mathcal{S}|} \sum_{s \in \mathcal{S}} \frac{1}{\sqrt{2}}
  \big| \bar{x}^{(1)}_{s} - \bar{x}^{(2)}_{s} \big| ,
\end{equation}
where $\bar{x}^{(1)}_{s}$ and $\bar{x}^{(2)}_{s}$ average the per-timestep MEt3R of clip $s$, relative to the split average, over the first and second halves of $\mathcal{T}$, with $1/\sqrt{2}$ compensating for the half-length averages.

\noindent\textbf{Robot manipulation.}
For LIBERO~\citep{liu2023libero}, we evaluate on the four standard suites, \ie,
LIBERO-Spatial, LIBERO-Object, LIBERO-Goal, and LIBERO-Long,
each consisting of 10 tasks, with 20 trials per task.
For LIBERO-Plus~\citep{fei2025liberoplus}, we evaluate the policy trained on LIBERO in a zero-shot manner on 10,030 tasks spanning seven perturbation dimensions with one trial per task.
For RoboCasa Kitchen~\citep{nasiriany2026robocasa365}, we evaluate on 24 manipulation tasks with 200 trials per task across five evaluation scenes (10 trials per scene), and average the success rates over three random seeds.
For real-world tasks, we use two exo cameras and one wrist camera, and conduct 24 trials per task.

\subsection{Robot Manipulation Tasks}
\label{app:robot_task}

\begin{figure}[t]
  \centering
  \includegraphics[width=\textwidth]{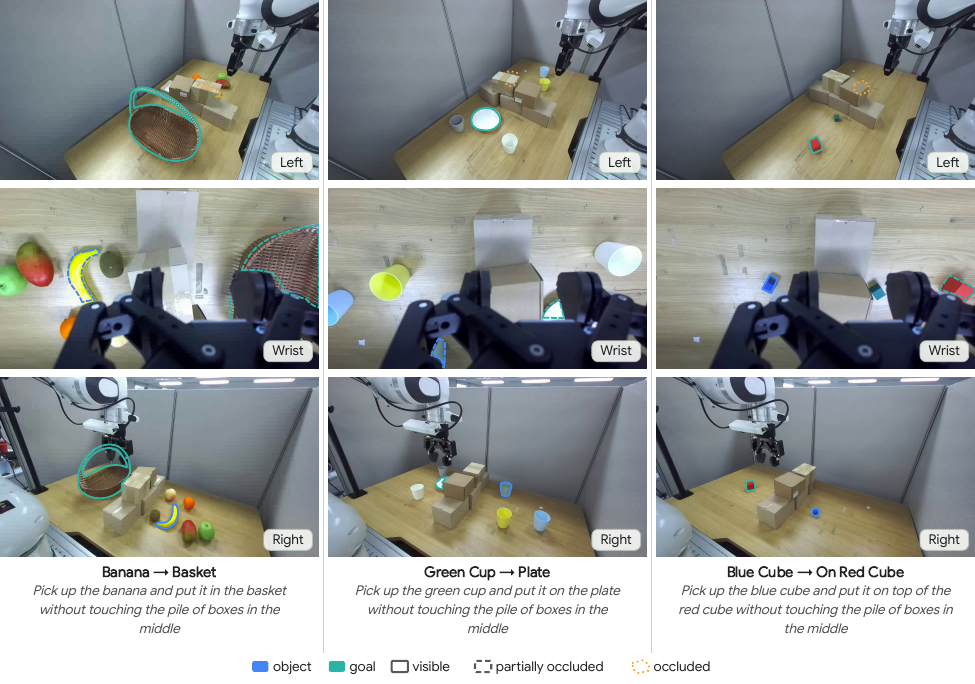}
  \captionsetup{justification=centering}
  \caption{\textbf{Examples of the robot manipulation tasks.}}
  \label{fig:robot_tasks}
\end{figure}

We evaluate all policies on three pick-and-place tasks with a Franka Research 3 arm and the DROID camera setup (left and right exterior cameras and a wrist camera), as shown in \Cref{fig:hardware}.
All tasks share a tabletop layout with a pile of boxes in the middle, and the object and its goal are placed on opposite sides of the pile so that no single camera shows the whole task (\Cref{fig:robot_tasks}).
The robot puts a banana into a basket among other fruits, places a green cup on a plate among cups of other colors, and stacks a small blue cube on top of a red cube next to a green distractor cube, which additionally requires precise placement in height.
Solving these tasks requires fusing the three views into a consistent 3D understanding of the scene, moving across the pile without touching it, and grounding the instruction among same-category distractors.

Each trial is scored as 1 if the object is placed at the goal without touching the pile of boxes, 0.5 if the pick-and-place succeeds but the robot hits the pile on the way, and 0 otherwise.
The success rate of a task is the average score over its 24 trials, which is why some entries in Table~\ref{tab:robot_manipulation} correspond to half-integer counts.

\subsection{Details of Robot Manipulation through 3D Point Trajectory}
\label{app:robot_manip}

\begin{figure}[t]
    \centering
    \includegraphics[width=\textwidth]{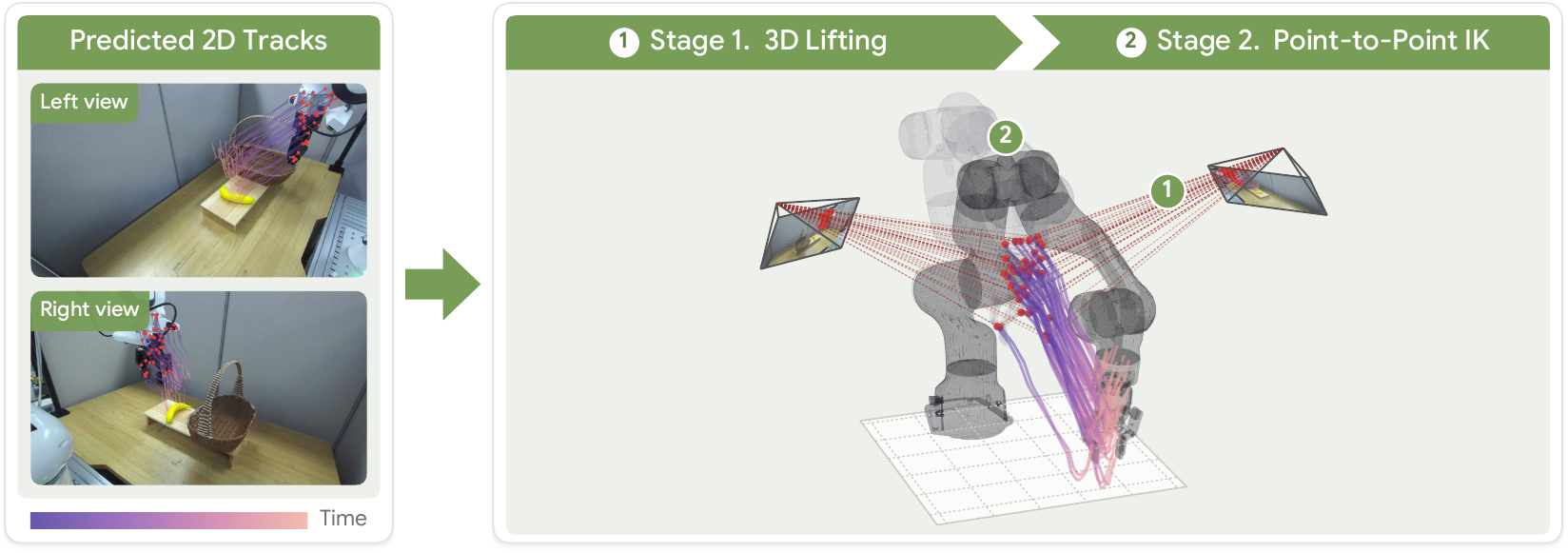}
    \caption{\textbf{From predicted point trajectories to joint-space actions.} Predicted 2D trajectories from both views are lifted to 3D (Stage~1) and converted into a joint trajectory by fitting the robot's kinematic model to the lifted points (Stage~2).}
    \label{fig:track2ik}
\end{figure}

To examine whether the forecast trajectories alone carry actionable information, we bypass the action tokens and convert the predicted trajectories directly into robot commands with a training-free two-stage procedure (\Cref{fig:track2ik} and \Cref{alg:track2ik}).
We sample $N{=}128$ query points on the robot surface from its kinematic model at the current joint state and project them into both exterior views.
Their trajectories are read from the trajectory head at the third of four denoising steps ($\sigma\approx0.83$).
We choose this step because $\sigma\approx0.83$ is the mode of the noise-level distribution used to train the trajectory head, where its predictions are most reliable.
In the first stage, each trajectory is lifted to 3D by triangulating the two views, weighting each observation by its predicted confidence $\widehat{c}^v_{\tau,n}$.
In the second stage, since every query point is attached to a known robot link, we obtain the joint configuration at each future frame by fitting the robot's forward kinematics (FK) to the lifted points with a robust loss, regularized toward the previous frame for a smooth trajectory.
Each point is weighted by $w_{\tau,n}$, its predicted confidence averaged over views, so that unreliable tracks contribute less to the fit.
The resulting joint trajectory, together with the gripper opening recovered from the finger points, is executed in a receding-horizon manner, without action labels or any robot-specific training.

\begin{algorithm}[h]
\caption{Zero-shot joint-space control from predicted point trajectories}
\label{alg:track2ik}
\begin{algorithmic}[1]
\REQUIRE calibrated exterior views $\mathcal{V}$, current joint state $\mathbf{q}_t$, robot kinematics $\mathrm{FK}$, read-out noise level $\sigma^\star$
\STATE Sample $N$ query points on the robot surface at $\mathbf{q}_t$ and project them into each view
\STATE Read 2D tracks $\{\widehat{\mathbf{u}}^v_{\tau,n}\}$ and confidences $\{\widehat{c}^v_{\tau,n}\}$ from the forward pass at noise level $\sigma^\star$
\FOR{$\tau = t+1,\dots,t+h$}
  \STATE \COMMENT{Stage 1: 3D lifting}
  \STATE $\widehat{\mathbf{P}}_{\tau,n} \gets \textsc{Triangulate}\big(\{\widehat{\mathbf{u}}^v_{\tau,n}, \widehat{c}^v_{\tau,n}\}_{v\in\mathcal{V}}\big)$ for all $n$
  \STATE $w_{\tau,n} \gets \frac{1}{|\mathcal{V}|}\sum_{v\in\mathcal{V}} \widehat{c}^v_{\tau,n}$ for all $n$
  \STATE \COMMENT{Stage 2: point-to-point IK}
  \STATE $\mathbf{q}_\tau \gets \arg\min_{\mathbf{q}} \sum_n w_{\tau,n}\,\rho\big(\|\mathrm{FK}_n(\mathbf{q})-\widehat{\mathbf{P}}_{\tau,n}\|\big) + \lambda\|\mathbf{q}-\mathbf{q}_{\tau-1}\|^2$
\ENDFOR
\STATE Execute the first steps of $\{\mathbf{q}_\tau\}$ and re-plan from the new observation
\end{algorithmic}
\end{algorithm}

\clearpage

\section{Dataset Curation}
To train \sname{}, we construct 600K samples from three sources: (i) \sname{}-Data, our multi-view videos with persistent 3D point trajectories from simulation, (ii) robot manipulation video datasets annotated with kinematics-grounded 3D point trajectories, and (iii) open-source video datasets from diverse domains, \eg,
MolmoMotion-1M~\citep{zhang2026molmomotion} and Ego-Exo4D~\citep{grauman2024egoexo4d}. We provide detailed statistics in \Cref{app:data_stats}, and elaborate on the construction of \sname{}-Data in \Cref{app:data} and robot data in \Cref{app:robotpipeline}, respectively.

\subsection{Data Statistics}
\label{app:data_stats}

We collect 600K samples in total, as summarized in
\Cref{tab:train_data}. \sname{}-Data contains 10,000 multi-view samples, consisting of 8,000 tabletop manipulation clips and 2,000 urban dynamics clips. The robot datasets contain 91K episodes from DROID~\citep{khazatsky2024droid} and  HRDexDB~\citep{lim2026hrdexdb}. The remaining clips are collected and sampled from open-source datasets.

\begin{table}[h]
\centering
\caption{\textbf{Statistics of the training data.}}
\label{tab:train_data}
\small
\resizebox{0.9\linewidth}{!}{%
\begin{tabular}{llcrccc}
\toprule
Source & Domain & Real/Sim & Samples & Views & 3D traj. & Action \\
\midrule
\sname{}-Data & Tabletop manipulation & Sim & 8,000 & 3 & \cmark & \xmark \\
\sname{}-Data & Urban dynamics & Sim & 2,000 & 4 & \cmark & \xmark \\
DROID & Tabletop manipulation & Real & 89,312 & 3 & \cmark & \cmark \\
HRDexDB & Tabletop manipulation & Real & 1,601 & 2 & \cmark & \cmark \\
MolmoMotion-1M & Mixed domain & Real & 480,000 & 1-2 & \cmark & \xmark \\
Ego-Exo4D & Human activities & Real & 24,000 & 1-4 & \cmark & \xmark \\
\midrule
Total & & & 604,913 & & & \\
\bottomrule
\end{tabular}
}
\end{table}

\subsection{3D Point Extraction Pipeline from Robot Videos}
\label{app:robotpipeline}
We describe our pipeline for extracting the 3D point trajectories used to train our model. Our objective is to recover metric, temporally consistent 3D trajectories of action-relevant geometry throughout each manipulation episode. Rather than relying on noisy pseudo-labels from visual point trackers, we construct these trajectories directly from known robot and object geometry together with recorded kinematic states. In particular, robot CAD models provide accurate surface geometry, which can be transformed according to the robot state at each timestep to obtain geometrically grounded 3D point trajectories. We detail this procedure below.

\noindent{\bf 3D Point Extraction.}
We begin with the CAD or visual meshes of each robot link and manipulated object. We assume calibrated cameras and a known transformation between the robot base and each camera coordinate frame. For each mesh, we sample surface points using area-weighted triangle sampling. Sampling is performed once in the mesh-local coordinate system with a fixed random seed, and the same points are reused throughout the entire episode. By subsequently transforming these fixed local points according to the corresponding link or object pose at each timestep, we obtain temporally consistent 3D point trajectories without resampling noise.

We allocate a fixed point budget to each action-relevant component: we sample 256 3D points from the whole robot body. In cases where the 3D points of the target object are available, such as in simulation settings, we also sample 256 points from the object being manipulated. For articulated components composed of multiple meshes, the budget is distributed approximately in proportion to mesh surface area. Let $\mathbf{p}^{\ell}_{n}$ denote a point sampled on the local CAD geometry of link or
object $\ell$. Its world-space location at timestep $\tau$ is obtained as
\[
  \mathbf{p}^{w}_{\tau,n} = \mathbf{R}^{w}_{\ell,\tau}\,\mathbf{p}^{\ell}_{n} + \mathbf{b}^{w}_{\ell,\tau},
\]
where $(\mathbf{R}^{w}_{\ell,\tau}, \mathbf{b}^{w}_{\ell,\tau})$ denotes the pose of the
corresponding link or object. Because $\mathbf{p}^{\ell}_{n}$ is fixed, point $n$ retains
the same physical identity over the complete trajectory.
For these sampled points, we apply forward kinematics of the robot via saved robot states, and obtain their 3D positions per time step. For simulated datasets, we replay the recorded simulator state and directly query the per-frame poses of robot links, objects, and cameras. The sampled CAD points can therefore be transformed using the simulator's underlying geometry without requiring visual pose estimation. For real-robot datasets, we load the corresponding robot model and visual meshes, and manipulate the robot CAD via saved robot states and forward kinematics.

\noindent{\bf Camera-frame trajectories and projection.}
Using the extracted 3D points, we first construct point trajectories in a common world frame and subsequently transform them into each camera coordinate system. For camera $v$, the position $\mathbf{p}^{v}_{\tau,n}$ of point $n$ at timestep $\tau$ is
given by
\[
  \mathbf{p}^{v}_{\tau,n} = \mathbf{R}^{v\leftarrow w}_{\tau}\,\mathbf{p}^{w}_{\tau,n} + \mathbf{b}^{v\leftarrow w}_{\tau},
\]
where $(\mathbf{R}^{v\leftarrow w}_{\tau}, \mathbf{b}^{v\leftarrow w}_{\tau})$ denotes the
world-to-camera extrinsics of camera $v$ at timestep $\tau$. Because the same point identities are preserved under these transformations, this representation provides explicit point correspondences~\citep{yue2025litept,chen2026prose,lee2026tora} across both time and viewpoints.

\noindent{\bf Data Filtering for Robot Data.}
Although the real-robot datasets provide camera calibration, even relatively small errors, such as errors in camera extrinsics or in the robot-base calibration can create substantial image-space misalignment between the projected CAD model and the observed robot. We therefore explicitly validate the geometric alignment of each real episode before retaining it.

We evaluate calibration quality using agreement between the projected CAD geometry and an independently obtained image-space robot mask. We first use the recorded joint configuration and camera calibration to solve the robot kinematics and render the corresponding CAD model into every frame, producing a binary projected CAD mask $M^{\mathrm{CAD}}_t$.

Rather than evaluating all frames equally, we select the frame $\tau^\star$ in which the projected robot occupies the largest image area, defined as
\[
\tau^\star
=
\arg\max_\tau
\left|M^{\mathrm{CAD}}_\tau\right|.
\]
This frame provides the strongest calibration signal because a larger visible robot region reduces sensitivity to small masks, truncation, and weak overlap statistics.

For the selected frame $\tau^\star$, we obtain an image-derived robot mask $M^{\mathrm{SAM}}_{\tau^\star}$ using SAM2~\citep{ravi2025sam} and compare it against the rendered CAD mask. We use CAD-mask recall, defined as
\[
\mathrm{Recall}_{\mathrm{CAD}}
=
\frac{
\left|
M^{\mathrm{CAD}}_{\tau^\star}
\cap
M^{\mathrm{SAM}}_{\tau^\star}
\right|
}{
\left|
M^{\mathrm{CAD}}_{\tau^\star}
\right|
},
\]
which measures the fraction of geometry predicted by the calibrated robot model that is supported by the observed image mask. We retain episodes with a recall of at least $0.9$ and discard those below this threshold. Low recall indicates substantial disagreement between the kinematically projected CAD model and the image evidence, suggesting that the resulting 3D supervision may be unreliable.

We find recall preferable to symmetric mask-overlap metrics, as the SAM2 mask may include regions that do not belong to the robot or may merge the robot with nearby objects. Our primary requirement is instead that the geometry used to construct the 3D supervision projects consistently onto the robot observed in the image.

\noindent{\bf Output representation.}
The resulting sample contains synchronized multi-view RGB observations together with persistent 3D surface-point trajectories for the robot arm, gripper or hand, and manipulated objects. We additionally retain the recorded joint states and all camera metadata required to reproduce the trajectories and their image-space projections, including intrinsics, distortion parameters, and per-frame extrinsics. Consequently, each point index refers to the same physical surface location across both time and camera viewpoints.

\clearpage

\section{Training Data Generation}
\label{app:data}

We use separate pipelines for tabletop manipulation and urban
capture, with a shared annotation format. This section describes
training-data generation; benchmark construction is detailed in
\Cref{app:bench}.

\subsection{Capture Settings and Environments}
\label{app:envs}
\label{app:stats}

\Cref{tab:backends} summarizes the capture settings; corpus
composition is reported in \Cref{tab:data_bench}.
Tabletop training uses city environment $C_1$ as static scenery,
whereas urban training captures live agents in $C_2$.
The benchmark additionally uses unseen kitchen environments $H$
for manipulation and moving agents in $C_1$ for urban evaluation.
The latter environment is therefore familiar as scenery, but
its traffic and crowd dynamics are absent from training.

\begin{table}[h]
\centering
\small
\setlength{\tabcolsep}{4pt}
\caption{\textbf{Training-data capture settings.} Both domains use
$1280{\times}1280$ images and distortion-free pinhole cameras.}
\label{tab:backends}
\label{tab:envs}
\begin{tabular}{@{}p{0.24\linewidth}p{0.31\linewidth}p{0.37\linewidth}@{}}
\toprule
Setting & Tabletop & Urban \\
\midrule
Motion source & Recorded manipulation & Live traffic and crowds \\
Views / frame rate & 3 / 20\,fps & 4 / 15\,fps \\
Clip length & 81--181 frames & 81, 100, 121, 144, or 181 frames \\
Intrinsics & Authored & Sampled per camera and clip \\
Camera motion & Static or planned hard-clip paths
& Static, arc, translation, random polyline, or pan--tilt \\
Visibility & Rendered depth, $3$\,cm tolerance
& Surface-point ray casting \\
Reference camera & \texttt{cam00} & \texttt{cam01} \\
\bottomrule
\end{tabular}
\end{table}

\subsection{Tabletop Generation}

\noindent\textbf{Motion and scene composition.}
We replay physics-simulated pick-and-place episodes of a
Fourier GR-1~T2 robot using recorded joint configurations
and object transforms. Each clip contains 1--4 targets or
receptacles, augmented with 5--8 non-contact and 0--2 contact
distractors drawn from 306 props with annotated resting poses.
All distractors avoid the manipulated objects' swept volumes.
Contact distractors additionally overlap the robot's swept
volume, with at least 25\% of a $3{\times}3{\times}3$ grid
inside their bounding box lying within that volume.
Non-contact props alternate between the robot's left and right.

\noindent\textbf{Physics validation.}
Distractors are simulated at 60\,Hz, with three substeps per
video frame. They first settle for up to 60 frames, requiring
translation below $0.1$\,cm and rotation below $0.5^\circ$
for five consecutive frames. Placements with falling props
are rejected. During replay, contact distractors must reach
$1$\,cm contact depth for at least two frames and move by
at least $1$\,cm. We allow up to five placement attempts,
then bake the accepted per-frame transforms for rendering.

\noindent\textbf{Camera and background placement.}
Camera poses are perturbed in robot-local coordinates:
distance is scaled by $[0.94,1.06]$, azimuth by $\pm4^\circ$,
and elevation by $\pm3^\circ$. Cameras are re-aimed at the
work surface while remaining at least $5$\,cm above it and
$3$\,cm clear of the robot. The complete scene and camera rig
are then relocated into $C_1$, accepting only placements
where the entire camera path remains clear of scene geometry.

\subsection{Urban Capture}

The camera rig is placed at one of 13 surveyed locations in
$C_2$. After the traffic and crowd simulation settles, we
select a visible pedestrian or vehicle satisfying the motion
criteria, with a separate quota of stationary targets.
Targets use 12 vehicle models and six pedestrian body meshes;
procedural combinations of clothing, hair, and footwear
provide additional appearance variation.

Each camera independently samples a horizontal field of view
between $25^\circ$ and $90^\circ$ and a trajectory type from
\Cref{tab:backends}. Cameras aim at the target except during
scheduled departures. Images and annotations are exported
from the same simulation timestep across all views.
The scheduler balances the five clip-length buckets, with
hard clips restricted to 181 frames.

\subsection{Geometric and Language Annotations}
\label{app:schema}

\noindent\textbf{Persistent surface points.}
We sample 256 points per annotated object. Tabletop samples
are cached per mesh and propagated using recorded transforms;
urban samples are initialized per acquired target. On deforming
meshes, component and triangle identities with barycentric
coordinates recover the same surface points from the deformed
vertices at each timestep. Tabletop clips additionally contain
512 robot points, split equally between the arms. Allocation
is proportional to link surface area with a minimum of four
points per link, reserving half of each arm's budget for
the hand.

\noindent\textbf{Visibility.}
Tabletop visibility uses rendered depth with a $3$\,cm tolerance.
Urban visibility tracks up to 32 in-frustum surface points
per camera and frame, excluding the target itself from
collision tests and allowing a $15$\,cm margin.
The target is visible if at least three points are unoccluded,
out of frustum if none is in frame, and occluded otherwise.
Invalid surface reconstructions count as occluded.
Urban clips retain separate bounding-box and traced visibility
arrays: the former supports the visibility acceptance check,
while the latter controls departures, re-acquisition, and
hard labels.

\noindent\textbf{Annotation format.}
Camera files store intrinsics and per-frame $4{\times}4$
world-to-camera matrices. Point arrays use reference-camera
coordinates in metres under the OpenCV axis convention.
Object trajectories have shape $(T, 256N_{\mathrm{obj}}, 3)$ for a $T$-frame clip with $N_{\mathrm{obj}}$ annotated objects, including contact distractors, and tabletop robot trajectories have shape $(T, 512, 3)$. Metadata records object index ranges,
roles, point validity, per-view visibility, and hard-label
measurements. Each clip contains per-camera RGB videos,
\texttt{depth.zip}, \texttt{cameras.json},
\texttt{points\_target.npz}, \texttt{visibility.json},
\texttt{meta.json}, and \texttt{caption.txt}.

\noindent\textbf{Captions.}
Motivated by SpatialBoost~\citep{jeon2026spatialboost}, we generate captions per clip through a vision-language model and filter them. Captions must agree with recorded motion, omit numerical claims and camera descriptions, and satisfy the single-sentence format. Failures are regenerated with the validation feedback
for up to two rounds; unresolved failures are excluded.

\subsection{Visibility-Controlled Clips}
\label{app:hard}

Hard clips introduce incomplete individual-camera observations
while retaining target coverage across the camera set.
\Cref{tab:hardlabel} states the training-label criteria;
labels are verified from measured visibility rather than
assigned from the camera schedule alone.

\begin{table}[t]
\centering
\small
\setlength{\tabcolsep}{4pt}
\caption{\textbf{Hard-clip criteria for the training pipelines.}
An invisible interval comprises frames labeled occluded or
out of frustum. Benchmark definitions and measured statistics
are reported separately in \Cref{app:bench}.}
\label{tab:hardlabel}
\begin{tabular}{@{}L{0.24\linewidth}L{0.32\linewidth}L{0.36\linewidth}@{}}
\toprule
Criterion & Tabletop & Urban \\
\midrule
Invisible interval & $\geq30$ consecutive frames per camera
& $\geq30$ per camera, followed by re-acquisition \\
Collective coverage & At least one visible camera per frame
& At least one visible camera per frame \\
Motion in interval & Target displacement $\geq5$\,cm
& Target moving in $\geq40\%$ of interval frames \\
Additional condition & $\geq20$ occluded frames summed over cameras
& Two-phase departure schedule \\
\bottomrule
\end{tabular}
\end{table}

\noindent\textbf{Tabletop construction.}
One camera starts $8$--$12$\,cm above the work surface.
Non-contact occluders are placed along camera--target segments
at relative positions $0.40$--$0.85$; when placement fails,
cameras may rotate away by up to $75^\circ$.
Paths are limited to $2.5$\,cm translation and $2^\circ$
rotation per frame, with $30$\,cm inter-camera separation,
$20^\circ$ yaw separation, and $15$\,cm forward lens clearance.
Invisible-window starts are searched within $\pm60$ frames
at two-frame intervals. Collision tracks to 32 target points
verify the schedule during physics replay. Failed attempts
revert to ordinary camera jitter and receive a normal label.

\noindent\textbf{Urban construction.}
All four cameras depart and re-acquire the target in two
phases. A new phase starts only while a camera from the
preceding phase remains traced-visible; capture aborts if
collective visibility is lost. Hard clips use the 181-frame
bucket, with a scheduler target of 10--20\% of the urban
training corpus. Failed hard attempts are discarded and
replaced.

\subsection{Validation and Rendering}
\label{app:gates}
\label{app:render}

Clips are checked against the criteria in \Cref{tab:gates}.
Failed samples are replaced, and samples with unavailable
validation inputs are held rather than accepted.
The training profile records departure--re-acquisition as
a label; the stricter benchmark profile additionally enforces
it as an acceptance condition.

\begin{table}[h]
\centering
\small
\setlength{\tabcolsep}{4pt}
\caption{\textbf{Configured acceptance checks.} Departure--re-acquisition
is recorded but not required by the training profile.}
\label{tab:gates}
\begin{tabular}{@{}p{0.25\linewidth}p{0.70\linewidth}@{}}
\toprule
Check & Threshold \\
\midrule
Target visibility & Minimum visible ratio $\geq0.1$ over all cameras \\
Survival and motion & Target present throughout; travel
$\geq1.0$\,m per 81 frames, with a $0.5$\,m floor \\
Track completeness & No track breaks; completeness $\geq0.99$ \\
Depth validity & Sky, near, and constant-value pixel ratios
each $\leq0.25$ \\
Camera clearance & At most two cameras with blocked paths;
view-block ratio $\leq0.6$ \\
Caption & Present, at least 20 characters, and passes format checks \\
Departure cycle & At least one cycle per scheduled camera,
with $\geq8$ frames out and $\geq8$ frames visible \\
Camera smoothness & Jolt $\leq40$\,cm; step $\leq160$\,cm;
rotation $\leq0.08$ of horizontal field of view per frame;
vertical vibration $\leq4$\,cm peak-to-peak at $1.3$--$3.0$\,Hz \\
Motion persistence & Target speed $\geq40$\,cm/s
in at least 40\% of frames \\
\bottomrule
\end{tabular}
\end{table}

For urban surface tracks, reprojection checks require at least
99\% of points to agree within one pixel, with no track breaks.
Tabletop rendering uses two spatial samples per pixel and
32 warm-up frames; urban capture uses a warm-up and settling
schedule. RGB is encoded as H.264, and optical-axis depth is
stored as 16-bit PNG in centimetres, valid from $1$\,cm to
$200$\,m. Per-camera valid-depth ratios are recorded.
Images and annotations use the same evaluated scene state.

Tabletop randomization uses per-clip seeds, and validated
transforms are replayed during rendering. Urban seeds are
recorded, but the live traffic and crowd simulation does not
guarantee bit-exact regeneration.

%--------------------------------------
\clearpage

\section{\bname{}: Construction and Statistics}
\label{app:bench}

This appendix documents the simulation component of \bname{},
which pairs synchronized multi-view videos with metric surface-point
trajectories, camera geometry, and per-view visibility. Its capture
pipelines are distinct from the training-data pipelines in
\Cref{app:data}. Real-world benchmark details are reserved for
\Cref{app:real_bench}.

\subsection{Composition and Capture Settings}
\label{app:bench_req}
\label{app:bench_stats}

The simulation benchmark contains 500 tabletop and 500 urban clips,
with 661,071 images and 716,544 persistent point trajectories
(\Cref{tab:bs_comp}). Images and annotations correspond to the same
simulation timesteps, and point identities remain defined during
occlusion. Reported statistics are measured from the released files.

\begin{table}[h]
\centering
\small
\setlength{\tabcolsep}{4pt}
\caption{\textbf{Simulation benchmark composition and capture settings.}
Timesteps are counted once per synchronized camera set; images count
all views. Trajectory counts sum persistent point identities over clips.
All images are $1280{\times}1280$, with distortion-free pinhole cameras.}
\label{tab:bs_comp}
\label{tab:bs_cams}
\begin{tabular}{@{}p{0.32\linewidth}p{0.29\linewidth}p{0.32\linewidth}@{}}
\toprule
Property & Tabletop & Urban \\
\midrule
Clips (hard) & 500 (27) & 500 (95) \\
Views / frame rate & 3 / 20\,fps & 4 / 15\,fps \\
Clip length & 81--680 frames & 81--181 frames \\
Synchronized timesteps & 129,441 & 68,187 \\
RGB images & 388,323 & 272,748 \\
3D point trajectories & 588,544 & 128,000 \\
Horizontal field of view & $93.0^\circ$, $100.5^\circ$, $67.0^\circ$
& $25.0^\circ$--$89.9^\circ$ \\
Camera translation & None within clips & Median 10.7\,cm/frame \\
Reference camera & \texttt{cam00} & \texttt{cam01} \\
Median target path length & 0.867\,m & 19.0\,m \\
\bottomrule
\end{tabular}
\end{table}

Tabletop clips cover city environment $C_1$ (436 clips) and two
kitchens, $k_1$ (28) and $k_2$ (36). They contain 1,299 tracked
objects: 693 manipulation targets and 606 receptacles.
Of the 500 clips, 392 retain their source-episode length and
108 are windowed to 81--181 frames. Urban clips contain
332 pedestrians and 168 vehicles spanning 55 target models,
with at most 11 clips per model. Their lengths are
81, 100, 121, 144, and 181 frames, containing
79, 79, 83, 82, and 177 clips, respectively.
All 95 hard urban clips belong to the 181-frame group.
The relationship of these environments to training is described
in \Cref{app:envs}.

\subsection{Tabletop Capture}
\label{app:bench_tabletop}

\noindent\textbf{Scene composition.}
Each clip replays a physics-simulated Fourier GR-1~T2
pick-and-place episode in Unreal Engine. Source sequences
contain 54 robot joint tracks, 1--4 replay objects, a work
surface, three cameras, and swept-volume geometry for the
robot and objects. Seeded randomization varies camera poses,
surface materials, and clutter. Distractors avoid the moving
objects' swept volumes, with a quota of 0--2 props intersecting
the robot's swept volume. Props settle and interact with the robot under physics, after which their transforms are recorded for rendering.
Released clips contain 1--12 distractors (median 8); contacts and subsequent motion are retained.

\noindent\textbf{Cameras and rendering.}
The three cameras retain their authored intrinsics and receive
robot-local pose perturbations: distance scaled by $[0.94,1.06]$,
azimuth by $\pm4^\circ$, and elevation by $\pm3^\circ$.
They are aimed at the manipulation region. Camera positions
remain fixed within each benchmark clip; hard clips additionally
use the rotations described in \Cref{app:bench_hard}.
Per-frame world transforms are read back for all objects,
36 robot links, and cameras, including their parent transforms.
RGB and depth are rendered at these timestamps with two spatial
samples per frame. All clips use a fixed AgX tone curve and
recorded lighting settings; lighting is not randomized in this
benchmark.

\subsection{Urban Capture}
\label{app:bench_urban}

\noindent\textbf{Agents and synchronization.}
Urban clips are captured in an open-source driving simulator
built on Unreal Engine, using six transformed city tiles
with 60 additional building models and fixed noon weather.
Vehicles follow the traffic manager and pedestrians follow
navigation destinations, amid approximately 30 background
vehicles and 40 pedestrians. A target is acquired once its
instance occupies at least 300 pixels in a camera; capture
starts after 15 frames of motion. The simulator advances at
$1/15$\,s intervals and waits for every sensor before the next
tick. Each view contains pixel-aligned RGB, depth, and
instance-segmentation sensors, with motion blur and depth of
field disabled.

\noindent\textbf{Camera configuration.}
Each camera independently samples its horizontal field of view
from $25^\circ$ to $90^\circ$. Initial positions are selected
from 32 ray-tested azimuths around the target, with distance
chosen to control its image coverage. Cameras follow static,
arc, translation, random-polyline, or pan--tilt paths and aim
at the target except during scheduled departures.
Path control limits translation to 150\,cm per frame,
changes in the translation step to 20\,cm, and rotation to
$0.06$ times the horizontal field of view per frame.
A two-stage height filter with a $0.5$\,s time constant
stabilizes camera aiming without modifying the target's
recorded trajectory.

\subsection{Ground-Truth Annotations}
\label{app:bench_gt}

\noindent\textbf{Persistent points.}
We sample 256 area-weighted surface points per object once
per mesh and preserve their identities across frames and clips.
Tabletop points follow recorded rigid transforms. The robot
has 512 additional points, divided equally between its arms
and allocated across 18 links per side, with at least four
points per link and half of each arm's budget reserved for
the hand and fingers. For urban targets, points store triangle
indices, barycentric coordinates, and skinning weights.
Per-frame bone transforms recover their positions through
linear blend skinning, with rigid components propagated from
their attachment transforms. These annotations follow skeletal
deformation, not simulated cloth or hair.

\noindent\textbf{Coordinates and depth.}
Stored trajectories use the reference camera's frame at each
timestep, in metres under the OpenCV convention
($+X$ right, $+Y$ down, $+Z$ forward).
Per-frame $4{\times}4$ world-to-camera matrices and intrinsics
are stored for every view, allowing conversion to world
coordinates. Motion-based acceptance checks use world-space
target motion. Depth is stored as 16-bit PNG in centimetres,
with far limits of 200\,m for kitchens, 650\,m for tabletop
city scenes, and 655\,m for urban scenes.

\noindent\textbf{Visibility labels.}
For tabletop points projected inside the image, visibility
requires camera-space depth $z \leq d+\epsilon$, where $d$
is rendered depth and $\epsilon=3$\,cm. An object is visible
when more than 10\% of its in-frustum points pass this test;
labels also distinguish occlusion, departure from the frustum,
and absence. Urban clips contain two target-visibility arrays.
\texttt{visibility} uses bounding-box projection and requires
at least 300 target-instance pixels for visibility.
\texttt{visibility\_strict} instead uses projected surface
points with the same instance-pixel criterion.
The bounding-box array supports the visibility acceptance
check, while the strict array determines departures,
re-acquisition, and hard labels.

\noindent\textbf{Language annotations.}
Each clip receives an English caption from a vision-language
model. Tabletop captioning uses the recorded object identities,
motion, active hand, and scene context; its style checks are
advisory. Urban captions are generated from representative
video frames after selection, and failing captions are
regenerated. All 500 urban captions pass the final style checks.

\subsection{Visibility-Controlled Subsets}
\label{app:bench_hard}

Hard subsets introduce sustained visibility losses in individual
views while preserving target visibility across the camera set.
\Cref{tab:bs_hard} summarizes their domain-specific criteria
and measured composition.

\begin{table}[t]
\centering
\small
\setlength{\tabcolsep}{4pt}
\caption{\textbf{Hard-subset criteria and composition.} An invisible
interval includes occluded and out-of-frustum frames.
Urban collective coverage is evaluated on the strict array.}
\label{tab:bs_hard}
\begin{tabular}{@{}p{0.31\linewidth}p{0.30\linewidth}p{0.32\linewidth}@{}}
\toprule
Property & Tabletop & Urban \\
\midrule
Hard clips & 27 & 95 \\
Cameras with $\geq30$ consecutive invisible frames
& All 3 & 2 or 3$^\dagger$ \\
Target motion in interval
& Displacement $\geq5$\,cm & Moving fraction $\geq0.4$ \\
Additional occlusion requirement
& $\geq20$ frames summed over views & None \\
Frames with no visible camera
& 0 & 0 (strict) \\
\bottomrule
\end{tabular}

\smallskip
\begin{minipage}{\linewidth}
\footnotesize
$^\dagger$All four urban cameras complete departure and re-acquisition cycles. Of the 95 hard urban clips, 31 have two and 64 have
three cameras meeting the 30-frame threshold.
\end{minipage}
\end{table}

\noindent\textbf{Tabletop construction.}
One camera is placed 8--12\,cm above the work surface and
55--70\,cm from the target region to induce occlusion by
arms and clutter. The other two rotate away in staggered
windows, overlapping only while the low camera is predicted
to retain visibility. Rotation is limited to $0.08$ times
the camera's half field of view per frame. Planned visibility
is estimated from 32 target surface points; the camera paths
are then recorded and the hard label recomputed from rendered
depth. Failed hard attempts are retried once before reverting
to normal clips. Of 100 attempts, 27 satisfy the hard criteria.

\noindent\textbf{Urban construction.}
All four cameras depart and re-acquire the target in two
phases. The second phase begins only while a camera from
the preceding phase is strict-visible, and capture aborts
if collective visibility is lost. These clips use 181 frames.
Normal clips may also include departures: each camera is
scheduled with probability 0.3, with at most two departures
per clip and none in the 81-frame group.

\noindent\textbf{Measured coverage.}
\Cref{tab:bs_ncams} reports target visibility across cameras.
No hard tabletop frame is invisible to all views. The urban
bounding-box array contains one such hard frame, whereas the
strict array used for labeling contains none.
These measurements characterize the temporal availability of
direct target observations across the camera set.

\begin{table}[h]
\centering
\small
\setlength{\tabcolsep}{5pt}
\caption{\textbf{Percentage of frames by the number of cameras seeing the target.} Urban columns use bounding-box visibility;
strict-array collective coverage is reported separately.}
\label{tab:bs_ncams}
\begin{tabular}{@{}lrrrr@{}}
\toprule
& \multicolumn{2}{c}{Tabletop ($V=3$)}
& \multicolumn{2}{c}{Urban ($V=4$)} \\
\cmidrule(lr){2-3}\cmidrule(lr){4-5}
Visible cameras & Normal & Hard & Normal & Hard \\
\midrule
4 & N/A & N/A & 71.81 & 36.48 \\
3 & 89.67 & 28.60 & 21.63 & 24.41 \\
2 & 9.65 & 48.49 & 5.78 & 29.34 \\
1 & 0.67 & 22.91 & 0.67 & 9.77 \\
0 & 0.012 & 0.000 & 0.12 & $0.01^\dagger$ \\
\midrule
Timesteps & 121,406 & 8,035 & 50,992 & 17,195 \\
\bottomrule
\end{tabular}

\smallskip
\begin{minipage}{\linewidth}
\footnotesize
$^\dagger$One urban hard frame has no visible camera under
bounding-box visibility; zero of 17,195 hard frames do so
under strict visibility.
\end{minipage}
\end{table}

\subsection{Validation and Limitations}
\label{app:bench_gates}

\noindent\textbf{Acceptance checks.}
Tabletop clips are validated after rendering for point--depth
agreement, camera-pose consistency, depth validity, target
visibility, object-trajectory clearance, hard-label agreement,
camera and pose-array completeness, sweep-geometry alignment,
and caption availability. Target, arm, and hand point--depth
checks require median error no greater than 3\,cm.
Urban clips follow the thresholds in \Cref{tab:bench_gates}.
Failed clips are replaced and their rejection reasons retained;
checks with missing inputs do not pass.

\begin{table}[t]
\centering
\small
\setlength{\tabcolsep}{4pt}
\caption{\textbf{Urban benchmark acceptance thresholds.} The caption check is applied after selection.}
\label{tab:bench_gates}
\begin{tabular}{@{}p{0.28\linewidth}p{0.67\linewidth}@{}}
\toprule
Check & Threshold \\
\midrule
Target visibility & Minimum visible ratio $\geq0.1$ over
cameras without scheduled departures \\
Survival and motion & Target present throughout; travel
$\geq1.0$\,m per 81 frames, with a 0.5\,m floor \\
Track completeness & Non-empty points, no track breaks,
completeness $\geq0.99$ \\
Depth validity & Sky, near, and constant-value pixel ratios
individually $\leq0.25$ \\
Camera clearance & At most two cameras with blocked paths;
view-block ratio $\leq0.6$ \\
Caption & Present, at least 20 characters, and passes style checks \\
Departure cycle & At least one cycle per scheduled camera,
with $\geq8$ frames out and $\geq8$ frames visible \\
Camera smoothness & Jolt $\leq40$\,cm; step $\leq160$\,cm;
rotation $\leq0.08$ times horizontal field of view per frame;
vertical vibration $\leq4$\,cm peak-to-peak at 1.3--3.0\,Hz \\
Motion persistence & Speed $\geq40$\,cm/s in at least 40\%
of frames \\
\bottomrule
\end{tabular}
\end{table}

\noindent\textbf{Annotation quality and selection.}
Urban surface reconstruction is checked against rendered
depth and instance silhouettes, with 90th-percentile
reprojection errors of 0.00\,px for pedestrians and
1.00\,px for vehicles. Transparent and unsupported surface
sections are excluded from this validation. All urban clips
have complete tracks without breaks. Of 793 judged urban
clips, 526 pass the strict profile and 500 are selected.
The hard subset contains 78 pedestrian and 17 vehicle clips,
reflecting a lower acceptance rate for fast-moving vehicles.

\noindent\textbf{Limitations and reproducibility.}
Simulation depth does not reproduce real sensor noise and
has invalid sky and out-of-range regions. Camera counts,
frame rates, and clip lengths differ by domain.
The hard urban subset over-represents pedestrians, and
skeletal point annotations omit independent cloth and hair
motion. Rendering profiles are fixed rather than calibrated
to real camera responses. Seeds are recorded, but physics
settling and live traffic prevent bit-exact regeneration
from seeds alone; recorded per-frame poses provide the
reference for replay.

\subsection{Real-World Benchmark Details}
\label{app:real_bench}

\noindent\textbf{Real-world human benchmark.}
Our real-world benchmark split comprises 100 newly recorded
10-second sequences featuring three participants in total.
Each sequence includes synchronized multi-view RGB videos,
a sequence of reconstructed 3D surfaces, and camera parameters.
To capture realistic everyday activities, we constructed a
$23.1\,\mathrm{m}^2$ home-like space furnished with a sink,
cabinets, drawers, a refrigerator, a sofa, tables, and
household objects.
Participants move through the space, lift and carry objects,
open drawers, and set objects down, producing changes in
both human pose and scene configuration.

\noindent\textbf{Capture system.}
We capture RGB videos using 48 FLIR industrial cameras
at 30 fps and a resolution of $2048\times1536$ pixels.
The camera array combines wide-angle lenses for broad spatial
coverage with standard-field-of-view lenses for capturing detail.
We synchronize image capture across all cameras using an
external hardware trigger.
Before recording, we calibrate camera intrinsics using
Zhang's method~\citep{zhang2000a} and estimate extrinsics
using 16 calibration boards distributed throughout the room.
Specifically, perspective-$n$-point estimates initialize
the relative camera-board poses, which are linked through
a pose graph in a common metric coordinate system. We further refine the calibration through feature matching
and bundle adjustment using COLMAP~\citep{schoenberger2016sfm}.
The calibrated camera parameters are used for image
undistortion, 3D reconstruction, and as model inputs.

\noindent\textbf{Reconstruction and filtering.}
We undistort the images using the calibrated camera parameters
and reconstruct each timestamp independently with
AmbiSuR~\citep{li2026ambisur}, using depth priors from
Depth Anything 3~\citep{lin2025da3}.
For all sequences, we optimize the Gaussians for 2,000 iterations
per timestamp before exporting the surfaces.
To reduce holes in the reconstructed surfaces, we extract surface points by back-projecting rendered depth maps rather than directly exporting Gaussian as surface points.

To reduce floating artifacts, we further filter the
reconstructed 3D points.
For each point, we count the views in which the rendered
depth at its projected image location agrees with its
camera-axis depth within $5\,\mathrm{mm}$.
We retain only points supported by at least two views
and having at least five neighbors within a
$10\,\mathrm{mm}$ radius.
Finally, to make downstream tasks computationally tractable,
we downsample the filtered points using a $1\,\mathrm{mm}$
voxel grid, retaining the point with the greatest view
support in each occupied voxel.

\subsection{Relation to Existing Resources}
\label{app:compare}

Related resources emphasize complementary annotations and tasks.
BEDLAM~2.0~\citep{tesch2026bedlam2} provides human-body
geometry, while ReCamMaster's Multi-Cam
Video dataset~\citep{bai2025recammaster} supports multi-view
re-rendering with camera trajectories.
TAPVid-MV~\citep{koppula2026tapvidmv} evaluates multi-view
point tracking, and
PointMotionBench~\citep{zhang2026molmomotion} evaluates
language-conditioned point-motion forecasting.
Our simulation benchmark pairs future multi-view appearance
with metric surface-point trajectories, camera geometry,
depth, visibility, and language on the same clips.

\clearpage

\section{Additional Experimental Results}

\subsection{Detailed Experiments on \bname{}}
In this section, we discuss the detailed results, varying the evaluation setup under \bname{}.
Specifically, in addition to the all-view Image-to-Video (I2V) setup in \Cref{tab:vp-worldbench}, we evaluate \bname{} under the following setups:
\begin{itemize}[leftmargin=1em, itemsep=2pt, topsep=4pt]
    \item \textbf{All-view I2V} (\Cref{tab:vp-worldbench-full}): We report performance using all views for each split.
    \item \textbf{2-view I2V} (\Cref{tab:vp-worldbench-2view}): We use two input views instead of all.
    \item \textbf{All-view V2V} (\Cref{tab:vp-worldbench-v2v}): We condition on input videos instead of images.
    \item \textbf{Trajectory head} (\Cref{tab:vp-worldbench-ours-head}): We directly measure the performance of \sname{} trajectory head.
\end{itemize}
\vspace{-0.1in}

\begin{table*}[h]
\centering
\caption{\textbf{Results on \bname{} (I2V with all-view).} V and T denote video and point-trajectory supervision. Best results are in \textbf{bold}. The upper bound is obtained by applying the same evaluation pipeline to the ground-truth videos.}
\label{tab:vp-worldbench-full}
\vspace{-0.1in}
\setlength{\tabcolsep}{3pt}
\resizebox{\linewidth}{!}{%
\begin{tabular}{lccccccccccccccccc}
\toprule
& & \multicolumn{8}{c}{Simulation} & \multicolumn{8}{c}{Real-world} \\
\cmidrule(lr){3-10} \cmidrule(lr){11-18}
& & \multicolumn{4}{c}{Robot (500)} & \multicolumn{4}{c}{General (500)}
  & \multicolumn{4}{c}{Robot (100)} & \multicolumn{4}{c}{General (100)} \\
\cmidrule(lr){3-6} \cmidrule(lr){7-10} \cmidrule(lr){11-14} \cmidrule(lr){15-18}
Method & MV
& FVD $\downarrow$ & PSNR $\uparrow$ & MEt3R $\downarrow$ & $\delta^{\text{3D}}_{\text{avg}}$ $\uparrow$
& FVD $\downarrow$ & PSNR $\uparrow$ & MEt3R $\downarrow$ & $\delta^{\text{3D}}_{\text{avg}}$ $\uparrow$
& FVD $\downarrow$ & PSNR $\uparrow$ & MEt3R $\downarrow$ & $\delta^{\text{3D}}_{\text{avg}}$ $\uparrow$
& FVD $\downarrow$ & PSNR $\uparrow$ & MEt3R $\downarrow$ & $\delta^{\text{3D}}_{\text{avg}}$ $\uparrow$ \\
\midrule
HV-1.5 & \xmark
& 98.34 & 20.21 & 0.023 & 41.62 & 115.63 & 11.45 & 0.061 & 23.10
& 426.76 & 16.04 & 0.102 & 36.17 & 105.83 & 20.00 & 0.014 & 21.35 \\
Wan2.2-5B & \xmark
& 315.13 & 15.84 & 0.108 & 29.51 & 205.73 & 10.43 & 0.076 & 18.00
& 550.55 & 14.58 & 0.159 & 29.65 & 115.52 & 16.97 & 0.030 & 16.82 \\
X-WAM & \cmark
& 203.83 & 19.97 & 0.038 & 38.80 & - & - & - & -
& 492.31 & 17.63 & 0.091 & 50.86 & - & - & - & - \\
Cosmos3-N & \xmark
& 104.74 & 20.05 & 0.040 & 39.03 & 167.02 & 10.99 & 0.062 & 22.12
& 326.11 & 16.37 & 0.099 & 41.50 & 66.24 & 20.89 & 0.015 & 20.47 \\
\midrule
\rowcolor{gray!15}
\textbf{\sname{} (V)} & \cmark
& 78.52 & 21.41 & 0.022 & 43.62 & 101.37 & 16.95 & 0.057 & 28.71
& 296.83 & 18.42 & 0.074 & 50.83 & 62.85 & 20.96 & 0.010 & 26.54 \\
\rowcolor{gray!15}
\textbf{\sname{} (V+T)} & \cmark
& \textbf{67.35} & \textbf{22.08} & \textbf{0.016} & \textbf{50.05} & \textbf{81.92} & \textbf{19.16} & \textbf{0.050} & \textbf{36.80}
& \textbf{255.19} & \textbf{20.57} & \textbf{0.064} & \textbf{64.75} & \textbf{56.48} & \textbf{22.10} & \textbf{0.007} & \textbf{33.92} \\
+ camera cond. & \cmark
& 65.34 & 22.75 & 0.014 & 55.05 & 79.33 & 19.39 & 0.046 & 40.04
& 244.84 & 21.03 & 0.062 & 66.75 & 49.77 & 22.45 & 0.005 & 38.46 \\
\midrule
\textit{Upper bound} & --
& -- & -- & 0.004 & 73.27 & -- & -- & 0.029 & 61.87
& -- & -- & 0.010 & 80.40 & -- & -- & 0.003 & 57.98 \\
\bottomrule
\end{tabular}
}
\end{table*}

In \Cref{tab:vp-worldbench-full}, we provide a full table for \Cref{tab:vp-worldbench}, reporting the performance with the camera condition in which the target camera trajectory is given for the forecasting, and the upper bound. We evaluate I2V performance with all-view from each split and follow the same evaluation protocols as \Cref{tab:vp-worldbench}.

\begin{table*}[h]
\centering
\caption{\textbf{Results on \bname{} (I2V with 2-view).} V and T denote video and point-trajectory supervision. Best results are in \textbf{bold}. The upper bound is obtained by applying the same evaluation pipeline to the ground-truth videos.}
\label{tab:vp-worldbench-2view}
\vspace{-0.1in}
\setlength{\tabcolsep}{3pt}
\resizebox{\linewidth}{!}{%
\begin{tabular}{lccccccccccccccccc}
\toprule
& & \multicolumn{8}{c}{Simulation} & \multicolumn{8}{c}{Real-world} \\
\cmidrule(lr){3-10} \cmidrule(lr){11-18}
& & \multicolumn{4}{c}{Robot (500)} & \multicolumn{4}{c}{General (500)}
  & \multicolumn{4}{c}{Robot (100)} & \multicolumn{4}{c}{General (100)} \\
\cmidrule(lr){3-6} \cmidrule(lr){7-10} \cmidrule(lr){11-14} \cmidrule(lr){15-18}
Method & MV
& FVD $\downarrow$ & PSNR $\uparrow$ & MEt3R $\downarrow$ & $\delta^{\text{3D}}_{\text{avg}}$ $\uparrow$
& FVD $\downarrow$ & PSNR $\uparrow$ & MEt3R $\downarrow$ & $\delta^{\text{3D}}_{\text{avg}}$ $\uparrow$
& FVD $\downarrow$ & PSNR $\uparrow$ & MEt3R $\downarrow$ & $\delta^{\text{3D}}_{\text{avg}}$ $\uparrow$
& FVD $\downarrow$ & PSNR $\uparrow$ & MEt3R $\downarrow$ & $\delta^{\text{3D}}_{\text{avg}}$ $\uparrow$ \\
\midrule
HV-1.5 & \xmark
& 85.91 & 20.72 & 0.021 & 39.22 & 141.01 & 11.66 & 0.062 & 20.46
& 426.76 & 16.04 & 0.102 & 36.17 & 117.21 & 20.17 & 0.016 & 19.54 \\
Wan2.2-5B & \xmark
& 294.52 & 16.42 & 0.089 & 27.62 & 180.34 & 9.96 & 0.070 & 17.33
& 550.55 & 14.58 & 0.159 & 29.65 & 151.03 & 16.76 & 0.033 & 16.20 \\
X-WAM & \cmark
& 154.26 & 20.03 & 0.033 & 36.73 & - & - & - & -
& 492.31 & 17.63 & 0.091 & 50.86 & - & - & - & - \\
Cosmos3-N & \xmark
& 86.05 & 20.17 & 0.037 & 34.66 & 202.55 & 10.06 & 0.058 & 19.99
& 326.11 & 16.37 & 0.099 & 41.50 & 88.21 & 20.08 & 0.018 & 19.48 \\
\midrule
\rowcolor{gray!15}
\textbf{\sname{} (V)} & \cmark
& 85.51 & 20.95 & 0.022 & 40.08 & 102.95 & 17.36 & 0.056 & 27.42
& 296.83 & 18.42 & 0.074 & 50.83 & 71.65 & 21.12 & 0.014 & 23.70 \\
\rowcolor{gray!15}
\textbf{\sname{} (V+T)} & \cmark
& \textbf{63.46} & \textbf{21.75} & \textbf{0.015} & \textbf{45.51} & \textbf{84.13} & \textbf{19.68} & \textbf{0.044} & \textbf{33.18}
& \textbf{255.19} & \textbf{20.57} & \textbf{0.064} & \textbf{64.75} & \textbf{57.49} & \textbf{21.58} & \textbf{0.008} & \textbf{31.62} \\
+ camera cond. & \cmark
& 60.14 & 22.22 & 0.014 & 52.42 & 79.07 & 19.91 & 0.043 & 37.42
& 244.84 & 21.03 & 0.062 & 66.75 & 48.76 & 22.29 & 0.007 & 39.38 \\
\midrule
\textit{Upper bound} & --
& -- & -- & 0.003 & 65.98 & -- & -- & 0.026 & 55.40
& -- & -- & 0.010 & 80.40 & -- & -- & 0.003 & 53.03 \\
\bottomrule
\end{tabular}
}
\vspace{-0.1in}
\end{table*}

In \Cref{tab:vp-worldbench-2view}, we provide a table with 2-view I2V setup on \bname{}.

\begin{table*}[h]
\centering
\caption{\textbf{Results on \bname{} (V2V with all-view).} V and T denote video and point-trajectory supervision. Best results are in \textbf{bold}. The upper bound is obtained by applying the same evaluation pipeline to the ground-truth videos.}
\label{tab:vp-worldbench-v2v}
\vspace{-0.1in}
\setlength{\tabcolsep}{3pt}
\resizebox{\linewidth}{!}{%
\begin{tabular}{lccccccccccccccccc}
\toprule
& & \multicolumn{8}{c}{Simulation} & \multicolumn{8}{c}{Real-world} \\
\cmidrule(lr){3-10} \cmidrule(lr){11-18}
& & \multicolumn{4}{c}{Robot (500)} & \multicolumn{4}{c}{General (500)}
  & \multicolumn{4}{c}{Robot (100)} & \multicolumn{4}{c}{General (100)} \\
\cmidrule(lr){3-6} \cmidrule(lr){7-10} \cmidrule(lr){11-14} \cmidrule(lr){15-18}
Method & MV
& FVD $\downarrow$ & PSNR $\uparrow$ & MEt3R $\downarrow$ & $\delta^{\text{3D}}_{\text{avg}}$ $\uparrow$
& FVD $\downarrow$ & PSNR $\uparrow$ & MEt3R $\downarrow$ & $\delta^{\text{3D}}_{\text{avg}}$ $\uparrow$
& FVD $\downarrow$ & PSNR $\uparrow$ & MEt3R $\downarrow$ & $\delta^{\text{3D}}_{\text{avg}}$ $\uparrow$
& FVD $\downarrow$ & PSNR $\uparrow$ & MEt3R $\downarrow$ & $\delta^{\text{3D}}_{\text{avg}}$ $\uparrow$ \\
\midrule
Cosmos3-N & \xmark
& 111.92 & 20.06 & 0.044 & 40.91 & 140.65 & 11.05 & 0.067 & 23.27
& 296.25 & 16.34 & 0.099 & 43.53 & 77.55 & 21.03 & 0.018 & 21.72 \\
+ camera cond. & \xmark
& 110.23 & 19.98 & 0.040 & 43.27 & 145.96 & 10.46 & 0.067 & 19.08
& 288.63 & 17.49 & 0.095 & 44.92 & 72.59 & 21.17 & 0.014 & 22.80 \\
\midrule
\rowcolor{gray!15}
\textbf{\sname{} (V)} & \cmark
& 73.31 & 21.44 & 0.028 & 43.75 & 125.83 & 17.53 & 0.060 & 28.92
& 243.62 & 18.32 & 0.073 & 53.12 & 66.96 & 21.21 & 0.012 & 28.09 \\
\rowcolor{gray!15}
\textbf{\sname{} (V+T)} & \cmark
& \textbf{63.77} & \textbf{22.18} & \textbf{0.021} & \textbf{50.17} & \textbf{81.49} & \textbf{19.36} & \textbf{0.055} & \textbf{38.31}
& \textbf{239.61} & \textbf{20.56} & \textbf{0.064} & \textbf{68.73} & \textbf{48.25} & \textbf{22.11} & \textbf{0.010} & \textbf{34.94} \\
+ camera cond. & \cmark
& 59.56 & 22.56 & 0.018 & 52.57 & 75.58 & 19.81 & 0.053 & 42.81
& 198.44 & 21.16 & 0.060 & 73.60 & 47.69 & 22.29 & 0.007 & 40.25 \\
\midrule
\textit{Upper bound} & --
& -- & -- & 0.004 & 73.27 & -- & -- & 0.029 & 61.87
& -- & -- & 0.010 & 80.40 & -- & -- & 0.003 & 57.98 \\
\bottomrule
\end{tabular}
}
\vspace{-0.1in}
\end{table*}

In \Cref{tab:vp-worldbench-v2v}, we provide a table with all-view Video-to-Video (V2V) setup on \bname{}. Compared to other baselines, Cosmos3-Nano~\citep{nvidia2026cosmos3} is the only model that supports V2V generation. With camera condition, \sname{} surpasses the baseline, \ie, Cosmos3-Nano on every split.

\begin{table*}[tbh]
\centering
\caption{\textbf{Results on \bname{} with \sname{} trajectory head.} V and T denote video and point-trajectory supervision. Best results are in \textbf{bold}.}
\label{tab:vp-worldbench-ours-head}
\vspace{-0.1in}
\setlength{\tabcolsep}{5pt}
\resizebox{0.75\linewidth}{!}{%
\begin{tabular}{lcccccc}
\toprule
& & \multicolumn{2}{c}{Simulation} & \multicolumn{2}{c}{Real-world} \\
\cmidrule(lr){3-4} \cmidrule(lr){5-6}
& & Robot (500) & General (500) & Robot (100) & General (100) \\
\midrule
Method & Tracker
& $\delta^{\text{3D}}_{\text{avg}}$ $\uparrow$
& $\delta^{\text{3D}}_{\text{avg}}$ $\uparrow$
& $\delta^{\text{3D}}_{\text{avg}}$ $\uparrow$
& $\delta^{\text{3D}}_{\text{avg}}$ $\uparrow$ \\
\midrule
\rowcolor{gray!15}
\sname{} (V) & MVTracker
& 43.62 & 28.71 & 50.83 & 26.54 \\
\rowcolor{gray!15}
\sname{} (V+T) & MVTracker
& 50.05 & 36.80 & 64.75 & 33.92 \\
+ camera cond. & MVTracker
& 55.05 & 40.04 & 66.75 & 38.46 \\
\midrule
\rowcolor{gray!15}
\sname{} (V) & Ours
& 45.06 & 29.92 & 51.78 & 28.39 \\
\rowcolor{gray!15}
\sname{} (V+T) & Ours
& \textbf{52.15} & \textbf{38.46} & \textbf{67.22} & \textbf{36.16} \\
+ camera cond. & Ours
& 56.81 & 42.82 & 69.17 & 39.41 \\
\bottomrule
\end{tabular}
}
\vspace{-0.1in}
\end{table*}

\sname{} has a trajectory head that predicts the positions of future 3D points. Unlike \Cref{tab:vp-worldbench}, where point tracks are extracted from MVTracker~\citep{rajivc2025mvtracker}, here we report the quality of point tracks directly predicted by our trajectory head.
As shown in \Cref{tab:vp-worldbench-ours-head}, direct prediction consistently improves 3D tracking accuracy, \eg, $\delta^{\text{3D}}_{\text{avg}}$ of \sname{} (V+T) increases from 64.75 to 67.22 on Real-world Robot.

For all variations, \sname{} (V+T) achieves state-of-the-art performance on I2V and V2V, indicating that the supervision of point trajectories consistently enhances the multi-view video fidelity.

\subsection{Detailed Experiments on 4D Generation}
\label{app:upper_bounds}

\begin{table*}[h]
\centering
\begin{minipage}[t]{0.54\linewidth}
\centering
\caption{\textbf{Results of 4D generation on ReCamMaster and CMU Panoptic.}}
\label{tab:4d-recamandcmu-upper}
\vspace{-0.07in}
\setlength{\tabcolsep}{3pt}
\resizebox{\linewidth}{!}{%
\begin{tabular}{lccccccc}
\toprule
& & \multicolumn{2}{c}{ReCamMaster (100)} & \multicolumn{4}{c}{CMU Panoptic (100)} \\
\cmidrule(lr){3-4} \cmidrule(lr){5-8}
Method & MV
& FVD $\downarrow$ & PSNR $\uparrow$
& FVD $\downarrow$ & PSNR $\uparrow$ & MEt3R $\downarrow$ & $\delta^{\text{3D}}_{\text{avg}}$ $\uparrow$ \\
\midrule
HV-1.5 & \xmark
& 183.48 & \underline{14.03}
& 352.15 & \underline{21.60} & \underline{0.041} & \underline{31.67} \\
Wan2.2-5B & \xmark
& 457.16 & 12.10
& 258.06 & 18.65 & 0.061 & 27.88 \\
Cosmos3-N & \xmark
& \underline{125.70} & 13.99
& \underline{162.36} & 20.72 & 0.060 & 30.12 \\
\midrule
\rowcolor{gray!15}
\textbf{\sname (V)} & \cmark
& 103.12 & 15.09
& 150.63 & 20.85 & 0.037 & 34.72 \\
\rowcolor{gray!15}
\textbf{\sname (V+T)} & \cmark
& \textbf{79.63} & \textbf{17.06}
& \textbf{128.64} & \textbf{23.24} & \textbf{0.029} & \textbf{42.03} \\
+ camera cond. & \cmark
& 78.03 & 17.60
& 120.23 & 23.49 & 0.026 & 44.61 \\
\midrule
\textit{Upper bound} & --
& -- & --
& -- & -- & 0.009 & 64.12 \\
\bottomrule
\end{tabular}
}
\end{minipage}
\hfill
\begin{minipage}[t]{0.44\linewidth}
\centering
\caption{\textbf{Results of 4D generation on RoboCasa Kitchen.}}
\label{tab:4d-robocasakitchen-upper}
\vspace{-0.07in}
\setlength{\tabcolsep}{3pt}
\resizebox{\linewidth}{!}{%
\begin{tabular}{lcccccc}
\toprule
 && \multicolumn{3}{c}{RGB} & \multicolumn{2}{c}{Geometry} \\
\cmidrule(lr){3-5} \cmidrule(lr){6-7}
Method & MV & PSNR$\uparrow$ & SSIM$\uparrow$ & LPIPS$\downarrow$ & MEt3R$\downarrow$ & $\delta^{\text{3D}}_{\text{avg}}\uparrow$ \\
\midrule
DreamZero & \xmark & 20.47 & 0.7612 & 0.1651 & 0.061 & 30.83 \\
Robot4DGen & \cmark & 22.05 & 0.8035 & 0.1078 & 0.025 & 46.96 \\
X-WAM & \cmark & 22.41 & 0.8729 & 0.0571 & 0.042 & 42.10 \\
\midrule
\rowcolor{gray!15}
\textbf{\sname (V)} & \cmark & 22.61 & 0.8752 & 0.0558 & 0.033 & 46.94 \\
\rowcolor{gray!15}
\textbf{\sname (V+T)} & \cmark & \textbf{22.83} & \textbf{0.8801} & \textbf{0.0537} & \textbf{0.020} & \textbf{53.70} \\
+ camera cond. & \cmark & 22.10 & 0.8835 & 0.0501 & 0.013 & 58.48 \\
\midrule
\textit{Upper bound} & -- & -- & -- & -- & 0.004 & 78.88 \\
\bottomrule
\end{tabular}
}
\end{minipage}\vspace{-10pt}
\end{table*}

In Tables~\ref{tab:4d-recamandcmu-upper} and~\ref{tab:4d-robocasakitchen-upper}, we provide detailed results for Tables~\ref{tab:4d-recamandcmu} and~\ref{tab:4d-robocasakitchen}, including the upper bounds of cross-view MEt3R and $\delta^{\text{3D}}_{\text{avg}}$ on CMU Panoptic and RoboCasa Kitchen.

\clearpage

\begin{table}[t]
\centering
\caption{\textbf{Per-frame geometry vs.\ 3D point trajectory supervision.} All variants share the backbone, data, feature layer, and training schedule of VPM (V+T). Per-frame variants are supervised with the same 3D point positions at every frame without correspondence across frames.}
\label{tab:per_frame_geo}
\small
\setlength{\tabcolsep}{4.2pt}
\begin{tabular}{l cccccccc}
\toprule
 & \multicolumn{4}{c}{Simulation} & \multicolumn{4}{c}{Real-world} \\
\cmidrule(lr){2-5} \cmidrule(lr){6-9}
 & \multicolumn{2}{c}{Robot (500)} & \multicolumn{2}{c}{General (500)} & \multicolumn{2}{c}{Robot (100)} & \multicolumn{2}{c}{General (100)} \\
\cmidrule(lr){2-3} \cmidrule(lr){4-5} \cmidrule(lr){6-7} \cmidrule(lr){8-9}
Supervision & FVD$\downarrow$ & $\delta^{\text{3D}}_{\text{avg}}\!\uparrow$ & FVD$\downarrow$ & $\delta^{\text{3D}}_{\text{avg}}\!\uparrow$ & FVD$\downarrow$ & $\delta^{\text{3D}}_{\text{avg}}\!\uparrow$ & FVD$\downarrow$ & $\delta^{\text{3D}}_{\text{avg}}\!\uparrow$ \\
\midrule
None & 78.52 & 43.62 & 101.37 & 28.71 & 296.83 & 50.83 & 62.85 & 26.54 \\
\midrule
\multicolumn{9}{c}{\textit{Per-frame geometry (no temporal identity)}} \\
Depth & 77.07 & 45.21 & 100.02 & 29.88 & 285.13 & 53.73 & 62.08 & 26.98 \\
Pointmap & 75.48 & 44.90 & 99.35 & 29.69 & 281.67 & 54.50 & 61.37 & 29.46 \\
\midrule
\multicolumn{9}{c}{\textit{3D point trajectory (temporal identity)}} \\
w/o $\mathcal{L}_{\text{corr}}$ & \underline{69.88} & \underline{48.54} & \underline{85.73} & \underline{31.54} & \underline{271.01} & \underline{59.94} & \underline{59.33} & \underline{30.75} \\
\rowcolor{gray!15} Full & \textbf{67.35} & \textbf{50.05} & \textbf{81.92} & \textbf{36.80} & \textbf{255.19} & \textbf{64.75} & \textbf{56.48} & \textbf{33.92} \\
\bottomrule
\end{tabular}
\vspace{-0.1in}
\end{table}

\subsection{Comparison with Per-Frame Geometry Supervision}
\label{app:per_frame_geo}

\noindent\textbf{Setup.}
To compare per-frame geometry supervision with ours, we replace
the trajectory head of \sname{} (V+T) with a per-frame geometry
head. Both heads are supervised at the pixels where the annotated
3D points project in each frame, using their depth or 3D position.
The per-frame head thus receives the same geometric supervision,
but without linking points across frames: it predicts geometry at
given pixel locations, whereas our head must predict where each
queried point moves.

\noindent{\bf Results.}
As shown in Table~\ref{tab:per_frame_geo}, per-frame geometry supervision improves over VPM (V) on every split, recovering only 19\% (depth) and 28\% (pointmap) of the FVD improvement of VPM (V+T).
Pointmaps improve over depth maps in FVD on every split, while $\delta^{\text{3D}}_{\text{avg}}$ is comparable, with depth slightly higher in simulation.
Even without $\mathcal{L}_{\text{corr}}$, 3D trajectories outperform both on every split while supervising the same 3D positions at every frame.
We therefore attribute this gain to temporal identity, which additionally constrains the motion of each surface point across frames.
The gap is largest on the dynamic splits, \eg, a $\delta^{\text{3D}}_{\text{avg}}$ of 54.50 vs.\ 59.94 on Real-world Robot and an FVD of 99.35 vs.\ 85.73 on Simulation General, where points move substantially and are often occluded.

\begin{table*}[h]
\centering
\caption{\textbf{Results on 3D point forecasting.} We evaluate the models on PointMotionBench~\citep{zhang2026molmomotion}. Best results are in \textbf{bold}, second best are \underline{underlined}.}
\label{tab:3d_point_head}
\setlength{\tabcolsep}{4pt}
\newcommand{\gc}{\cellcolor{gray!15}}
\resizebox{\linewidth}{!}{%
\begin{tabular}{llcccccccccccc}
\toprule
\multirow{2}{*}{Paradigm} & \multirow{2}{*}{Model}
& \multicolumn{2}{c}{Inputs} & \multicolumn{3}{c}{HOT3D} & \multicolumn{3}{c}{WorldTrack} & \multicolumn{3}{c}{DAVIS} \\
\cmidrule(lr){3-4} \cmidrule(lr){5-7} \cmidrule(lr){8-10} \cmidrule(lr){11-13}
& & Frames & Text
& ADE$\downarrow$ & FDE$\downarrow$ & PWT$\uparrow$
& ADE$\downarrow$ & FDE$\downarrow$ & PWT$\uparrow$
& ADE$\downarrow$ & FDE$\downarrow$ & PWT$\uparrow$ \\
\toprule
\multirow{2}{*}{\shortstack{Non-\\parametric}}
& Static      & 1 & \xmark & 0.180 & 0.316 & 0.293 & 0.167 & 0.317 & 0.390 & 2.281 & 4.360 & 0.085 \\
& Extrapolate & 3 & \xmark & 0.159 & 0.309 & 0.351 & 0.184 & 0.432 & 0.436 & 2.683 & 5.741 & 0.104 \\
\midrule
\multirow{2}{*}{Pixel-space}
& Wan2.2-5B      & 1 & \cmark & 0.200 & 0.308 & 0.253 & 0.852 & 1.046 & 0.090 & 3.074 & 5.192 & 0.051 \\
& Cosmos-Predict & 5 & \cmark & 0.225 & 0.294 & 0.199 & 0.831 & 0.988 & 0.072 & 4.191 & 6.368 & 0.033 \\
\midrule
\multirow{3}{*}{3D model}
& ObjectForesight & 3 & \xmark & 0.129 & \underline{0.192} & 0.353 & -- & -- & -- & -- & -- & -- \\
& EgoScaler       & 1 & \cmark & 0.170 & \textbf{0.179} & 0.218 & -- & -- & -- & -- & -- & -- \\
& Robot4DGen      & 3 & \xmark & 0.212 & 0.271 & 0.112 & 0.548 & 0.704 & 0.121 & 2.120 & 3.382 & 0.081 \\
\midrule
2D track
& Track2Act & 1 & \xmark & 0.294 & 0.413 & 0.202 & 1.230 & 1.567 & 0.053 & 4.853 & 8.110 & 0.018 \\
\midrule
\multirow{5}{*}{3D track}
& MolmoMotion-FM & 1 & \cmark & 0.183 & 0.311 & 0.286 & 0.165 & 0.305 & 0.401 & 1.380 & 2.205 & 0.165 \\
& MolmoMotion-FM & 3 & \cmark & 0.135 & 0.255 & 0.382 & 0.158 & 0.295 & 0.438 & 1.480 & 2.520 & 0.130 \\
& MolmoMotion-AR & 1 & \cmark & 0.157 & 0.290 & 0.303 & 0.148 & 0.269 & 0.424 & \underline{1.146} & \underline{1.843} & \textbf{0.199} \\
& MolmoMotion-AR & 3 & \cmark & \underline{0.109} & 0.217 & \textbf{0.444} & \underline{0.143} & \underline{0.261} & \underline{0.445} & 1.227 & 2.108 & 0.153 \\
& \gc\textbf{\sname{} (ours)} & \gc 3 & \gc\cmark
& \gc\textbf{0.098} & \gc 0.209 & \gc\underline{0.396}
& \gc\textbf{0.129} & \gc\textbf{0.248} & \gc\textbf{0.470}
& \gc\textbf{1.124} & \gc\textbf{1.804} & \gc\underline{0.171} \\
\bottomrule
\end{tabular}
}
\vspace{-0.1in}
\end{table*}

\subsection{Performance of 3D Point Trajectory Head}
\label{app:own_head}
To evaluate our 3D point trajectory head, we test various methods on PointMotionBench~\citep{zhang2026molmomotion}, a 3D point
forecasting benchmark. As shown in \Cref{tab:3d_point_head}, \sname{} achieves
the lowest ADE on HOT3D~\citep{banerjee2024hot3d} and the best results on all three metrics on WorldTrack~\citep{feng2025st4rtrack}. 
Compared with MolmoMotion-AR under the same three-frame input, \sname{} lowers both ADE and FDE on every dataset and improves PWT on WorldTrack and DAVIS~\citep{pont2017davis}.
These results highlight that our trajectory head produces consistent 3D tracks across diverse domains, from egocentric manipulation to general scenes.

\begin{table}[h]
\centering
\caption{\textbf{Results on single-view I2V world modeling.}}
\label{tab:paibench_rbench}
\setlength{\aboverulesep}{0pt}
\setlength{\belowrulesep}{0pt}
\renewcommand{\arraystretch}{1.1}
\resizebox{0.65\linewidth}{!}{%
\begin{tabular}{lcccc}
\toprule
& \multicolumn{3}{c}{PAIBench-G ($\uparrow$)} & RBench ($\uparrow$) \\
\cmidrule(lr){2-4} \cmidrule(lr){5-5}
Model & Overall & Domain & Quality & Score (\%) \\
\midrule
Wan2.2-A14B            & 81.3 & 85.3 & 77.3 & 50.7 \\
HunyuanVideo-1.5       & 81.7 & 85.9 & 77.6 & 46.0 \\
Cosmos-Predict2.5-2B   & 81.2 & 84.6 & 77.9 & 46.4 \\
Cosmos-Predict2.5-14B  & 81.1 & 84.0 & \underline{78.1} & - \\
Wan2.1-14B             & 80.2 & 83.6 & 76.8 & - \\
Wan2.2-5B              & 81.0 & 84.6 & 77.4 & - \\
\midrule
Veo-3.1                & 82.6 & \underline{87.6} & 77.6 & 56.3 \\
Seedance-1.5-Pro       & 80.8 & 84.7 & 76.9 & 58.4 \\
Wan2.6                & 81.9 & 85.9 & 77.8 & \textbf{60.7} \\
\midrule
Cosmos3-Super          & \underline{82.8} & 87.3 & \textbf{78.2} & 58.1 \\
Cosmos3-Nano           & 82.7 & 87.2 & \underline{78.1} & 58.4 \\
\rowcolor{gray!15}
\textbf{\sname{} (ours)} & \textbf{83.0} & \textbf{87.9} & \textbf{78.2} & \underline{60.1} \\
\bottomrule
\end{tabular}
}
\end{table}

\subsection{Performance on Single-View World Modeling}

\noindent{\bf I2V world modeling.}
We evaluate image-to-video (I2V) generation on PAIBench-G~\citep{zhou2026pai} and RBench~\citep{deng2026rbench}, where
each model predicts future frames from a single-view image. As shown in
\Cref{tab:paibench_rbench}, \sname{} achieves the best Overall and Domain scores
on PAIBench-G and ties for the best Quality score, outperforming both
open-source models and proprietary models such as Veo-3.1. On RBench, \sname{} ranks
second, trailing only Wan2.6 by 0.6 points.

\begin{table}[h]
\centering
\caption{\textbf{Results on single-view V2V world modeling.}}
\label{tab:v2v_monocular}
\begin{tabular}{l|cc}
\toprule
Model & Mode & Score ($\uparrow$) \\
\midrule
VideoPoet     & V2V & 29.5 \\
VideoGPT     & V2V & 35.0 \\
Magi-1        & V2V & \underline{56.0} \\
Cosmos3-Super & V2V & \textbf{59.7} \\
Cosmos3-Nano  & V2V & 50.2 \\
\sname{} (ours) & V2V & 52.1 \\
\bottomrule
\end{tabular}
\end{table}

\noindent{\bf V2V world modeling.}
We further evaluate video-to-video (V2V) generation, where each model
continues a single-view input video. As shown in \Cref{tab:v2v_monocular},
\sname{} reaches a score of 52.1, outperforming VideoPoet~\citep{kondratyuk2023videopoet}, VideoGPT~\citep{yan2021videogpt}, and Cosmos3-Nano while remaining behind Magi-1~\citep{teng2025magi} and Cosmos3-Super. These results demonstrate that \sname{} remains competitive even in the single-view setting.

\begin{table*}[h]
\centering
\caption{\textbf{Modality-wise ablation on ReCamMaster and CMU Panoptic.} V, T, and A denote
video, point-trajectory, and action supervision. The upper bound applies the evaluation pipeline to the ground-truth videos.}
\label{tab:analysis_modality_recam_panoptic}
\setlength{\tabcolsep}{4pt}
\resizebox{0.7\linewidth}{!}{%
\begin{tabular}{lccccccc}
\toprule
& & \multicolumn{2}{c}{ReCamMaster (100)} & \multicolumn{4}{c}{CMU Panoptic (100)} \\
\cmidrule(lr){3-4} \cmidrule(lr){5-8}
Method & MV
& FVD $\downarrow$ & PSNR $\uparrow$
& FVD $\downarrow$ & PSNR $\uparrow$ & MEt3R $\downarrow$ & $\delta^{\text{3D}}_{\text{avg}}$ $\uparrow$ \\
\midrule
Cosmos3-Nano & \xmark
& 125.70 & 13.99
& 162.36 & 20.72 & 0.060 & 30.12 \\
\sname (V) & \cmark
& 103.12 & 15.09
& 150.63 & 20.85 & 0.037 & 34.72 \\
\sname (V+A) & \cmark
& 101.35 & 15.63
& 149.06 & 21.14 & 0.035 & 36.51 \\
\sname (V+T) & \cmark
& 79.63 & 17.06
& 128.64 & 23.24 & 0.029 & 42.03 \\
\sname (V+T+A) & \cmark
& 76.12 & 17.48
& 118.05 & 24.08 & 0.029 & 44.33 \\
+ camera cond. & \cmark
& 73.44 & 18.01
& 113.40 & 24.25 & 0.027 & 45.70 \\
\midrule
\textit{Upper bound} & --
& -- & --
& -- & -- & 0.009 & 64.12 \\
\bottomrule
\end{tabular}
}
\end{table*}

\subsection{Additional Results on Modality-Wise Ablation Study}
We repeat the modality-wise ablation of \Cref{tab:analysis_modality} on the zero-shot 4D generation benchmarks of \Cref{exp:zero_shot_4d_gen}, ReCamMaster and CMU Panoptic, which contain no robots and are not used for training.
As shown in \Cref{tab:analysis_modality_recam_panoptic}, the trends on \bname{} carry over to these out-of-domain scenes.
Adding point trajectories to video yields the largest gain, reducing FVD on ReCamMaster from 103.12 to 79.63 and raising $\delta^{\text{3D}}_{\text{avg}}$ on CMU Panoptic from 34.72 to 42.03, whereas adding actions yields only 101.35 and 36.51, respectively.
Action supervision nonetheless improves or matches every metric, both on its own and on top of point trajectories (\eg, $+2.30$ vs.\ $+7.31$ $\delta^{\text{3D}}_{\text{avg}}$ from trajectories), indicating that it complements point supervision even without robots in the scene.
Adding camera conditioning to \sname{} (V+T+A) further improves all metrics, achieving the best results on both benchmarks.

\begin{figure}[t]
    \centering
    \centerline{\includegraphics[width=\textwidth]{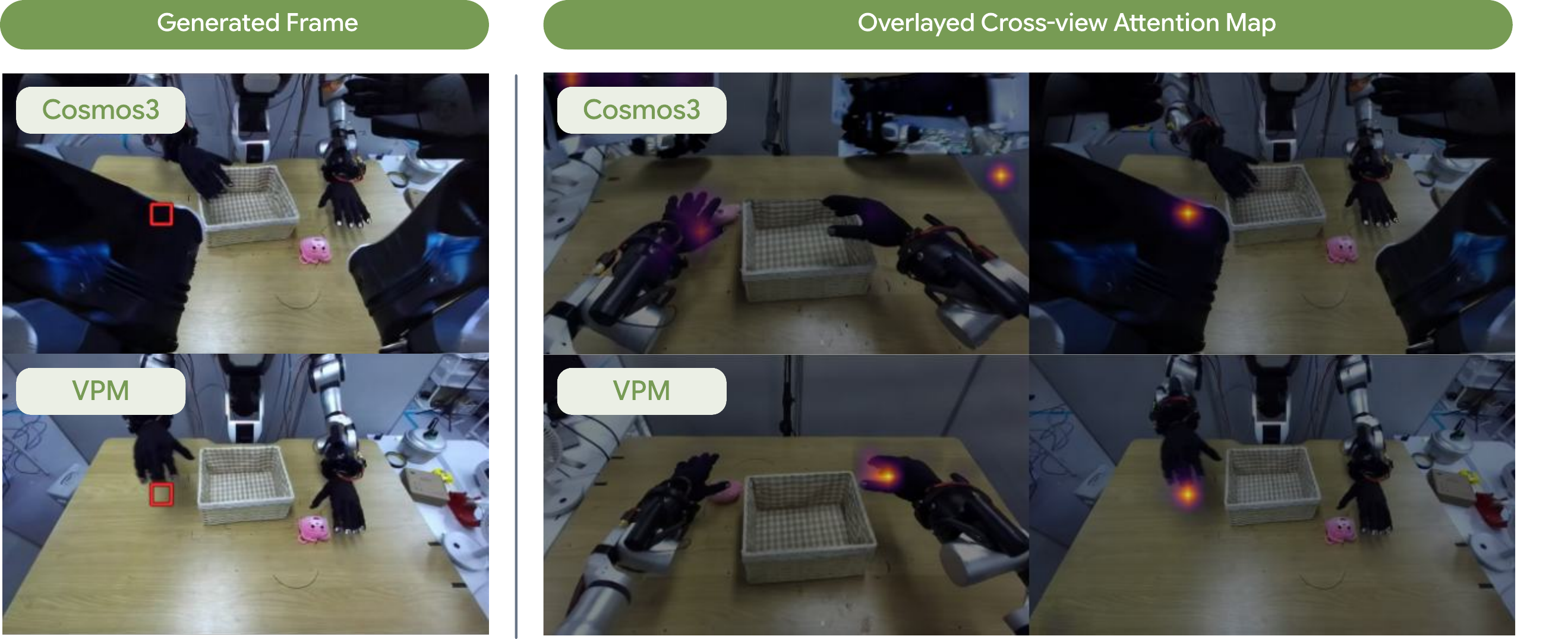}}
    \vspace{-0.05in}
   \caption{\textbf{Visualization of cross-view attention map.}}
   \label{fig:cross_view_attn}
   %\vspace{-15pt}
\end{figure}

\subsection{Analysis on Multi-View Video Generation}
\label{app:multi-view-attn}
To examine how \sname{} maintains consistency across views, we visualize its cross-view attention in \Cref{fig:cross_view_attn}. We select a query grid in one generated view (red box) and overlay its attention map on the other views.
Since Cosmos3-Nano does not natively take multi-view input, we concatenate the views into a single image. For Cosmos3-Nano, the attention scatters over the background and unrelated parts of the robot, and the arm in its generated frame is visibly distorted. 
In contrast, the attention of \sname{} lands on the same hand in each of the other views, matching the
location of the query grid. 
This highlights that \sname{} learns the cross-view correspondence~\citep{an2025cross,hong2024pf3plat,hong2024unifying,hong2021deep,hong2024unifying2,cho2021cats,cho2022cats++} and leverages it to generate consistent multi-view videos.

\begin{table}[h]
\centering
\caption{\textbf{Clock-time analysis.} Wall-clock time (sec) to generate one clip of 64 frames at 480p per view on one GH200 GPU. Cosmos3-Nano generates the views independently, whereas \sname{} generates them jointly. Parentheses show the time relative to Cosmos3-Nano with the same number of views.}
\label{tab:clock_time}
\small
\setlength{\tabcolsep}{5pt}
\begin{tabular}{llccc}
\toprule
Method & Output & 1 view & 2 views & 4 views \\
\midrule
Cosmos3-Nano & Video & 15.9 & 31.8 & 63.5 \\
VPM (V) & Video & 16.3 {\scriptsize(1.03$\times$)} & 38.7 {\scriptsize(1.22$\times$)} & 101.4 {\scriptsize(1.60$\times$)} \\
VPM (V+T) & Video + 3D traj. & 16.5 {\scriptsize(1.04$\times$)} & 39.1 {\scriptsize(1.23$\times$)} & 102.3 {\scriptsize(1.61$\times$)} \\
VPM (V+T+A) & Video + 3D traj. + action & 16.6 {\scriptsize(1.04$\times$)} & 39.2 {\scriptsize(1.23$\times$)} & 102.5 {\scriptsize(1.61$\times$)} \\
\bottomrule
\end{tabular}
\vspace{-10pt}
\end{table}

\subsection{Clock-Time Analysis}
\noindent{\bf Setup.}
We measure the wall-clock time to generate one clip under the I2V protocol of Table~\ref{tab:vp-worldbench}, \ie, 64 future frames at 480p per view from a single conditioning frame, on one NVIDIA GH200 GPU with BF16 precision and batch size 1.
\sname{} (V+T+A) additionally denoises 16 action chunks for action generation.
All models use 35 denoising steps with classifier-free guidance, following the default sampler of Cosmos3. 
The reported time includes VAE decoding.
As in our evaluation, Cosmos3-Nano generates the views one at a time, whereas ours generates all views jointly in a single token sequence.

\noindent{\bf Results.}
For four views, decoding trajectories and actions adds only 0.9\,s and 0.2\,s, respectively, and with a single view, camera conditioning makes VPM only 3\% slower than Cosmos3-Nano.
The views can attend to one another, which contributes to the gains over Cosmos3-Nano.

\begin{figure}[t]
  \centering
  \includegraphics[width=\linewidth]{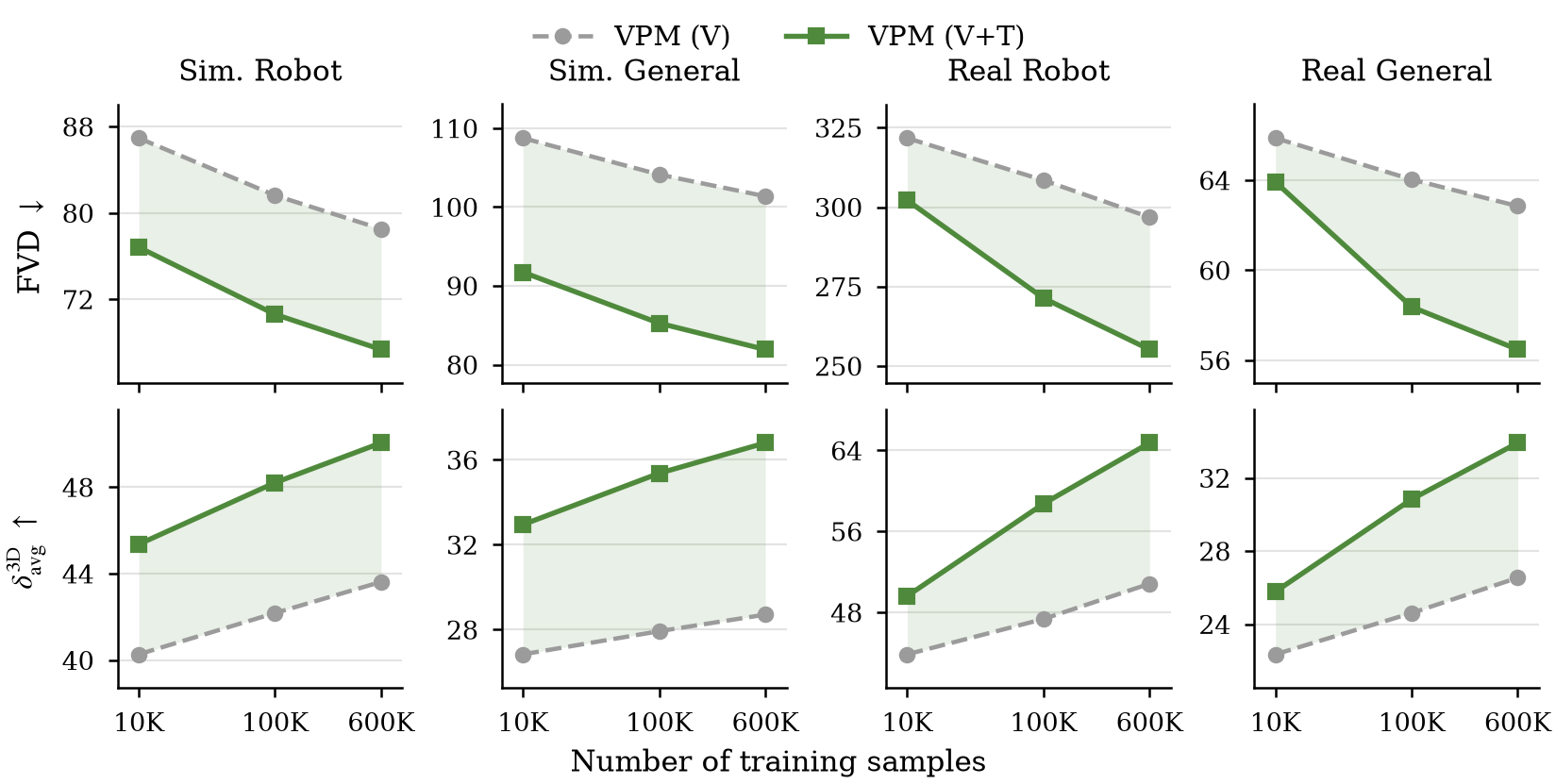}
  \caption{\textbf{Data scalability.} FVD (top) and $\delta^{\text{3D}}_{\text{avg}}$ (bottom) of \sname{} (V) and \sname{} (V+T) trained on 10K, 100K, and 600K samples. The shaded area marks the gain from point supervision.}
  \label{fig:data_scaling}
\end{figure}

\subsection{Data Scalability}
\label{app:data_scaling}

\noindent{\bf Setup.}
We train \sname{} (V) and \sname{} (V+T) on 10K, 100K, and 600K samples.
The subsets are random subsamples of the full data  that preserve their proportions in \Cref{tab:train_data}.
At each scale, both models use the same training schedule.

\noindent{\bf Results.}
As shown in \Cref{fig:data_scaling}, both models improve steadily with more training data, and point supervision improves over the video-only model at every scale.
The gap widens as the data grows on every split. Specifically, the $\delta^{\text{3D}}_{\text{avg}}$ gap on Real-world Robot grows from 5.76 at 10K to 13.92 at 600K, as the robot datasets add trajectory supervision from real robot episodes.
\sname{} (V+T) trained on 100K samples already outperforms \sname{} (V) trained on all 600K samples on every split, \eg, 70.61 vs.\ 78.52 FVD on Simulation Robot, indicating that point supervision is a more data-efficient way to improve forecasting than scaling video data alone.

\clearpage

\section{Detailed Ablation Studies}
\label{app:detailed_ablation}

In this section, we ablate the design choices of \sname{}.
Unless otherwise stated, each variant starts from \sname{} (V+T) and changes a single component, keeping the training data, the three-stage schedule, and fixed iterations in Table~\ref{tab:hyperparams}.
All variants are evaluated on VP-WorldBench with the I2V protocol of Table~\ref{tab:vp-worldbench}, and we report FVD for video fidelity and $\delta^{\text{3D}}_{\text{avg}}$ for 3D motion accuracy.

\begin{table}[h]
\centering
\caption{\textbf{Ablation on camera conditioning.} Add.\ and Rot.\ denote the additive Pl\"{u}cker embedding and the camera-dependent rotation of queries and keys in \Cref{eq:camera}, and View denotes the view axis of the 4D RoPE in \Cref{eq:rope4d}.}
\label{tab:abl_camera}
\small
\setlength{\tabcolsep}{4.2pt}
\begin{tabular}{ccc cccccccc}
\toprule
 &  &  & \multicolumn{4}{c}{Simulation} & \multicolumn{4}{c}{Real-world} \\
\cmidrule(lr){4-7} \cmidrule(lr){8-11}
\multicolumn{3}{c}{Camera cond.} & \multicolumn{2}{c}{Robot (500)} & \multicolumn{2}{c}{General (500)} & \multicolumn{2}{c}{Robot (100)} & \multicolumn{2}{c}{General (100)} \\
\cmidrule(lr){1-3} \cmidrule(lr){4-5} \cmidrule(lr){6-7} \cmidrule(lr){8-9} \cmidrule(lr){10-11}
Add. & Rot. & View & FVD$\downarrow$ & $\delta^{\text{3D}}_{\text{avg}}\!\uparrow$ & FVD$\downarrow$ & $\delta^{\text{3D}}_{\text{avg}}\!\uparrow$ & FVD$\downarrow$ & $\delta^{\text{3D}}_{\text{avg}}\!\uparrow$ & FVD$\downarrow$ & $\delta^{\text{3D}}_{\text{avg}}\!\uparrow$ \\
\midrule
\xmark & \xmark & \xmark & 91.22 & 33.29 & 140.06 & 24.64 & 299.51 & 45.79 & 67.80 & 23.94 \\
\xmark & \xmark & \cmark & 75.17 & 41.70 & 128.41 & 29.08 & 281.67 & 49.25 & 61.43 & 26.99 \\
\cmark & \xmark & \cmark & 70.70 & 48.18 & \underline{84.60} & \underline{33.94} & 268.34 & 59.00 & 59.01 & 30.05 \\
\cmark & \cmark & \xmark & \underline{70.01} & \underline{48.59} & 85.15 & 33.40 & \underline{265.37} & \underline{61.88} & \underline{58.90} & \underline{31.55} \\
\rowcolor{gray!15} \cmark & \cmark & \cmark & \textbf{67.35} & \textbf{50.05} & \textbf{81.92} & \textbf{36.80} & \textbf{255.19} & \textbf{64.75} & \textbf{56.48} & \textbf{33.92} \\
\bottomrule
\end{tabular}
\vspace{-0.1in}
\end{table}

\subsection{Ablation Study for Camera Embeddings}
\label{app:abl_camera}

\noindent{\bf Setup.}
\sname{} conditions the generator on camera geometry through three components (Section~\ref{sec:camera}): the additive Pl\"{u}cker embedding $\mathbf{W}\mathbf{e}_i$ (Add.) and the camera-dependent rotation of queries and keys $\mathbf{R}(\psi(\mathbf{e}_i))$ (Rot.) in \Cref{eq:camera}, and the view axis of the 4D RoPE in \Cref{eq:rope4d} (View).
Table~\ref{tab:abl_camera} removes them individually and jointly.

\noindent{\bf Results.}
Removing all three components degrades every split of VP-WorldBench, most severely on Simulation General (FVD 140.06 vs.\ 81.92), whose cameras move during each clip.
The view axis alone partially recovers performance, and adding the Pl\"{u}cker embedding further reduces FVD on Simulation General from 128.41 to 84.60.
The camera-dependent rotation improves both metrics on every split, most notably $\delta^{\text{3D}}_{\text{avg}}$ on Real-world Robot (59.00 $\rightarrow$ 64.75).
Removing only the view axis raises FVD by approximately 4\% across all splits.

\begin{table}[h]
\centering
\caption{\textbf{Ablation on training objectives.} $\mathcal{L}_{\text{FM}}$ is used in all rows, and $\mathcal{L}_{\text{conf}}$ is used together with $\mathcal{L}_{\text{seq}}$. The first row corresponds to \sname{} (V) in Table~\ref{tab:vp-worldbench}.}
\label{tab:abl_objective}
\small
\setlength{\tabcolsep}{4.2pt}
\begin{tabular}{ccc cccccccc}
\toprule
 &  &  & \multicolumn{4}{c}{Simulation} & \multicolumn{4}{c}{Real-world} \\
\cmidrule(lr){4-7} \cmidrule(lr){8-11}
\multicolumn{3}{c}{Objective} & \multicolumn{2}{c}{Robot (500)} & \multicolumn{2}{c}{General (500)} & \multicolumn{2}{c}{Robot (100)} & \multicolumn{2}{c}{General (100)} \\
\cmidrule(lr){1-3} \cmidrule(lr){4-5} \cmidrule(lr){6-7} \cmidrule(lr){8-9} \cmidrule(lr){10-11}
$\mathcal{L}_{\text{seq}}$ & $\mathcal{L}_{\text{depth}}$ & $\mathcal{L}_{\text{corr}}$ & FVD$\downarrow$ & $\delta^{\text{3D}}_{\text{avg}}\!\uparrow$ & FVD$\downarrow$ & $\delta^{\text{3D}}_{\text{avg}}\!\uparrow$ & FVD$\downarrow$ & $\delta^{\text{3D}}_{\text{avg}}\!\uparrow$ & FVD$\downarrow$ & $\delta^{\text{3D}}_{\text{avg}}\!\uparrow$ \\
\midrule
\xmark & \xmark & \xmark & 78.52 & 43.62 & 101.37 & 28.71 & 296.83 & 50.83 & 62.85 & 26.54 \\
\xmark & \xmark & \cmark & 78.44 & 42.96 & 100.45 & 29.90 & 284.17 & 52.00 & 61.57 & 27.50 \\
\cmark & \xmark & \xmark & 74.82 & 43.90 & 91.92 & 30.00 & 275.24 & 54.69 & 59.71 & 27.99 \\
\cmark & \cmark & \xmark & \underline{69.88} & \underline{48.54} & \underline{85.73} & \underline{31.54} & \underline{271.01} & \underline{59.94} & \underline{59.33} & \underline{30.75} \\
\rowcolor{gray!15} \cmark & \cmark & \cmark & \textbf{67.35} & \textbf{50.05} & \textbf{81.92} & \textbf{36.80} & \textbf{255.19} & \textbf{64.75} & \textbf{56.48} & \textbf{33.92} \\
\bottomrule
\end{tabular}
\vspace{-0.1in}
\end{table}

\subsection{Ablation Study for Objectives}
\label{app:abl_objective}

\noindent{\bf Setup.}
We ablate the point-supervision objectives: the 2D position loss $\mathcal{L}_{\text{seq}}$ and the depth loss $\mathcal{L}_{\text{depth}}$ of the trajectory head in \Cref{eq:track}, and the cross-view correspondence loss $\mathcal{L}_{\text{corr}}$ in \Cref{eq:corr}.
$\mathcal{L}_{\text{FM}}$ is always used, and $\mathcal{L}_{\text{conf}}$ is kept whenever $\mathcal{L}_{\text{seq}}$ is used, since its target depends on the predicted 2D position.
The first row of \Cref{tab:abl_objective}, trained with $\mathcal{L}_{\text{FM}}$ only, corresponds to \sname{} (V) in Table~\ref{tab:vp-worldbench}.

\noindent{\bf Results.}
Combining all objectives performs best on every split.
$\mathcal{L}_{\text{corr}}$ alone has limited effect, except for FVD on Real-world Robot (296.83 $\rightarrow$ 284.17).
$\mathcal{L}_{\text{seq}}$ mainly improves FVD (\eg, 101.37 $\rightarrow$ 91.92 on Simulation General), and adding $\mathcal{L}_{\text{depth}}$ substantially raises $\delta^{\text{3D}}_{\text{avg}}$ (\eg, 54.69 $\rightarrow$ 59.94 on Real-world Robot).
Adding $\mathcal{L}_{\text{corr}}$ on top further improves every split, with the largest $\delta^{\text{3D}}_{\text{avg}}$ gain on Simulation General (31.54 $\rightarrow$ 36.80).

\begin{table}[t]
\centering
\caption{\textbf{Ablation on training stages.} Iterations of a removed stage are reassigned to Stage~3, keeping the total number of iterations fixed.}
\label{tab:abl_stage}
\small
\setlength{\tabcolsep}{4.2pt}
\begin{tabular}{ccc cccccccc}
\toprule
 &  &  & \multicolumn{4}{c}{Simulation} & \multicolumn{4}{c}{Real-world} \\
\cmidrule(lr){4-7} \cmidrule(lr){8-11}
\multicolumn{3}{c}{Stage} & \multicolumn{2}{c}{Robot (500)} & \multicolumn{2}{c}{General (500)} & \multicolumn{2}{c}{Robot (100)} & \multicolumn{2}{c}{General (100)} \\
\cmidrule(lr){1-3} \cmidrule(lr){4-5} \cmidrule(lr){6-7} \cmidrule(lr){8-9} \cmidrule(lr){10-11}
1 & 2 & 3 & FVD$\downarrow$ & $\delta^{\text{3D}}_{\text{avg}}\!\uparrow$ & FVD$\downarrow$ & $\delta^{\text{3D}}_{\text{avg}}\!\uparrow$ & FVD$\downarrow$ & $\delta^{\text{3D}}_{\text{avg}}\!\uparrow$ & FVD$\downarrow$ & $\delta^{\text{3D}}_{\text{avg}}\!\uparrow$ \\
\midrule
\xmark & \xmark & \cmark & 70.45 & 48.50 & 85.34 & 34.29 & 262.40 & 60.86 & 58.11 & 31.75 \\
\xmark & \cmark & \cmark & \underline{69.83} & \underline{49.15} & \underline{83.27} & \underline{34.51} & \underline{259.42} & \underline{62.00} & \underline{58.02} & \underline{32.80} \\
\cmark & \xmark & \cmark & 77.50 & 39.51 & 99.81 & 29.38 & 292.53 & 49.77 & 62.49 & 27.04 \\
\rowcolor{gray!15} \cmark & \cmark & \cmark & \textbf{67.35} & \textbf{50.05} & \textbf{81.92} & \textbf{36.80} & \textbf{255.19} & \textbf{64.75} & \textbf{56.48} & \textbf{33.92} \\
\bottomrule
\end{tabular}
\end{table}

\subsection{Ablation Study for Training Stages}

\noindent{\bf Setup.}
\sname{} is trained in three stages: Stage~1 adapts the generator to multi-view input and camera conditioning with $\mathcal{L}_{\text{FM}}$, Stage~2 warms up the trajectory head with $\mathcal{L}_{\text{track}}$ while the generator is frozen, and Stage~3 jointly optimizes both with $\mathcal{L}_{\text{forecast}}$.
We remove Stage~1, Stage~2, or both, and reassign their iterations to Stage~3, so that every variant is trained for the same 12K iterations in total.

\noindent{\bf Results.}
As shown in Table~\ref{tab:abl_stage}, the full curriculum performs best on every split.
Removing Stage~2 is the most harmful, performing worse than training with Stage~3 alone on every split, with the largest drop on Real-world Robot ($\delta^{\text{3D}}_{\text{avg}}$ 64.75 $\rightarrow$ 49.77).
Removing Stage~1 leads to a smaller but consistent drop, most pronounced on Real-world Robot ($\delta^{\text{3D}}_{\text{avg}}$ 64.75 $\rightarrow$ 62.00).

\begin{table}[h]
\centering
\caption{\textbf{Ablation on the generator layer.} The trajectory head reads query and key features from the $l$-th self-attention layer of the generator.}
\label{tab:abl_layer}
\small
\setlength{\tabcolsep}{4.2pt}
\begin{tabular}{c cccccccc}
\toprule
 & \multicolumn{4}{c}{Simulation} & \multicolumn{4}{c}{Real-world} \\
\cmidrule(lr){2-5} \cmidrule(lr){6-9}
 & \multicolumn{2}{c}{Robot (500)} & \multicolumn{2}{c}{General (500)} & \multicolumn{2}{c}{Robot (100)} & \multicolumn{2}{c}{General (100)} \\
\cmidrule(lr){2-3} \cmidrule(lr){4-5} \cmidrule(lr){6-7} \cmidrule(lr){8-9}
Layer $l$ & FVD$\downarrow$ & $\delta^{\text{3D}}_{\text{avg}}\!\uparrow$ & FVD$\downarrow$ & $\delta^{\text{3D}}_{\text{avg}}\!\uparrow$ & FVD$\downarrow$ & $\delta^{\text{3D}}_{\text{avg}}\!\uparrow$ & FVD$\downarrow$ & $\delta^{\text{3D}}_{\text{avg}}\!\uparrow$ \\
\midrule
6 & 72.81 & 45.92 & 96.28 & 31.10 & 277.11 & 57.04 & 60.26 & 28.75 \\
12 & 69.45 & 48.30 & 85.76 & 34.90 & 266.16 & 61.21 & \underline{58.30} & \underline{31.88} \\
\rowcolor{gray!15} 18 & \textbf{67.35} & \textbf{50.05} & \textbf{81.92} & \textbf{36.80} & \textbf{255.19} & \textbf{64.75} & \textbf{56.48} & \textbf{33.92} \\
24 & \underline{68.76} & \underline{49.16} & \underline{84.96} & \underline{35.28} & \underline{259.84} & \underline{63.80} & 58.65 & 31.60 \\
\bottomrule
\end{tabular}
\end{table}

\subsection{Ablation Study for Layers}
\label{app:abl_layer}

\noindent{\bf Setup.}
The trajectory head reads query and key features from a single self-attention layer $l$ of the generator, following the DiT~\citep{peebles2023scalable} architecture. We vary $l \in \{6, 12, 18, 24\}$, where $l=18$ is the default.

\noindent{\bf Results.}
As shown in Table~\ref{tab:abl_layer}, attaching the head to the 18th layer performs best on every split.
Shallow layers perform clearly worse, \eg, FVD on Simulation General rises from 81.92 to 96.28 with $l=6$.
Moving the head deeper to $l=24$ causes a slight drop, increasing FVD by about 2--4\% across splits.

\begin{table}[t]
\centering
\caption{\textbf{Ablation on the trajectory refiner.} (a) Number of refinement iterations $M$. (b) Attention types of the refiner, with $M=4$.}
\label{tab:abl_refine}
\small
\setlength{\tabcolsep}{4.2pt}
\begin{tabular}{l cccccccc}
\toprule
 & \multicolumn{4}{c}{Simulation} & \multicolumn{4}{c}{Real-world} \\
\cmidrule(lr){2-5} \cmidrule(lr){6-9}
 & \multicolumn{2}{c}{Robot (500)} & \multicolumn{2}{c}{General (500)} & \multicolumn{2}{c}{Robot (100)} & \multicolumn{2}{c}{General (100)} \\
\cmidrule(lr){2-3} \cmidrule(lr){4-5} \cmidrule(lr){6-7} \cmidrule(lr){8-9}
Setting & FVD$\downarrow$ & $\delta^{\text{3D}}_{\text{avg}}\!\uparrow$ & FVD$\downarrow$ & $\delta^{\text{3D}}_{\text{avg}}\!\uparrow$ & FVD$\downarrow$ & $\delta^{\text{3D}}_{\text{avg}}\!\uparrow$ & FVD$\downarrow$ & $\delta^{\text{3D}}_{\text{avg}}\!\uparrow$ \\
\midrule
\multicolumn{9}{l}{\textit{(a) Refinement iterations}} \\
$M=1$ & 69.46 & 47.11 & 87.23 & 33.44 & 267.42 & 61.05 & 59.83 & 30.19 \\
$M=2$ & 68.77 & 49.00 & 83.25 & 35.75 & 260.49 & 62.18 & 57.07 & 33.20 \\
\rowcolor{gray!15} $M=4$ & \underline{67.35} & \textbf{50.05} & \textbf{81.92} & \textbf{36.80} & \underline{255.19} & \underline{64.75} & \textbf{56.48} & \textbf{33.92} \\
$M=8$ & \textbf{67.14} & \underline{49.96} & \underline{82.63} & \underline{36.75} & \textbf{255.01} & \textbf{64.92} & \underline{56.66} & \underline{33.58} \\
\midrule
\multicolumn{9}{l}{\textit{(b) Refiner attention}} \\
w/o spatial & \underline{70.10} & 47.54 & \underline{86.79} & 34.25 & \underline{267.59} & \underline{61.16} & \underline{57.31} & \underline{32.50} \\
w/o temporal & 70.87 & \underline{48.21} & 88.54 & \underline{34.28} & 268.79 & 59.64 & 59.18 & 30.77 \\
w/o cross-view & 74.95 & 46.49 & 91.14 & 29.89 & 272.50 & 57.09 & 61.07 & 29.95 \\
\rowcolor{gray!15} Full & \textbf{67.35} & \textbf{50.05} & \textbf{81.92} & \textbf{36.80} & \textbf{255.19} & \textbf{64.75} & \textbf{56.48} & \textbf{33.92} \\
\bottomrule
\end{tabular}
\end{table}

\subsection{Ablation Study for Refinement}
\label{app:abl_refine}

\noindent{\bf Setup.}
The trajectory head refines its estimates over $M$ iterations with a transformer whose attention is factorized over space, time, and views.
We vary $M \in \{1, 2, 4, 8\}$ and remove each attention type from the refiner.

\noindent{\bf Results.}
As shown in \Cref{tab:abl_refine}, performance consistently improves up to $M=4$, \eg, FVD on Simulation General decreases from 87.23 at $M=1$ to 81.92 at $M=4$.
Further increasing to $M=8$ changes every metric by less than 1 while doubling the cost of the head.
Among the attention types, removing cross-view attention causes the largest degradation on every split, \eg, increasing FVD by 11.3\% and lowering $\delta^{\text{3D}}_{\text{avg}}$ from 36.80 to 29.89 on Simulation General.
Removing temporal attention increases FVD more than removing spatial attention on every split, while the two are comparable in $\delta^{\text{3D}}_{\text{avg}}$ on the simulation splits.

\subsection{Ablation Study for Training Data}

\begin{figure}[t]
  \centering
  \includegraphics[width=\linewidth]{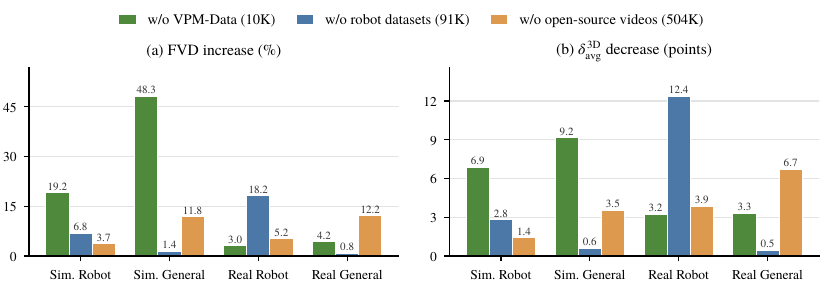}
  \caption{\textbf{Ablation on training data sources.} Relative FVD increase (a) and $\delta^{\text{3D}}_{\text{avg}}$ decrease (b) of \sname{} (V+T) when one source is removed from training, measured against the model trained on all sources. Higher is worse.}
  \label{fig:abl_data}
\end{figure}

\noindent{\bf Setup.}
Our training data combines three sources (\Cref{tab:train_data}): \sname{}-Data (10K samples), robot manipulation datasets with kinematics-grounded point tracks (91K), and open-source videos (504K).
We train \sname{} (V+T) while leaving out one source at a time and report the change in FVD and $\delta^{\text{3D}}_{\text{avg}}$ with respect to the model trained on all sources in Figure~\ref{fig:abl_data}.

\noindent{\bf Results.}
Each source mainly benefits the evaluation domain it covers.
Removing \sname{}-Data causes the largest degradation on the simulation splits, especially on Simulation General ($+$48.3\% FVD and $-$9.16~$\delta^{\text{3D}}_{\text{avg}}$), since it is the only source that pairs synchronized multi-view urban videos with persistent 3D trajectories. Removing \sname{}-Data also lowers $\delta^{\text{3D}}_{\text{avg}}$ on Real-world General by 3.31, suggesting that its accurate simulated trajectories transfer to real human motion.
Removing the robot datasets mainly affects Real-world Robot ($+$18.2\% FVD and $-$12.39~$\delta^{\text{3D}}_{\text{avg}}$), and removing the open-source videos mainly affects Real-world General ($+$12.2\% FVD and $-$6.73~$\delta^{\text{3D}}_{\text{avg}}$).
Removing any source degrades every split, indicating that the three sources are complementary.

\clearpage
\raggedbottom

\section{Qualitative Results}

\subsection{Multi-View Video and Point Trajectory Prediction}

\begin{figure}[H]
    \centering
    \includegraphics[width=\textwidth]{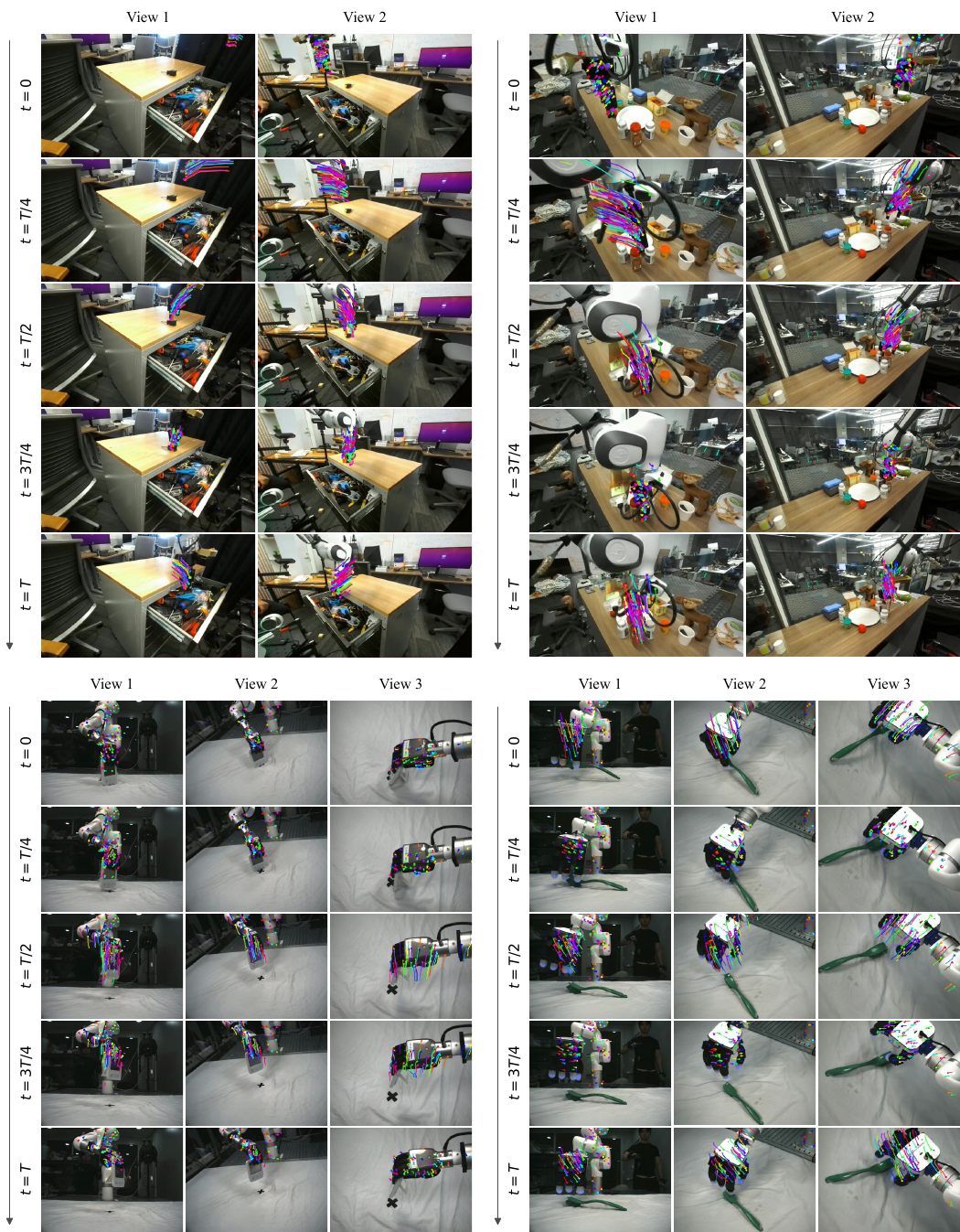}
    \caption{\textbf{Qualitative results of multi-view prediction.}
    Predicted video with overlaid predicted point track for two-view (top) and three-view (bottom) scenes.
    Conditioning frames are omitted.}
    \label{fig:qual_mvgen_pointgen}
\end{figure}

\clearpage

\begin{figure}[H]
    \centering
    \includegraphics[width=\textwidth]{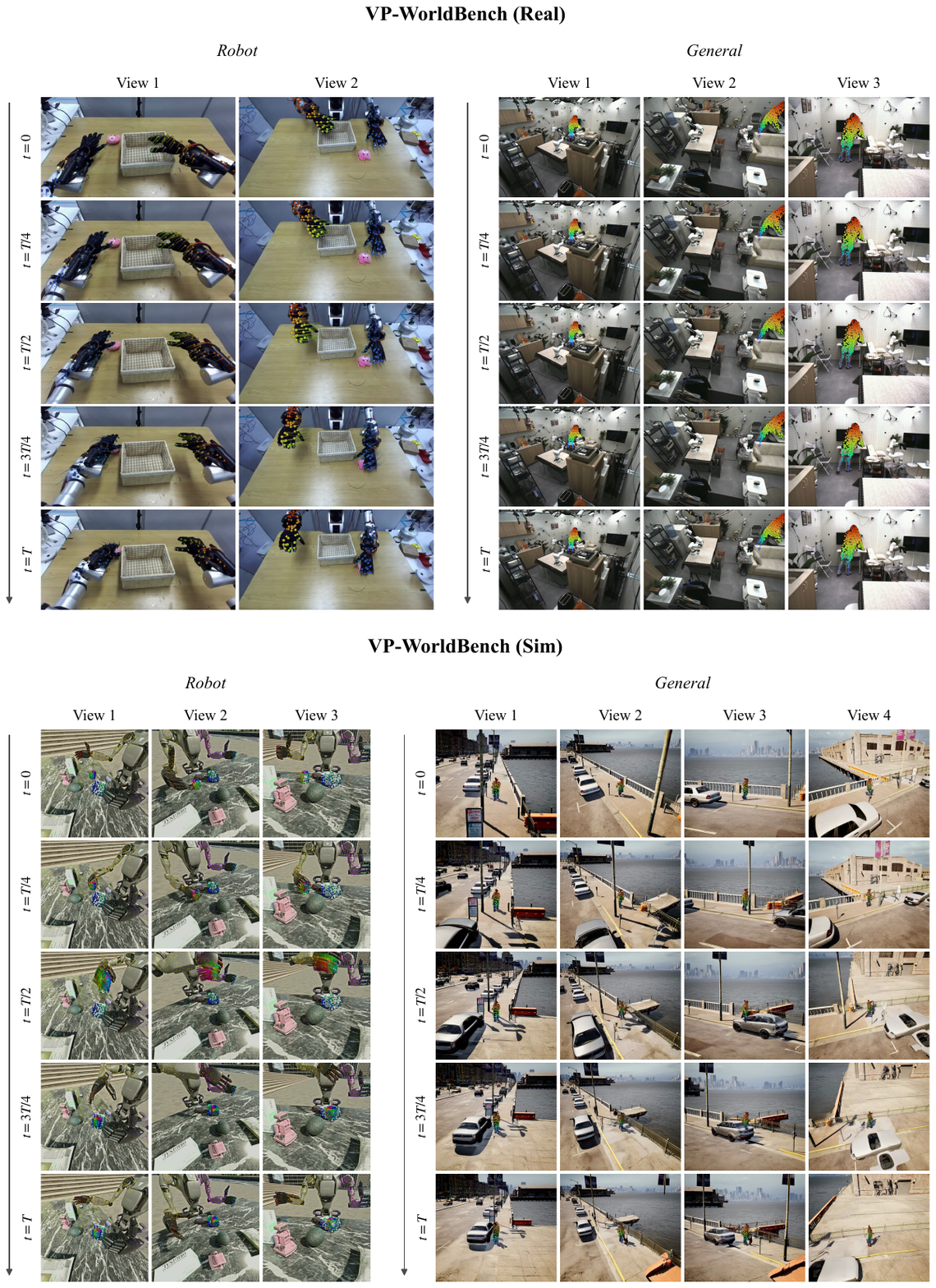}
    \caption{\textbf{Additional qualitative results of multi-view prediction.}
    Predicted video with overlaid predicted point track on robot and general scenes from VP-WorldBench (Real, top) and VP-WorldBench (Sim, bottom).}
    \label{fig:qual_mvgen_pointgen2}
\end{figure}

\clearpage

\subsection{Predicted vs.\ Ground-Truth Videos}

\begin{figure}[H]
    \centering
    \includegraphics[width=\textwidth]{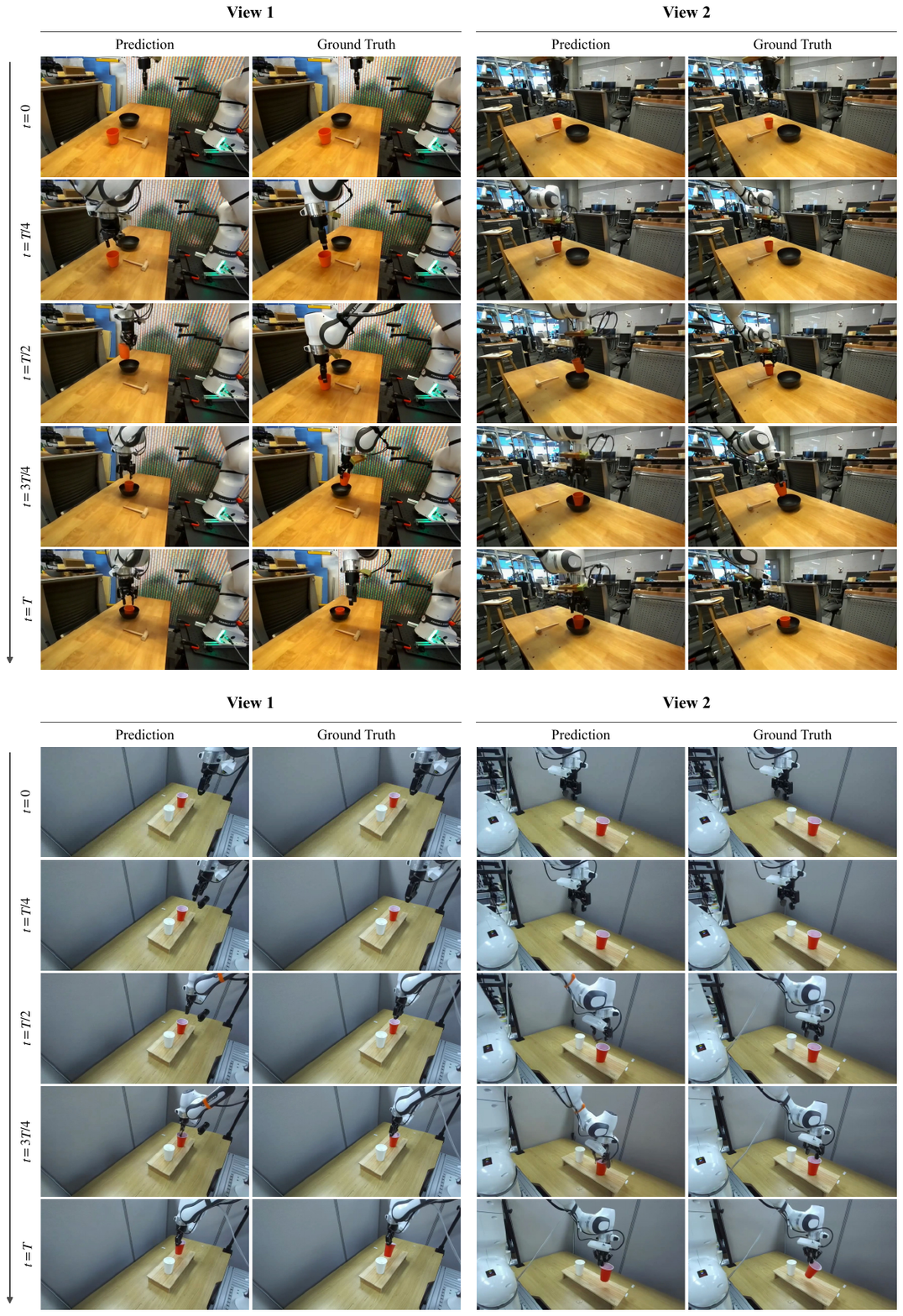}
    \caption{\textbf{Predicted vs.\ ground-truth videos.}
    Each view shows the prediction next to the ground truth.
    Conditioning frames are omitted.}
    \label{fig:pred_vs_gt}
\end{figure}

\clearpage

\subsection{Example Rollouts of Robot Manipulation Tasks}

\begin{figure}[H]
    \centering
    \includegraphics[width=\textwidth]{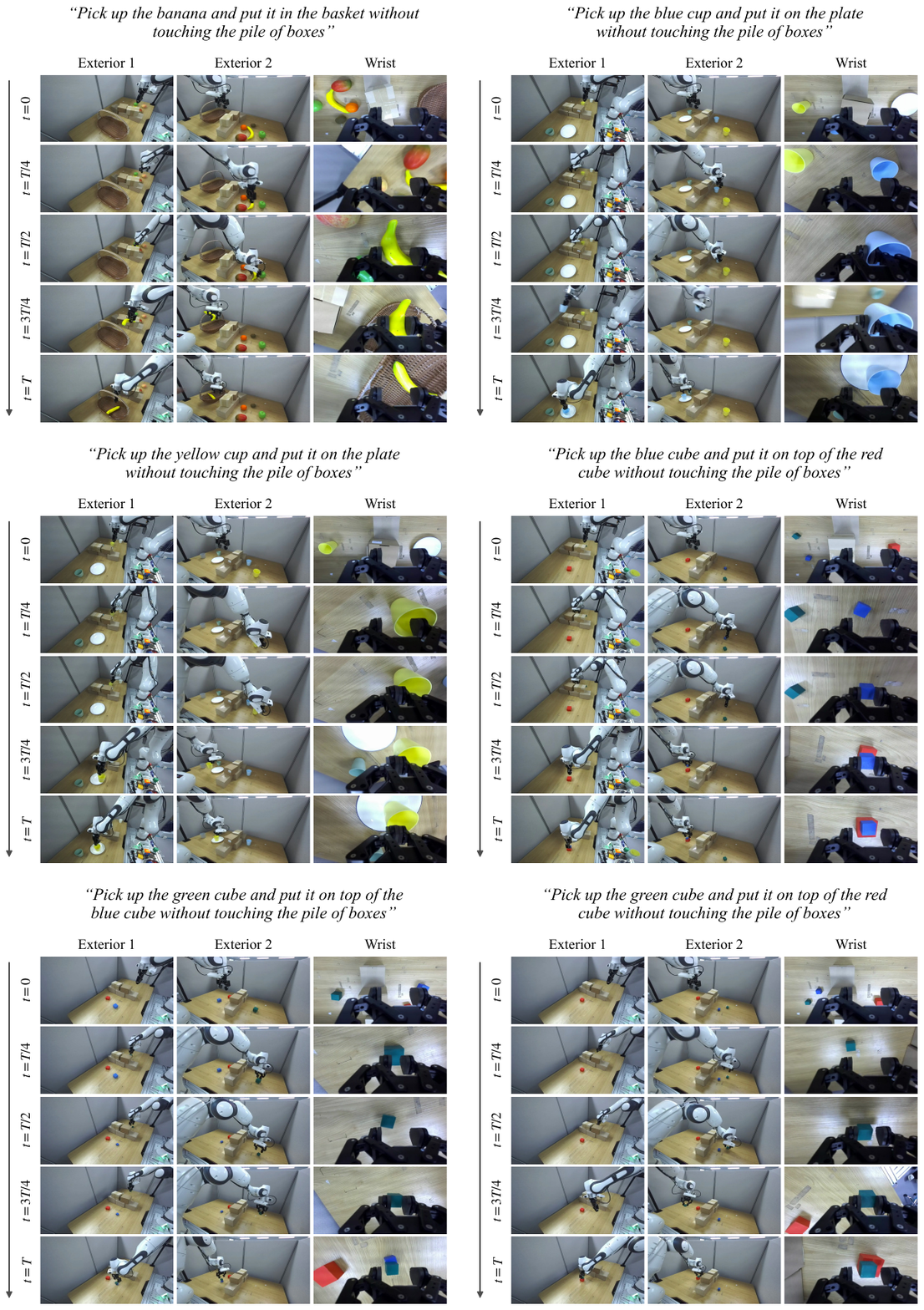}
    \caption{\textbf{Example rollouts of robot manipulation tasks.}
    Successful rollouts of our policy on six tasks, shown from two exterior cameras and a wrist camera.
    Rows are time steps sampled uniformly over each episode of length $T$.}
    \label{fig:droid_rollouts}
\end{figure}

\clearpage
\flushbottom

\end{document}